\documentclass[10pt]{article}
\usepackage[a4paper,margin=22mm]{geometry}
\usepackage{iftex}
\ifXeTeX\usepackage{fontspec}\else\usepackage[T1]{fontenc}\usepackage[utf8]{inputenc}\usepackage{lmodern}\fi
\usepackage{textcomp}
\usepackage{amsmath,amssymb,amsthm}
\usepackage{graphicx,booktabs,array,xcolor,longtable}
\usepackage[font=small,labelformat=empty,justification=justified,singlelinecheck=false]{caption}
\usepackage{microtype}
\usepackage[numbers,super,sort&compress]{natbib}
\usepackage[hidelinks]{hyperref}
\usepackage{xurl}


\newcommand{\figlabel}[2]{\textbf{#1 $|$ #2}}
\newcounter{edfig}

\title{Attention at Absolute Zero}
\author{Mostafa `Neo' Mohsenvand\\ {\small MIT Media Lab}}
\date{}

\begin{document}
\maketitle

\begin{abstract}
\noindent Attention scores the candidates that could support an answer, which we call witnesses, and reads those scores at
a temperature. We build a higher-order attention layer whose state lives on pairs of items, so that it can
answer questions about relations, such as whether one town can be reached from another. For a pair of items $a$
and $b$, every third item $k$ is a candidate witness, judged through its two legs $(a, k)$ and $(k, b)$. The layer reads
these witnesses at zero temperature, keeping the best witness and the number of witnesses that tie with it, and passes on
only bits that compare such counts. It is looped, rebuilding its pair state from the input and these bits at every
loop, and it is trained from answers alone. Each choice removes one of three ways in which soft attention
fails as inputs grow: its weight disperses over ever more candidates, learned scores cannot resolve the shrinking gaps
between them, and a soft state drifts from loop to loop. Trained in this way,
the layer stays exact far beyond the sizes it saw. In maximum retrieval it is right on at least 99.95\%
of sets a thousand times larger than any in training, on every one of ten seeds, where the same network trained with a soft count falls to
42\% at 4,096 items. On the thirty algorithms of CLRS-30, trained without hints, a general form of the
layer, which reads half its channels at zero temperature but embeds the input's numbers and carries a real-valued state
from loop to loop, beats the best published model, even though that model is trained with hints on some tasks; the full
cold layer is right on every test input of 9 of the 12 tasks whose answers depend on the
numbers only through comparisons. From yes-or-no answers it learns to decide 2-SAT and two-colouring exactly
at 500 variables and, as our theory predicts, not parity equations, which the layer cannot express. Beside a
frozen language model of 7 billion parameters it chains family relations exactly in contexts of 1,000 statements
after training on 16, where the model alone falls to 44\%, and finite checks on
the trained weights prove every block that learned correct at every size, on contexts without cycles and given what the
model reads. On CLUTRR, given
the stories' graphs, it is right on every story of five to ten relations and learns exactly the benchmark's
62 rules; from their text, with no hand-written rule, the model frozen and only a small read-in trained
from the answers, it lifts an open 7B model from 33\% to
65\%, above every system we found reported without hand-written rules except frontier reasoning
models.
\end{abstract}

\section{Introduction}

\begin{figure}[tbp]
\centering
\includegraphics[width=\linewidth]{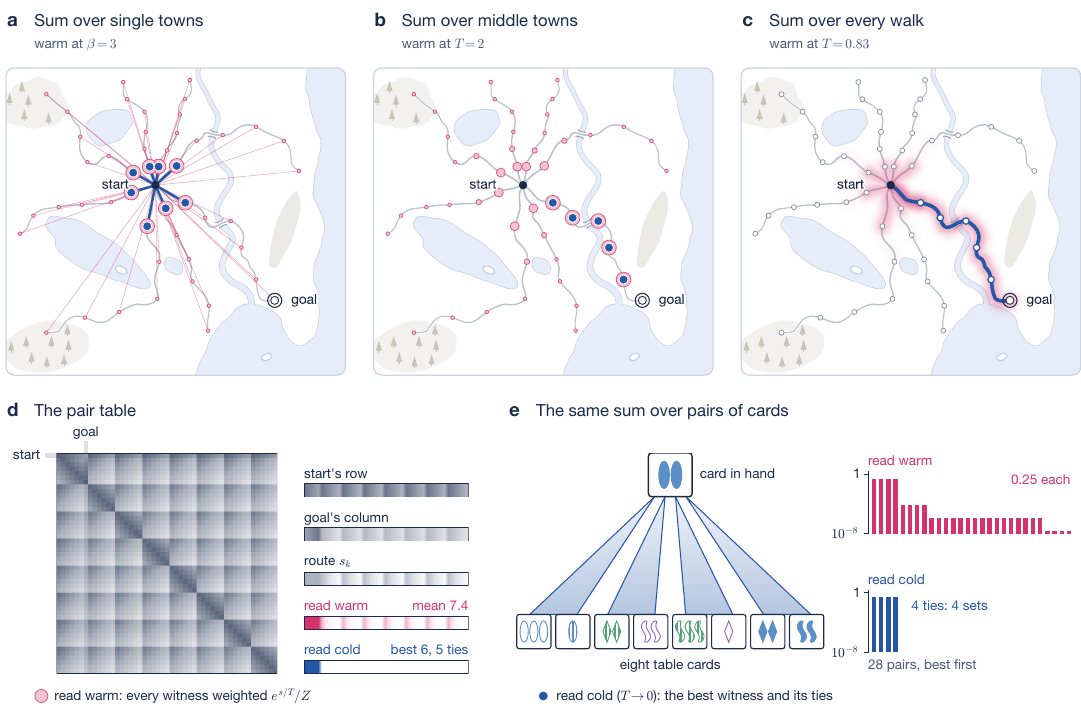}
\caption{\figlabel{Fig. \thefigure}{One sum over witnesses, read warm and read cold.} Each panel forms the sum
$Z_T = \sum_w e^{s(w)/T}$ over one family of witnesses and reads it in two ways: warm, every witness weighted by
$e^{s(w)/T}/Z_T$ (rose: area, glow or tint), and cold, $T \to 0$, which keeps the best witness and the number of
witnesses tied with it (cobalt). Nothing is trained. The map is one search question of 49 towns from the
benchmark of ref.~\cite{saparov2025transformers}, with roads counted in both directions. \textbf{a}, The start's lookup:
the witnesses are single towns, scored by their nearness to the start and read warm at the inverse temperature $\beta = 1/T = 3$; every town keeps some
weight, while the cold read keeps the eight towns next to the start, tied. \textbf{b}, The pair (start, goal): its
witnesses are the 47 middle towns $k$, each scored by minus the length $s_k$ of the route
start $\to k \to$ goal in roads, read warm at $T = 2$ roads; the cold read keeps the 5 towns of the
shortest route. \textbf{c}, Every walk from the start to the goal, weighted by its length and summed exactly by one matrix
inverse, read warm at $T = 0.83$ road: the warm weight glows along every road, and the cold read keeps the one
shortest walk. \textbf{d}, The pair table. Every ordered pair of towns keeps such a sum over its middle towns; the table is
tinted by the number of roads between the two towns. The start's row and the goal's column (light bands), unrolled into
strips over the 47 middle towns, give the legs from the start and the legs into the goal, which add up to the
route lengths $s_k$ of b, and below them the warm and cold weight of each middle town. Read
warm, the pair reports a mean length of 7.4 roads, which no route has; read cold, it reports the best
route, 6 roads, and its 5 ties. \textbf{e}, The same sum over the 28 pairs of
eight table cards in a SET deal, scored by a hand-set layer of four clocks at the inverse temperature tuned for 8 cards
($\beta = 2.68$). Read warm, the four pairs that complete a set with the card in hand
share 0.9997 of the weight, about a quarter each, so the number of sets is lost; read cold, the
four tied best pairs are kept, and their count is the number of sets. Schematic, not data: the map's
geography (water, woods, road curves and town positions draw the question's roads), the size of the smallest warm discs,
the width of the glow, the order of the towns in the table and the layout of the cards.}
\label{fig:big}
\end{figure}

An algorithm is correct on inputs of every size. Breadth-first search finds every town that can be reached from a
starting town on a map of ten towns or of ten thousand, by the same few steps. A neural network trained to carry out
such an algorithm is different: it is fitted on small inputs, and as the inputs grow it usually loses accuracy. The
CLRS-30 benchmark measures this gap on thirty algorithms from a standard textbook, training on graphs and arrays of at
most 16 items and testing at 64\cite{velickovic2022clrs}, and much of the progress on it has come from hints, which
supervise a network on the algorithm's intermediate states as well as on its final answer\cite{ibarz2022generalist}.
This paper builds an attention layer that needs no hints and stays exact far beyond the sizes it was trained on. Trained
on answers alone, a general form of it, which keeps part of its state as real numbers, beats the strongest published model
on CLRS-30, FloydNet, even though 10 of
FloydNet's thirty scores come from runs with hints\cite{yu2026floydnet}, and on a task where softmax attention is known
to fail it is still right at a thousand times its training size. To explain how, we follow two small examples from the
point of view of an attention layer (Fig.~\ref{fig:big}).

Attention answers a question by scoring the candidates that could support an answer, which we call
witnesses\cite{mohsenvand2026witnesses}, and by reading those scores together. On a road map, the question ``which town
is nearest the start?'' has single towns as its witnesses: each town is scored on its own, and the reading picks the
best. We call such a question a lookup, and ordinary attention, which scores each item against a query, is built for
it\cite{vaswani2017attention}. The question ``can the goal be reached from the start by two roads?'' is of another kind.
Its witnesses are the middle towns $k$, and a middle town counts only if one road runs from the start to $k$ and another
from $k$ to the goal, so its score depends on two pairs of towns at once. We call this a join. A layer whose state lives
on single towns can answer a join only if each town has already gathered, in its own state, the facts that the join
needs; transformers trained to search graphs learn to do exactly this, and they stop learning as the graphs
grow\cite{saparov2025transformers}. A layer whose state lives on pairs of towns can test both legs of the route directly.

\begin{figure}[tbp]
\centering
\includegraphics[width=\linewidth]{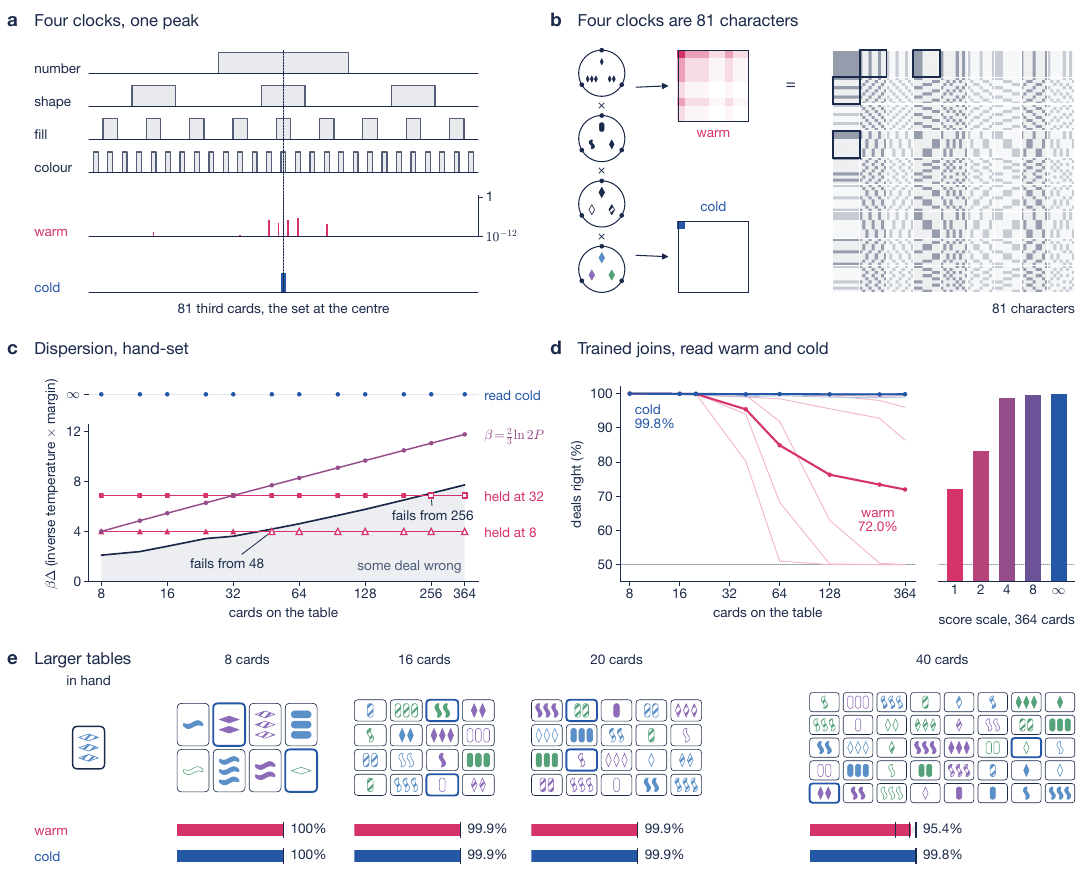}
\caption{\figlabel{Fig. \thefigure}{SET: read warm, the weight spreads; read cold, the set is kept.} A deal is a card in
hand and a table of cards, and the question is whether two table cards complete a set with the card in hand.
\textbf{a}, One join trained in the first version of this paper\cite{mohsenvand2026witnesses}, with the card in hand and one table card fixed:
each attribute's clock over the 81 possible third cards (number, shape, fill and colour, turning 1, 3, 9 and 27 times
across them), whose product peaks only on the card that completes the set (centre). Read warm, every third card keeps
some weight (log scale; the next largest is $7 \times 10^{-7}$); read cold, only the set. \textbf{b}, The four clocks
are four of the 81 characters of $(\mathbb{Z}_3)^4$ (outlined), and their product at the learned inverse temperature,
$\beta = 4.86$, is one bright card on the $9 \times 9$ map of cards: warm, a haze with one peak; cold, one card.
\textbf{c}, Dispersion with weights set by hand: SET with six attributes, inverse temperature times margin,
$\beta\Delta$, against the number of cards on the table (1,200 deals per size); below the navy boundary some deal is read
wrong. Held at its value for 8 cards, the temperature fails from 48 cards, and held at its value for 32
cards, from 256; grown as the size law says, $\beta = \ln(2P)/\Delta = \tfrac23 \ln 2P$ for $P$ pairs of cards (a set outscores the nearest miss by $\Delta = 3/2$), it is right on
every deal to 364; read cold, the layer is exact at every size. \textbf{d}, The six joins of the first version of this paper\cite{mohsenvand2026witnesses} trained at 8 cards with all 81 cards in hand, on deals with no set or exactly one:
read warm as published (rose: their mean and each join) they fall to 72.0\% at 364 cards, with
3 of the six at chance, while the same weights read cold (cobalt) stay at
99.8\% (lowest join 99.0\%). Right: the same joins at 364 cards with their scores
multiplied by 1, 2, 4 and 8 (72.0, 83.1, 98.7 and 99.7\%) and read
cold (99.8\%); the cold read is the end of that dial. Cards repeat from 64 cards on, since 81 distinct
cards allow no table without a set beyond 40. \textbf{e}, Deals with exactly one set at 8, 16, 20 and 40 cards (the
completing pair outlined), and below each the trained joins' accuracy at that size read warm and cold, never rounded up
to 100\%. Schematic, not data: the drawing of the clock dials, the grey tones of the characters, the order of the cards
in the deals and the logarithmic tint of the warm product in b, whose weights are exact.}
\label{fig:set}
\end{figure}

The card game SET makes the same point without a map (Fig.~\ref{fig:set}). Each card has four attributes, such as colour and shape, and three
cards form a set when every attribute is either the same on all three or different on all three. Asked whether two of
the cards on the table complete a set with the card in hand, a layer must judge pairs of table cards, and no score given
to single cards ranks those pairs correctly on every table\cite{mohsenvand2026witnesses,zweig2022exponential}.
Higher-order attention answers both examples by keeping a state for every pair of items and updating it through a third.
For a pair $(a, b)$, every other item $k$ is a candidate witness, judged through the two legs $(a, k)$ and $(k, b)$, just
as one step of the Floyd--Warshall algorithm judges a route from $a$ to $b$ through $k$. Layers of this shape are the core
of protein structure prediction\cite{jumper2021highly}, of the Edge Transformer for systematic
reasoning\cite{bergen2021systematic} and of FloydNet\cite{yu2026floydnet}. Looped, such a layer can double its reach at every loop, because each loop can join two routes
found by the one before\cite{fischer1971boolean}, so that $L$ loops cover routes of $2^L$ roads; whether a trained layer
does so depends on what it learns, and we show where it does.

How the witnesses are read decides whether such a layer stays exact as the input grows. Attention scores each witness
$w$ and forms the sum $Z_T = \sum_w e^{s(w)/T}$, which gives witness $w$ the weight $e^{s(w)/T}/Z_T$; the temperature $T$
sets how sharply the weight falls from the best witness to the rest. At a high temperature the weight is spread over
many witnesses, and at a low one it gathers on the best. To single out the right witness among $n$ others, the margin by
which it wins, divided by the temperature, must exceed about $\log n$\cite{chiang2022overcoming,mohsenvand2026witnesses}.
A larger map has more middle towns, and a larger table more pairs of cards, so a reading that is sharp enough on the
inputs of training can blur on the inputs of testing.

This blur takes three distinct forms, which we separate by measuring each on a task that shows it alone (Section~3). The
first, dispersion, is that a softmax at a fixed temperature spreads its weight as candidates are
added\cite{velickovic2025softmax}. On SET, a layer whose weights we set by hand, so that nothing else can go wrong, reads
correctly up to the size its temperature was tuned for and fails beyond it: tuned for tables of 8 cards, it fails from
tables of 48, and the joins trained on SET in the first version of this paper fail in the same way when their tables grow
(Fig.~\ref{fig:set}c,d). The second, resolution, is subtler. The gaps between witnesses narrow as the input grows, and the learned
scores must then resolve differences that training never asked them to see. In maximum retrieval, where a network must
pick the item of highest priority from a set, the softmax networks we trained carry offsets of a few thousandths of the priority range,
invisible on the 16 items of training and decisive among thousands. The third, carried error, belongs to looped
networks. A continuous state passed from loop to loop drifts, and on depth-first search the margin between the right
parent and a wrong one, comfortable at the depths seen in training, erodes deeper in the search tree until the network
names the wrong parent. Reading at zero temperature, as a maximum does, cures the first failure but not the other two,
because what a maximum passes on is still a learned, continuous score.

Each failure has its own remedy, and the three remedies fit in one layer, which we call cold attention. The first remedy
is to read at zero temperature. What survives that limit is exact: as $T$ falls, the sum $Z_T$ is governed by its best
score, and the next term of the limit is the number of witnesses that reach that score\cite{litvinov2007maslov}, so a
layer that keeps the best witness and its tie count keeps everything that a cold enough softmax would see, with no
temperature left to set. The second remedy is to compare inside the layer. Instead of learning score offsets that stand
in for comparisons, the layer counts the witnesses that pass learned tests on their two legs and passes on only two bits
for each pair of counts, whether one is larger and whether the two are equal. The third remedy is to rebuild the pair
state at every loop from the input and the bits found so far, so that nothing continuous is carried from one loop to the
next and nothing can drift. We train the layer from final answers alone, with its hard tests in the forward pass from the
first update and straight-through gradients behind them\cite{bengio2013estimating} (Section~4 and Figs.~\ref{fig:arch} and~\ref{fig:layer}).

We first test the layer where softmax was shown to fail\cite{velickovic2025softmax}. In maximum retrieval each item's
witnesses are the rivals that outrank it, and the cold count picks the item that has none. Trained on sets of at most
16 items, the layer is right on at least 99.95\% of sets of every size up to
16,384, a thousand times the largest set it saw, on every one of ten seeds; at 4,096 items the best
published method is right on 62.7\%\cite{vasylenko2026long}. A contrast within our own runs shows where
the failure lies. A network that compares priorities inside the layer but is trained with a soft count falls to
42\% at 4,096 items, yet its own weights, read at zero temperature, are right on at least
99.96\% of sets at every size on every seed: the network had learned the right comparisons, and its soft reading
lost them (Fig.~\ref{fig:fail}b).

On CLRS-30, trained without hints, the general block beats the best published model even when that model uses hints
(Fig.~\ref{fig:clrs}). The general block takes only part of the recipe: half of its channels count witnesses and pass on
comparison bits, but the other half keep the strength of the best witness as a real number, the input's numbers enter as
embeddings, and its state is carried from loop to loop. With the input's numbers entering the block as embeddings, one
configuration trained once per task averages 97.04\% over
three seeds, against FloydNet's published 96.64\%, and leads FloydNet's own runs without hints by
2.2 points. A single block trained on all thirty tasks at once does better still: it averages
97.20\%, above one block per task and above FloydNet with its hints, with each of its three seeds above
FloydNet. When each task takes the setting our theory prescribes, a cold block, into
which numbers enter only through comparisons made inside the layer, on the 12 tasks whose answers
depend on the numbers only through comparisons, and the general block on the other 18, the average
rises to 97.98\% on the benchmark's standard inputs; Section~6 marks which of those choices were made after
seeing results. In this cold setting depth-first search becomes exact on every graph of 64, 96 and 128
nodes, and every answer provably stays the same when the input's numbers are replaced by any others in the same order.

The largest errors that remain each have a cause that we can name. Strong components ask about cases that small training
graphs rarely contain, string matching needs a count that the layer cannot store, and the optimal search tree needs a
recurrence that gradient descent does not find; segment intersection we have not fixed. For strong components we also let
the learner find its missing cases itself. A committee of saved models is shown small graphs changed one edge at a time
and keeps those on which its members disagree, and adding them to the training pool lifts the official score from
90.0\% to 97.1\% over three seeds, close to the 98.21\% that our
hand-designed graphs give (Fig.~\ref{fig:committee}).

The theory also draws a sharp line around the yes-or-no questions that the layer can decide. A constraint problem asks
whether variables can be given values that satisfy a set of constraints, and a class of such problems has bounded width
when checking small groups of variables for consistency is enough to decide it; by Barto's theorem, checking pairs of
variables through a third is then enough\cite{barto2016collapse}. For constraints between two variables over a
finite domain, a pair block that sees no positions decides a class at every size if and only if the class has bounded
width, and for constraints on more variables the same holds once each constraint has a token of its own. If the class
has bounded width, one looped block with fixed weights decides it on every instance: its kept bits record, for each
pair of variables, which of the $|D|^2$ pairs of values from the domain $D$ have been ruled out, and it comes to rest
within $n^2 |D|^2 + 2$ loops. If it does not, no network whose answer depends only on what pair refinement sees decides
it, whatever its weights, its join or its number of loops, even with a different network for each size. The two halves
rest on Barto's collapse of bounded width to pairs\cite{barto2016collapse} and on the counting-logic lower bound of
Atserias, Bulatov and Dawar, a construction of the Cai--F\"urer--Immerman kind that they carry to every class without
bounded width\cite{cai1992optimal,atserias2009affine,barto2014local}; the statement for networks is ours. 2-SAT and two-colouring have bounded width, and parity equations over three
variables do not. Trained on yes-or-no answers alone, on
at most 16 variables for 2-SAT and 32 for two-colouring, the layer is right on every
instance at 500 variables on three seeds of each, apart from a single 2-SAT instance on one seed (Fig.~\ref{fig:csp}d). On parity
equations, as predicted, its answer does not depend on whether the instance can be satisfied.

Language models read single sentences well and chain many of them poorly, and the pair layer can do the chaining for
them. Beside a frozen language model of 7 billion parameters\cite{qwen2024qwen25}, which reads each sentence, the layer
learns from yes-or-no answers to chain family statements such as ``Maria Lopez is the mother of Tom Baker'', written in
eight wordings, two of them never seen in training. Trained on contexts of at most
16 statements, it answers every question of the form ``is $X$ an ancestor of $Y$?'' in contexts of
1,000 statements, on each of six seeds, where the model alone falls to
44\% (Fig.~\ref{fig:organ}); on a simpler task, the general block of Section~4, which carries a continuous state from loop to loop and reads the same inputs,
falls to 71\% at 256 facts. Because each trained layer is a finite table, two finite checks on its
weights decide whether it is right at every size, and they prove every one of the 23 trained layers that
learned correct on contexts of every size, given what the model reads; for nine of them the proof is also
checked in Lean. On the CLUTRR benchmark of crowd-written family stories\cite{sinha2019clutrr}, given the stories' graphs, the
layer is right on every story of five to ten relations, the top of the published table, and misses only
nine short stories in which the benchmark's own rules derive two different relations. From the stories' text, with no rule
written or supplied by hand, the model frozen and a small read-in trained from the answers alone, it lifts the open 7B
model from 33.3\% to 64.6\% of the test stories. That is above every other system
we found reported on this test without hand-written rules, among them Llama-3-70B, ten times its size
(56.3\%), and GPT-4o (57.6 to 61.9\%)\cite{zhang2026extracting}; only frontier
reasoning models match or beat it, the best at 73.2\%.

The layer has two costs. It keeps a state for every pair of items and does work cubic in their number at every loop, and
depth-first search remains sequential, so its loops must grow with the input.

\section{Background}

Our layer draws on four lines of work: networks that learn to execute algorithms, the known loss of sharpness of soft
attention as inputs grow, the zero-temperature limit of a sum, and the logic that describes what networks over pairs can
compute. We take them in turn, and close with what this paper adds to each.

CLRS-30 fixes thirty algorithms from a standard textbook, together with their inputs and the way their answers are
scored\cite{velickovic2022clrs,cormen2009introduction}. Its strongest results train on the algorithms' intermediate
states as hints\cite{ibarz2022generalist}, while a second line of work trains on answers
alone\cite{mahdavi2023towards,bevilacqua2023neural,rodionov2023neural,georgiev2024deep}. Bohde et al.\ observed that each
intermediate state of these algorithms fixes the next one, and rebuilt the network's state at every step from the input
and its own prediction of the current hints, discarding the state it would otherwise carry\cite{bohde2024markov};
Mahdavi et al.\ found no gain from rounding the carried state to discrete codes\cite{mahdavi2023towards}. Our cold loop
also rebuilds its state at every step, but from the input and from bits that the layer itself computes at zero
temperature for every pair of items; nothing it rebuilds from predicts the textbook's state, and no target supervises it.
Rodionov and Prokhorenkova already let a network trained without hints feed back a soft state of its own on every edge,
beside the carried one\cite{rodionov2023neural}. Discrete neural algorithmic reasoning makes the executor's states
discrete and its attention hard, so that every transition can be checked one by one\cite{rodionov2025discrete}; it
learns those states from the benchmark's hints at every step, feeding the true states back in during training and
running for the true number of steps, and without hints its breadth-first search labels 79\% of nodes
correctly at 64 nodes. FloydNet, the strongest published model, keeps a state for every pair of items and updates
it through a middle item, as a step of the Floyd--Warshall algorithm does\cite{yu2026floydnet}; our layer has the same
shape and differs in how it reads the middle items. Whether a learner finds the right rule depends on its training sample
as much as on its architecture. Small samples can pin an algorithm down exactly when they contain the right
cases\cite{nerem2025graph,wittig2026which}, and transformers trained to search a graph, the task of the road map in the
introduction, find it harder to learn as the graphs grow\cite{saparov2025transformers}.

That soft attention loses its sharpness as inputs grow is well established. A softmax at a fixed temperature spreads its
weight over more candidates as they are added\cite{velickovic2025softmax}, and separating one candidate from $n$ others
requires a margin that, scaled by the inverse temperature, grows like $\log n$\cite{chiang2022overcoming,hayase2026unified}.
Scaling the temperature with $\log n$, or replacing the softmax by a sparse map, follows from this
law\cite{vasylenko2026long}. The first version of this paper stated the same law for sums over witnesses of any kind,
such as the pairs of cards in SET, and checked it on several layers\cite{mohsenvand2026witnesses}; this paper asks what
remains when the temperature is taken to zero.

Reading attention at zero temperature gives saturated, or hard, attention, whose expressive power the theory of
transformers has characterised\cite{merrill2022saturated,merrill2023logic,strobl2024what}. As the temperature falls, a
log-sum-exp tends to a maximum, the tropical limit of Maslov's dequantisation\cite{litvinov2007maslov}; this limit is
why dynamic programs can be smoothed into differentiable ones\cite{mensch2018differentiable}, and why tropical attention
reads a maximum over witnesses\cite{hashemi2025tropical}. Layered tropical numbers keep a count beside the
maximum\cite{izhakian2010layered}, and that pair, the best score together with the number of witnesses that tie with it,
is the leading term that our layer keeps.

Logic describes what such networks can compute. Message-passing networks distinguish exactly what the Weisfeiler--Leman
test distinguishes, and networks with a state for every pair of items what its folklore two-dimensional form
distinguishes\cite{morris2019weisfeiler,maron2019provably,cai1992optimal}. Looped networks have been placed in logic with
floating-point numbers\cite{ahvonen2024logical}, with maximum or minimum aggregation\cite{bonet2026recurrent}, and with
unbounded precision, where a looped network reaches the full message-passing limit\cite{grohe2021logic,
rosenbluth2026repetition}. We train hard tests with straight-through gradients\cite{bengio2013estimating}.

What we add is a layer whose zero-temperature limit keeps the tie count beside the best witness and passes on only
comparisons, trained hard from the first update on answers alone, and measured against each of the three ways in which
soft attention fails, on CLRS-30 without hints and beyond it: on constraint problems, where the logic above draws a sharp
line around what the layer can decide, and beside frozen language models, where each trained block can be proven correct
at every size.

\section{Three ways soft attention fails at size}

\begin{figure}[tbp]
\centering
\includegraphics[width=\linewidth]{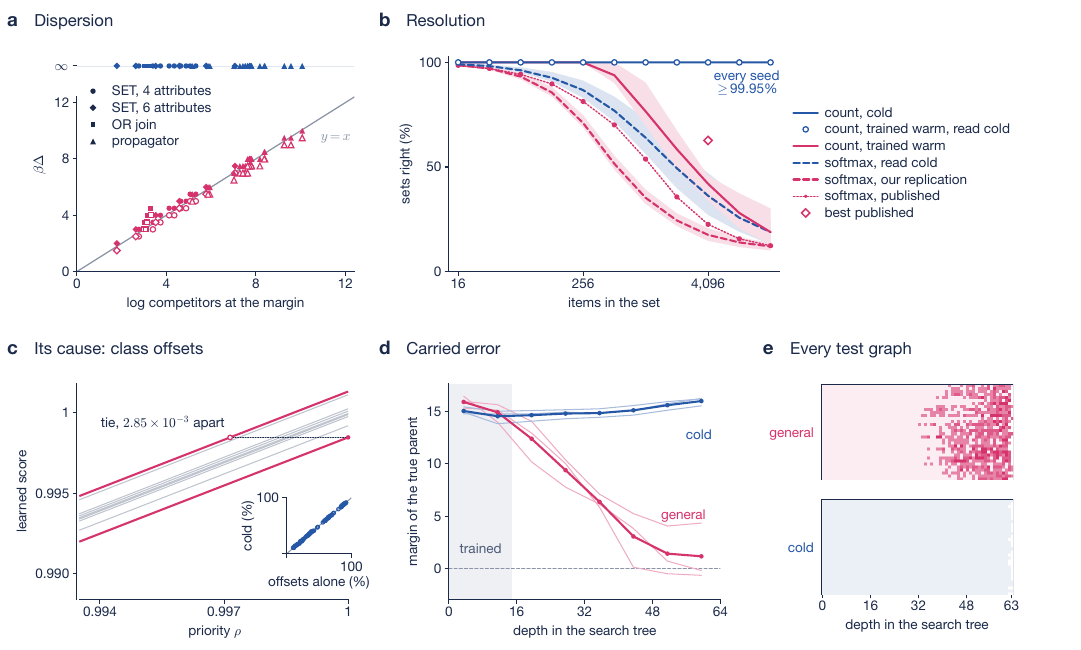}
\caption{\figlabel{Fig. \thefigure}{Three ways soft attention fails at size.} \textbf{a}, Dispersion: the size law across
four hand-set layers, nothing trained (the instances of the first version of this paper\cite{mohsenvand2026witnesses}: SET with four and
six attributes, the one-layer OR join and the propagator on random acyclic graphs). Each instance is shown at the two
values of $\beta\Delta$, half a nat apart, that bracket its critical value, read warm: filled, every question right; open,
some wrong; against the logarithm of the competitors at the margin. Grey: $y = x$. Top row: the same layers read cold,
exact by construction. \textbf{b}, Resolution, measured on maximum retrieval (pick the class of the item of highest
priority; trained on sets of 5 to 16 items; ten seeds each, 10,240 test sets per size and seed; lines, the mean; bands,
lowest to highest seed): comparisons inside the layer read cold (count, cold; every seed at least 99.95\% at
every size), the same comparisons trained with a soft count and read warm or cold (count, trained warm), and a softmax network from our replication,
which falls up to 18.5 points below the published curve, read warm or cold, beside the published softmax
curve\cite{velickovic2025softmax} and the best published method at 4,096 items\cite{vasylenko2026long}. \textbf{c}, The
cause of resolution: one softmax network's learned key score near the top of the priority range, one line per class (the
two extreme classes in rose, the other eight grey), in priority units; the top item of the lowest class ties an item of
the highest class about three thousandths of the priority range lower. Inset: accuracy read cold, predicted from the class
offsets alone against measured, for 20 networks at four sizes. \textbf{d}, Carried error:
depth-first search on the official 64-node test, the margin of the true parent against the node's depth: the logit of the
true parent minus the largest logit of any other candidate, in raw logits, after 64 loops (median per bin of eight depths;
three seeds, thin, and their mean, bold), for the general block, which carries a continuous strength from loop to loop, and
the cold block with positions compared, which rebuilds its bits every loop. The switch also replaces real strengths by
bits and moves positions into comparisons; with positions encoded instead, the cold bit loop keeps a margin of
10.0 near the root and 9.1 at depth 63, but loses accuracy at 128 nodes
(Fig.~\ref{fig:depth}c). Below zero the parent named is wrong; grey, the depths seen in training.
\textbf{e}, Every official test graph (rows) by depth in its depth-first forest (columns): the parents named wrong, summed
over three seeds (darker, more seeds wrong), for the general block after 64 loops (96.63, 76.03 and
87.60\% right) and the cold block at its registered read (every parent right on every seed). Pixel rows and columns
are cells of the data; nothing in this figure is schematic.}
\label{fig:fail}
\end{figure}

An attention layer scores each witness $w$ of an answer and reads the sum
\begin{equation}
Z_T = \sum_{w} e^{s(w)/T}
\label{eq:z}
\end{equation}
at a temperature $T$, attending to witness $w$ with weight $e^{s(w)/T}/Z_T$. On the road map of Fig.~\ref{fig:big}b the
witnesses of the pair (start, goal) are its 47 middle towns, each scored by minus the length of the route
through it; read warm, the weight is spread over all of them, and the pair reports an average length of
7.4 roads, which no route has, while read cold it reports the shortest route, 6 roads, and the
5 towns that lie on routes of that length. As the input grows, the warm reading fails in three distinct ways.
We separate them by choosing, for each, a task on which it fails alone (Figs.~\ref{fig:set} and~\ref{fig:fail}).

Dispersion is the failure that the temperature itself causes. At a fixed temperature, the weight on the right witness
shrinks as competitors are added\cite{velickovic2025softmax}. To see it with nothing else going wrong, we read SET with a
layer whose weights are set by hand. A temperature held at its best value for tables of 8 cards fails from
48 cards, and one held at its best value for 32 cards fails from 256; a temperature that falls as
the logarithm of the number of competing pairs grows is right at every size we tried, up to 364 cards
(Fig.~\ref{fig:set}c), and the same law holds across the other layers of the first version of this paper (Fig.~\ref{fig:fail}a). Trained networks fail in the same way. The six joins that our earlier work trained on
tables of 8 cards\cite{mohsenvand2026witnesses}, read as published, fall to 72.0\% of deals with no set or exactly one at
364 cards, with 3 of the six at chance, because the one pair that completes the set is
outweighed by the many that do not. The same weights read cold stay at 99.8\% (lowest join
99.0\%), and multiplying the scores by 2, 4 and 8 moves the warm read step by step towards the cold one,
to 83.1, 98.7 and 99.7\% (Fig.~\ref{fig:set}d). The cold read is the end of the dial
that a softmax turns as its temperature falls, and reading at zero temperature removes dispersion altogether.

Resolution is the failure that zero temperature cannot fix, because the scores themselves are wrong in their last
digits. We trained the softmax network of the maximum-retrieval task\cite{velickovic2025softmax}, in which a network must
pick the item of highest priority from a set, on sets of at most 16 items, with ten seeds under each of two readings of
its weight penalty, and read its learned key score. Our runs fall below the published curve, by up to
18.5 and 11.4 points under the two readings (Supplementary Note~4), so we use them to
find the cause of the failure and not as the published method's score. Within each class of item the score rises with
priority, but each class carries its own fixed offset, between 1.9 and 6.0
thousandths of the priority range across 20 trained networks (Fig.~\ref{fig:fail}c). Sets of 16
items never ask the network to rank two items that close, so nothing in training removes the offsets; at thousands of
items, neighbouring priorities are closer than the offsets, and the network ranks by class instead of by priority. Read
at zero temperature, the same weights are right on only 36.3\% of sets of 4,096 items, and the offsets
alone predict each network's accuracy there to within 0.5 points ($r = 0.99$).
The remedy is to compare priorities inside the layer, where no learned offset can reorder them.

Carried error belongs to looped networks. The general block of Section~4 passes a continuous strength from loop to loop,
so it must keep the margin between the right answer and a wrong one intact at every loop, and on depth-first search it
does so only near the depths it was trained on. There the margin is about 15, but deeper in the search tree it falls to
between -0.6 and 4.4, and once it falls below zero the parent the block names is
wrong (Fig.~\ref{fig:fail}d,e). A block that rebuilds its state at every loop from the input and the bits computed so far has
no margin to lose, because its bits are made anew at each loop rather than carried from the last; hinted reasoners that
rebuild their state from their own predictions of the algorithm's state drop the carried history in the same
way\cite{bohde2024markov}.

Depth sets a separate limit, which no reading removes. A loop over two-step routes can double the reach of the state, so
that $L$ loops cover routes of $2^L$ roads, whereas depth-first search cannot be doubled and needs a number of loops that
grows with the input (Section~5). Whether a trained layer does double depends on what it has learned, and we report where
it does and where it does not (Sections~4 and~6). That bound concerns the worst case; the fade measured here follows the carried
margin. Each of the three failures thus suggests its own remedy, and the next section builds a layer that applies all
three.

\section{The cold layer}

The three remedies meet in one layer. Its starting point is what survives the zero-temperature limit of the sum of
equation~(\ref{eq:z}). As $T \to 0$, the sum keeps two exact quantities: the best score $s^\ast = \max_w s(w)$ and the
number of witnesses that reach it, $g^\ast = |\{w : s(w) = s^\ast\}|$. Indeed $T \log Z_T \to s^\ast$ and
$Z_T e^{-s^\ast/T} \to g^\ast$, and whenever $(s_1^\ast, g_1^\ast)$ exceeds $(s_2^\ast, g_2^\ast)$ in lexicographic order,
$Z_T(F_1) > Z_T(F_2)$ at every small enough temperature (Proposition~1). Comparing these pairs is therefore what a warm
sum does once its temperature is low enough, and a layer that compares them directly has no temperature left to set. On
the map, the pair (start, goal) keeps the best route of 6 roads and its 5 ties; in the SET deal, the
best score is that of a set and its tie count is the number of sets (Fig.~\ref{fig:big}d,e).

\begin{figure}[tbp]
\centering
\includegraphics[width=\linewidth]{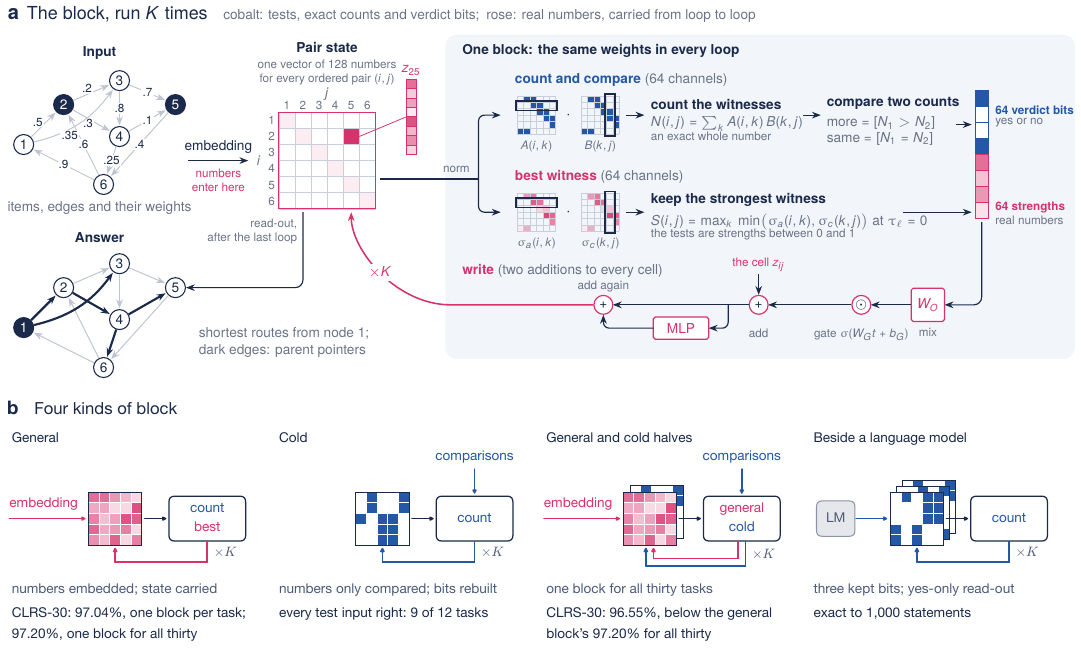}
\caption{\figlabel{Fig. \thefigure}{One block, run $K$ times, and the four kinds of block we train.} Cobalt: the yes-or-no
tests, exact counts and verdict bits (cold); rose: real numbers, which a block carries from loop to loop. \textbf{a}, The
network keeps one vector of 128 numbers for every ordered pair of items, its pair state. In the general
setting the input's items, facts and numbers enter it once, through a learned embedding. One block then reads the state
and writes back to it, with the same weights in every loop. It reads through two learned tests per channel, $A(i, k)$ and
$B(k, j)$. Half of its 128 channels count, for every pair, the middle items $k$ that pass both tests (the
witnesses) and hand on only two verdict bits per pair of channels: is one count larger, and are the two equal? The other
half keep the strongest witness and hand on its strength, a number between 0 and 1. A gated mix of the bits and strengths
is added to each cell, then the output of a small network applied to each pair. After the last loop a read-out takes the
answer from the pair state: here, for shortest routes, each node points to the one before it. The block has
146,552 parameters, shared by every loop, and one loop costs $n^3$ per channel. \textbf{b}, The four kinds of
block in one grammar: the pair state (small square; rose tints, real numbers; cobalt cells, bits), the block (box,
naming the channels it holds: count, the count-and-compare channels of a; best, the best-witness channels; or the general
and the cold half), what one loop hands to the next (the arrow back; rose, a carried real state; cobalt, bits from which
the state is rebuilt), and where numbers enter (into the state through an embedding, rose; into the block as comparisons
only, cobalt; from a frozen language model, grey box). General: the general block of a, 97.04\% on CLRS-30 with one block per task and
97.20\% with one block for all thirty tasks. Cold: the cold block, into which numbers enter only as comparisons made inside the layer, its state rebuilt from bits
every loop (kept bits, which only turn on, or free bits; with kept bits it rebuilds twice, and its comparison channels
read a state without the bits their learned tests made); right on every test input, on every seed, on 9
of the 12 tasks whose answer depends on the numbers only through comparisons. General and cold halves: the
general-and-cold block, one block for all thirty tasks, 96.55\% at its registered read (weight-averaged
parameters, $K(n)$ loops) and 96.40\% read like the general block (final parameters, eight loops), a loss of
0.46 points against our first general block for all thirty read the same way. Beside a language model: three kept bits per
pair of mentions, monotone tests, tied legs and a read-out that can only say yes, reading relations from a frozen model;
exact to 1,000 statements. Schematic, not data: everything in b (the icons and their cell patterns are drawn,
not read from a model); in a, the tests drawn in the block are set by hand, as in Extended Data
Fig.~\ref{edfig:arch}, and the tints of the pair state and of the bit and strength columns are illustrative.}
\label{fig:arch}
\end{figure}

The layer holds a vector $z_{ij}$ for every ordered pair of items (Fig.~\ref{fig:arch}a), like the cell of the pair table in Fig.~\ref{fig:big}d,
and the witnesses of a pair $(i, j)$ are the middle items $k$. A middle item is admitted when two learned tests pass, one
on the leg $(i, k)$ and one on the leg $(k, j)$, just as a middle town counts when a road leads to it from the start and
another from it to the goal. With $t = \mathrm{norm}(z)$, $A_\ell(i, k) = [a_\ell \cdot t_{ik} + \alpha_\ell > 0]$ and
$B_\ell(k, j) = [c_\ell \cdot t_{kj} + \gamma_\ell > 0]$, so that channel $\ell$ counts
\begin{equation}
N_\ell(i, j) = \sum_k A_\ell(i, k)\, B_\ell(k, j),
\label{eq:count}
\end{equation}
a whole number at every size. Channels come in pairs, and each pair passes on two bits, $[N_{2m-1} > N_{2m}]$ (``more'')
and $[N_{2m-1} = N_{2m}]$ (``same''), so that what the layer passes on is a comparison, never a score that could drift.
Where the witnesses carry numbers of their own, as keys and positions do, a channel keeps the best value among the
admitted witnesses together with the number of witnesses that reach it, and compares these pairs through the same two
bits. Route sums, which the cold block builds on six graph tasks (below), carry no tie count: their two bits, ``less''
and ``level'', compare two routes with the same two ends by value alone. In the general
setting a second half of the channels keeps the strength of the best witness,
\begin{equation}
S_\ell(i, j) = \exp\Big(-\min_k \big[\max(\gamma^a_{ik}, \gamma^c_{kj}) + \tau_\ell \min(\gamma^a_{ik}, \gamma^c_{kj})\big]\Big),
\label{eq:best}
\end{equation}
with leg costs $\gamma = -\log\sigma(\cdot)$ and a learned $\tau_\ell \in [0, 1]$. Each loop writes these outputs back into
the state through a gate and a small network applied to each pair. In the cold setting the loop instead, as in the Markov reasoners of Bohde et al.\cite{bohde2024markov} but without
their hints, rebuilds the
state from the input and the bits alone, so that nothing continuous survives from one loop to the next. It rebuilds
twice per loop. The best-witness half, which here passes on only whether some middle item passes both tests, reads the
state rebuilt from all the bits, and its bits, once on, stay on (kept bits). The comparison half then reads a state
rebuilt from the kept bits and the comparison bits that no learned test made, such as those comparing a pair's own two
numbers; the bits of its learned tests are zero in this state only, and still reach the read-out and the next loop's
best-witness half. Proposition~3 needs them withheld; the second rebuild costs one more pass of the per-pair network,
$O(n^2)$ beside the joins. In the first loop every kept bit and every learned comparison bit is zero. One loop costs
$O(n^3)$ per channel.

The two settings differ only in how numbers enter the block. In the general setting they enter the state as
embeddings, and one configuration of 146,552 parameters, the same on all thirty tasks, is trained per task. In
the cold setting they never enter the state at all: they are kept exactly in registers, on items or on pairs, and the
state receives only
the bits that compare them inside the layer, which we count as part of the method. We call the block of the first setting
the general block, which embeds its numbers and carries a real-valued state from loop to loop, and the block of the second
the cold block, which passes on only comparisons and rebuilds its state from bits; a general-and-cold block, used for
the shared runs of Section~6, has one half of each kind, each a whole block, and writes the cold half's bits into the general half's state at every
loop (Fig.~\ref{fig:arch}b; Supplementary Note~2). We keep the word warm for a read at a finite temperature, and for
the one block that is trained and read that way, the soft witness sum of an ablation (Fig.~\ref{fig:committee}). On six weighted
graph tasks the cold block also builds route sums by a fixed rule, $R(u, v) \leftarrow \min(R(u, v), \min_k R(u, k) +
W(k, v))$ over the admitted middle items. Arithmetic built in this way is a ceiling rather than a result, so we report
those tasks apart from the learned ones. A few parts are set by task rather than learned: for each combination of
register types two channel pairs have fixed tests that admit only the two items of the pair, so they hand on how those
items' own numbers compare; depth-first search, topological sort and strong components loop on bits without keeping
any and are read after $n$ loops; task scheduling takes its deadlines in units of the input's size and its positions as
given; Prim's algorithm also builds hop counts along the tree; and Floyd--Warshall's read-out sees which admitted middle
item wins the join. Supplementary Table~2 and Supplementary Note~2 give what each setting builds in, task by task.

We train from answers alone, with no intermediate states, and use the hard tests in the forward pass from the first
update. Their gradient is the slope of a sigmoid, as in straight-through estimation\cite{bengio2013estimating}; ``more''
takes the slope of $\sigma(N_{2m-1} - N_{2m})$, and ``same'' a bump that is zero at a tie and points towards it. The cold
block is read when its bits come to rest, at the first loop that changes nothing, or after $n$ loops on the tasks whose
answer must be built one step at a time.

\begin{figure}[tbp]
\centering
\includegraphics[width=\linewidth]{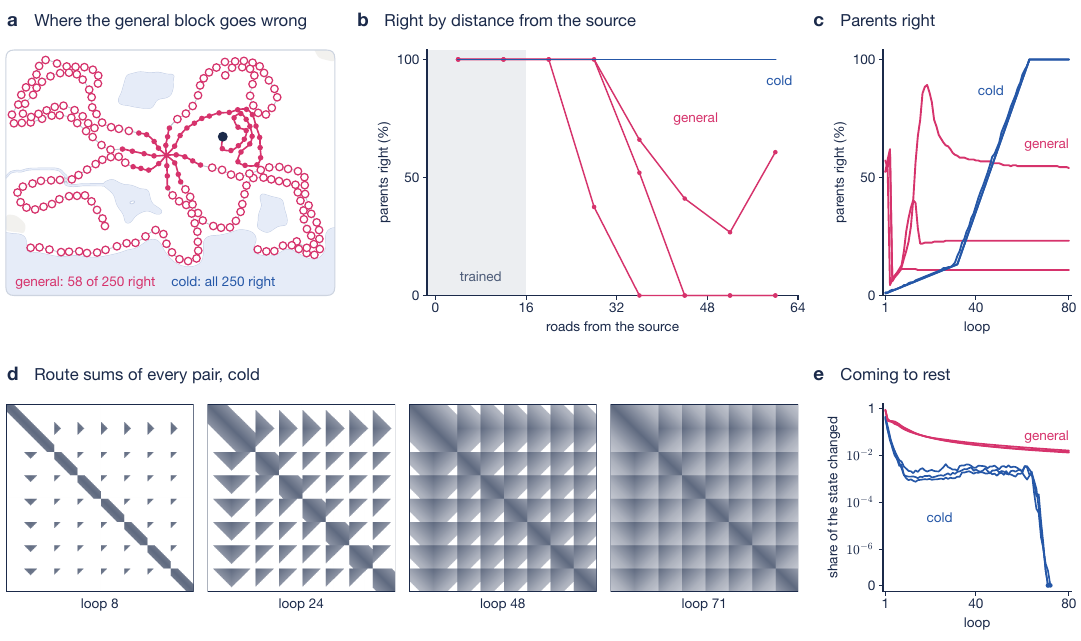}
\caption{\figlabel{Fig. \thefigure}{The cold layer on a map 63 roads deep.} The 250-town
map of a graph-search question, given to saved Bellman--Ford blocks as an input (every road of weight
0.5, roads in either direction, the source at one end of the map's longest route, 63
roads); nothing is trained. Cold: the cold block, which rebuilds its bits every loop and builds its route sums by the
fixed rule over admitted middle items. General: the general block, which carries a continuous state from loop to loop.
\textbf{a}, The general block (seed 0) read at loop 71, where the cold block (seed 0) comes to rest: a town
filled where it names the town's parent rightly (the road to it drawn), a ring where it names it wrong. It names
58 of 250 parents right and none of the 168 towns 40 or
more roads deep; the cold block names all 250, so its map, filled everywhere, is not drawn. \textbf{b},
Parents right at loop 71 by the number of roads from the source, in bins of eight roads, three seeds of each
block; every cold seed is right in every bin. Shaded, the depths that a training graph can reach (at most
15 roads). \textbf{c}, Parents right after each loop, three seeds of each block. The general blocks
peak at 53.6, 62.0 and 89.2\% (loops
2, 3 and 19) and end at
23.2, 10.8 and 54.0\% at loop 80; every cold
block names every parent from loop 63. \textbf{d}, The cold block's route sums for every pair of towns
after loops 8, 24, 48 and 71 (a cell filled once its route is known, tinted by its length; towns in
depth-first order of the shortest-path tree, so that each branch is a run of rows). \textbf{e}, The share of each block's
state that changes in each loop (a loop that changes nothing is drawn at 0): the cold block's bits and route sums stop
changing exactly after 71, 72 and 71 loops, which is where its
registered read stops; the general block's state still changes by 1.4 to 1.7\% per
loop at loop 80. On the 49-town map of Fig.~\ref{fig:big} both blocks name every
parent, the general blocks from loop 3 and the cold blocks from loop 6; the
cold blocks rest after 19, 16 and 18 loops, and the general block's
state still changes by about 8\% per loop at loop 16. Schematic, not data: the
map's geography (the course of each road and the towns' positions are a drawing of the question's edges) and the order of
the towns in the tables.}
\label{fig:layer}
\end{figure}

Figure~\ref{fig:layer} shows saved Bellman--Ford blocks, trained on small graphs, at work on a map of
250 towns whose farthest town lies 63 roads from the source, with 168 towns
40 or more roads away. The cold block extends its routes by one road per loop and names every parent
from loop 63, the depth of the map, on each of its three seeds. Its bits and route sums then stop
changing exactly, after 71, 72 and 71 loops, which is where its
read stops, with every parent right. The general block, which carries a continuous state, never comes to rest: after
80 loops its state still changes by 1.4 to 1.7\% per loop,
and its share of parents right rises in the first loops and then falls back. Read at loop 71, its three
seeds name 58, 27 and 137 of the 250 parents right,
and of the 168 towns 40 or more roads deep, 0,
0 and 72. Every town that any of them names wrong lies at least
27 roads from the source, deeper than any answer in training (Fig.~\ref{fig:layer}b). These
trained blocks walk rather than double. The layer can double
its reach, since each loop can join two routes found by the one before, but a block learns to do so only where its
training rewards it, as on 2-SAT under a tight loop budget (Section~6) and in the blocks beside a language model
(Section~8).

The next section states what can be proven about this layer, and about single
blocks once they are trained.

\section{What can be proven}

The layer can be described exactly in four respects; its reach over yes-or-no questions of a whole class, the constraint
problems, follows a sharp line; and a single trained block can be proven correct on inputs of every size. We take these
in turn.

The first result is the limit already stated (Proposition~1): the cold reading keeps $(s^\ast, g^\ast)$ and orders two
sums as every small enough temperature does. The pairs also compose. Over a route of two legs, such as start $\to k$ and
$k \to$ goal on the map, the leading terms combine as $(s_1 + s_2, g_1 g_2)$, and across disjoint families of witnesses as
the better of the two pairs, with the counts added when the scores tie.

The second is invariance (Theorem~2). If the input's numbers enter the state only through comparison bits, then any
recoding of them that keeps every comparison, such as any increasing map of the keys, leaves every bit, every state, the
point of rest and the answer unchanged. Node potentials, $W'(u, v) = W(u, v) + \varphi(u) - \varphi(v)$, act in the same
way on the built route sums, because $R(u, k) + W(k, v)$ moves by $\varphi(u) - \varphi(v)$ whatever $k$ is, provided the
layer only ever compares two routes with the same two ends; the symmetry thus dictated how the comparisons are wired. We
tested the theorem on every cold model: across 41 models and 5,303,808 answers,
recoding the numbers changed 0. The general block, which embeds its numbers, changes up to
636 of 2,048 answers on Bellman--Ford under the same test.

The third is rest (Proposition~3). Without negative cycles, the cold block's kept bits only turn on and its built numbers
never rise and take finitely many values, so together they change only finitely often, $M$ times in all. A change
reaches the learned tests at most two loops later, so once nothing has changed for three loops the state repeats, and
the block comes to rest within $3M + 3$ loops. Reading at rest therefore needs no loop count fixed in advance.

The fourth is a limit on depth (Proposition~4). One loop is computable in NC, so a polylogarithmic number of loops
computes a function in NC. Since the order in which depth-first search visits the nodes is
P-complete\cite{reif1985depth,greenlaw1995limits}, no block looped polylogarithmically many times computes it on every
graph unless NC = P: depth-first search has to be built one step per loop, and its loops have to grow with the input.
Proofs of these four results are in Supplementary Note~1.

The fifth result draws the line around constraint problems (Theorem~5). A constraint problem asks whether variables can
be given values from a finite domain $D$ so that every constraint of an instance holds; 2-SAT and two-colouring are
examples, and so are parity equations over three variables. A class of such problems has bounded width when checking
small groups of variables for consistency is enough to decide it, and by Barto's theorem checking pairs of variables
through a third is then enough\cite{barto2016collapse}. For classes whose constraints relate two variables, the theorem
reads: a pair block that sees no positions decides the class on instances of every size if and only if the class has
bounded width. In one direction, if the class has bounded width, one looped cold block with fixed weights decides it on every instance,
by running Barto's check of pairs through a third item as synchronous loops. Since kept bits only turn on, they record
what the check has ruled out: for each pair of variables $(x, y)$, which of the $|D|^2$ pairs of values are excluded.
The check excludes $(a, b)$ on $(x, y)$ when, for some third variable $z$, every value $c$ of $z$ is excluded with $a$ on
$(x, z)$ or with $b$ on $(z, y)$. That condition is not one test on each leg, but it is an OR, over the subsets $U$ of
$D$, of one: every $c$ outside $U$ excluded on $(x, z)$, and every $c$ in $U$ excluded on $(z, y)$. So $|D|^2 2^{|D|}$
channels run the check, and the block comes to rest within $n^2 |D|^2 + 2$ loops. In the other direction, if the class
does not have bounded width, no network whose answer depends only on what the folklore two-dimensional
Weisfeiler--Leman refinement of the instance sees decides it, whatever its weights, its join (exists, best or count) or
its number of loops, and even with a different network for each size. This half rests on the counting-logic lower
bound of Atserias, Bulatov and Dawar: they build, as Cai, F\"urer and Immerman did for graphs, two systems of equations
over a finite abelian group, one solvable and one not, that counting logic cannot tell apart, and carry the pair to
every class without bounded width through gadget reductions\cite{cai1992optimal,atserias2009affine,barto2014local}. For constraints on three or more variables the same holds once each
constraint has a token of its own. The theorems are cited; the statement for networks, the construction and the learned
results are ours (Supplementary Note~6). 2-SAT and two-colouring have bounded width, and for them the loops needed are
few: a contradiction $L$ implications deep is found within $\lceil \log_2 L \rceil$ loops, because each loop composes two
chains of implications and so squares the implication graph, which matches the one loop per doubling of depth that we measure (Section~6). Parity equations
over three variables do not have bounded width\cite{atserias2009affine}, so no position-free pair block decides them,
whatever it is trained on. With unbounded precision a looped network reaches the full message-passing
limit\cite{rosenbluth2026repetition}; which class the cold layer computes at fixed precision, with its best witness and
tie count, remains open.

The last result concerns a trained block rather than the layer. In the blocks we trained beside language models
(Section~8), a test adds a number looked up from the pair's input pattern to non-negative weights on
the kept bits that are on, so each trained block is a finite table, and two finite computations on that table decide
whether it answers right at every size (Supplementary Note~5). For the answer no, every context whose answer is no maps
onto one small template, whose items are of two kinds, the descendants of the question's first person and the rest; the
answer no is what puts the question's second person among the rest. If
the block says no on the template, it says no on every such context, because by induction on the loops every bit kept
by a pair is also kept by the pair's image in the template. For the answer yes, the chain of statements that makes the
answer yes carries the same patterns inside any context without cycles as it does alone, since a pair's pattern depends
only on its own two statements and each further statement brings mentions of its own, and further items can only add
bits. Both checks assume that each statement and the question name two different people. If some set of kept bits is rebuilt whenever a pair of the chain is split at its middle, a chain of $L$ statements
says yes within $t_0 + \lceil \log_2 L \rceil + 1$ loops, for every $L$. Discrete reasoners trained with hints have been
checked transition by transition for every graph size\cite{rodionov2025discrete}; rule learners trained from answers
reach every size through the program extracted from them, not through the trained network\cite{evans2018learning};
computer-assisted proofs about a trained transformer have fixed its input size\cite{gross2024compact}; a recurrent
adder trained with a hard forward pass has been proven in Lean for every length, with its carry supervised after every
step, while a variant supervised only after the last step was not proven\cite{tealsea2026lean} (a repository that is not
peer reviewed and that states its proofs were written and reviewed by AI agents); and in earlier work we proved a
language model's arithmetic exact at every length by checking the finite set of bounded views from which each of its
steps is computed\cite{mohsenvand2026sketchpad}. Here the network is trained from yes-or-no answers alone, on relational
input whose loops must grow with its size, and the proof concerns its own table, read exhaustively from its weights.

\section{Results}

We test the layer in four settings, each harder to read in a different way: a task chosen because softmax fails on it,
the thirty algorithms of CLRS-30, graphs deeper than any in the official test, and constraint problems decided from
yes-or-no answers alone.

\subsection{Maximum retrieval}

We begin where softmax attention was shown to fail\cite{velickovic2025softmax}. Each set holds items with a class and a
priority, and the answer is the class of the item of highest priority. In our layer, each item's witnesses are the rivals
with higher priority, compared inside the layer, and the cold count picks the item that has none. Trained on sets of 5 to
16 items, it is right on at least 99.95\% of sets at every size up to 16,384 on every one of
ten seeds (99.97\% on average at the hardest size), and every error it makes falls on
a set whose two highest priorities are equal in floating point (Fig.~\ref{fig:fail}b). The published softmax network is
right on 22.6\% at 4,096 items\cite{velickovic2025softmax}, and the best published method on
62.7\%\cite{vasylenko2026long}. Our own replication of the softmax network, which falls below the
published curve (Supplementary Note~4), is right on 17.6\% there, and on 36.3\% when its
weights are read cold, because its learned offsets survive the cold read (Section~3). A network that compares priorities
inside the layer but is trained and read with a soft count is right on 42\% at 4,096 items, and read
cold, the same weights are right on at least 99.96\% at every size on every seed. Each remedy alone
fails; together they are exact.

\subsection{Thirty algorithms without hints}

\begin{figure}[tbp]
\centering
\includegraphics[width=\linewidth]{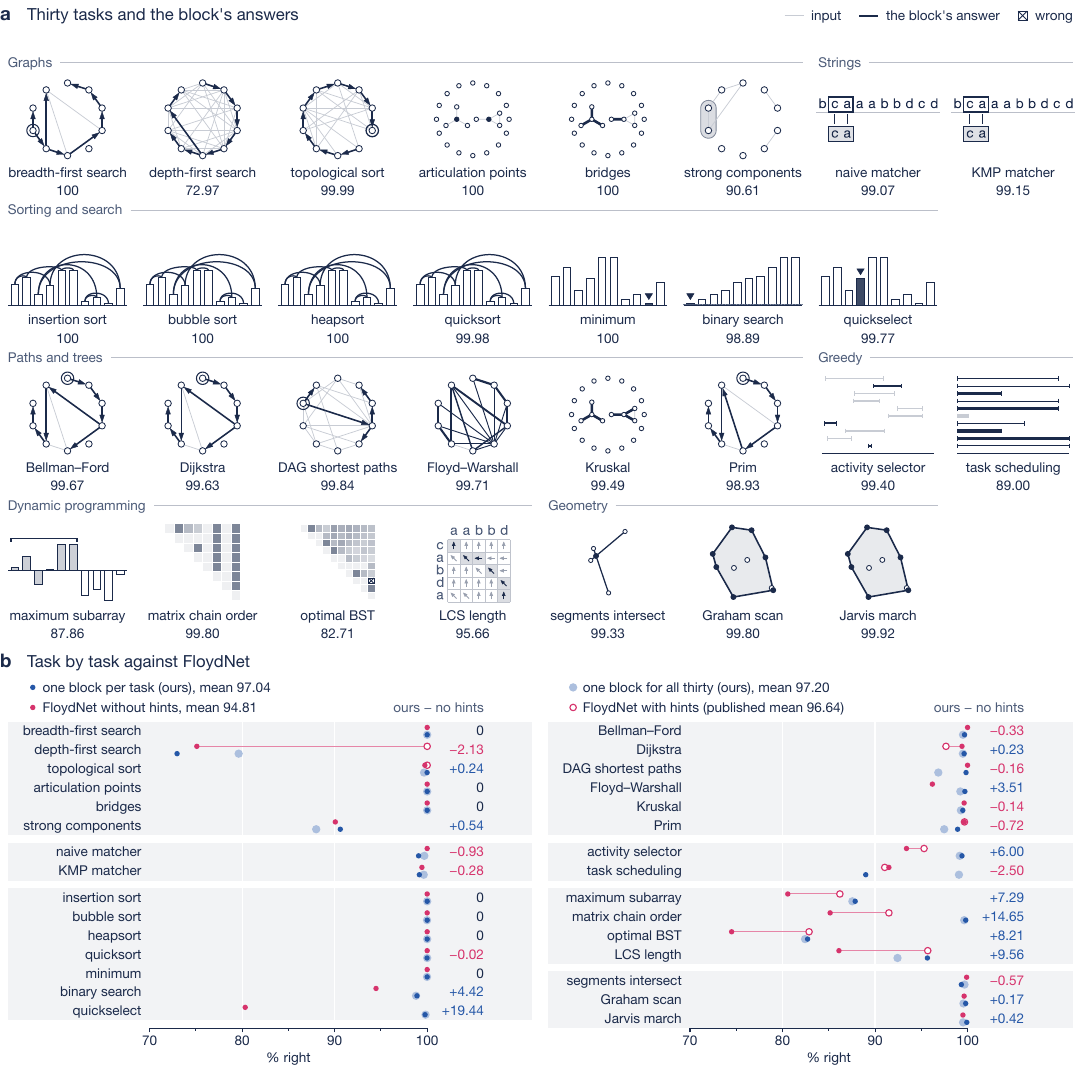}
\caption{\figlabel{Fig. \thefigure}{Thirty algorithms without hints.} \textbf{a}, Each task drawn on a small input from the
benchmark's generator (ten items; articulation points, bridges and Kruskal at 24 nodes, the smallest size whose first
input has a cycle), with the answer of one block per task (one of its three seeds) drawn in navy over the input in grey.
The one wrong answer on these inputs, a single cell of the optimal search tree's table, is crossed. Under each tile, the
share right on the official 64-item test, mean of three seeds. \textbf{b}, Task by task on the official test, in the
families of a. In each row the upper lane is FloydNet (rose): its runs without hints (filled) and, where its published
score comes from a run with hints, that score (open, joined by a hairline). The lower lane is ours (cobalt): one block per
task (filled) and one block for all thirty tasks (pale disc; mean 97.20\%, depth-first search
read after $n$ loops). The legend gives each series' mean over the thirty tasks; FloydNet's published mean mixes its runs
with and without hints. Right: ours, one block per task, minus FloydNet without hints; cobalt where ours is ahead, rose
where it is behind. All of our models are trained without hints. Schematic, not data: the layouts of the tiles, including
the order of the nodes around each ring (chosen to minimise crossings), the arcs of the sorting tiles, the tints of the
split tables (where the chosen split falls in its range) and the placement of the 24-node sparse graph.}
\label{fig:clrs}
\end{figure}

On CLRS-30, trained without hints, our layer beats the best published model even when that model uses hints. In the
general setting, one block per task averages 97.04\% at 64 nodes over three seeds (Fig.~\ref{fig:clrs}), while
FloydNet reports 96.64\%, with 10 of its thirty scores from runs with hints, and averages
94.81\% in its runs without them\cite{yu2026floydnet}. Against those runs without hints, which are trained as ours
are, our block leads by 2.23 points, winning 13 tasks, tying 7 and losing
10; against every other published model trained without hints it is ahead on every task each one
covers (Supplementary Note~4). Against FloydNet's published scores it wins 10, ties 7 and loses
13, and only 2 of those losses exceed a point: depth-first search (72.97\% against
100.00\%) and task scheduling (89.00\% against 91.05\%). One block trained on all
thirty tasks at once does better than one block per task: it averages 97.20\% over three seeds (97.01 to
97.34), every seed above FloydNet's published average, and 2.39 points above FloydNet's runs without
hints. It is our first shared block (96.87\%) with two changes. Task scheduling's deadlines reach the block
divided by the input's size, the units its positions already have, which lifts that task from 91.85 to
99.07\%; this is an encoding of the input that the benchmark does not prescribe. Depth-first search is read after
$n$ loops, 79.62\% against 73.03\% at 8 loops, with the seeds far apart (76.42,
91.02 and 71.44\%). Read like the first shared block, every task after 8 loops, it averages
96.92\%, and it loses up to 2.10 points on other tasks (Supplementary Note~4).

The cold setting does better where the theory says it should. Choosing for each task the setting our theory prescribes,
a cold or bit-loop block on the 12 tasks whose answers depend on the numbers only through comparisons
and the general block on the other 18, with no arithmetic built into the block, gives a composite of
97.98\% when every task is trained on the benchmark's standard sampler and inputs,
1.34 points above FloydNet's published average, with 9 wins,
10 ties and 11 losses. Two cold rows were trained off the standard inputs: task
scheduling, with deadlines given in units of the input's size and positions not randomised, and quickselect, with eight
extra training sizes and a loss from the second loop. No standard run of their cold setting has three seeds, so the
standard column takes the general block on both; with the two rows as trained, the composite is 98.35\%
(10 wins, 10 ties and 10 losses). Some choices were made
after seeing results, for depth-first search, strong components, topological sort and task scheduling, and with every
choice as first registered the average is 94.77\% (Supplementary Table~2). Where the answer depends
on the numbers only through comparisons and no route sum is built, the cold block is right on every test input, on every
seed, on 9 of 12 tasks; the exceptions are Strong components, Activity selector, Task scheduling.

\subsection{Depth beyond the official test}

The official test cannot say whether a rule keeps doubling its reach, as a loop over two-step routes can.
On Bellman--Ford its answers are 4 to 7 hops deep, fewer than the
13 loops at which it is read, so a block that stops at a fixed depth and one that doubles score the same.
We therefore read the shared blocks trained on eleven of the hardest tasks on deeper graphs, of 128 and 256 nodes with
answers up to 244 hops deep, and on chain-shaped graphs of 64 nodes (Fig.~\ref{fig:depth}; one seed
per block). The general block falls to 77.3\% on Bellman--Ford at 256 nodes, and a second half of the same
kind lifts it to 85.5\%; with a cold half in its place, the general-and-cold block holds
94.2\%, and 98.2\% when eight of the 64 best-witness channels of its cold half test both legs with one test,
so that they square their relation at every loop. On the chain graphs the same four blocks score 50.9,
64.9, 77.3 and 96.8\%. Prim's algorithm breaks the
pattern: at 256 nodes the general block with two halves (96.5\%) beats both general-and-cold blocks
(91.2\% and 88.3\%). Depth-first search, which must be built one step per
loop, is right on every graph at 64, 96 and 128 nodes once positions are compared rather than embedded, whereas the same
bit loop with positions embedded is exact at 64 nodes and not at 128 (Fig.~\ref{fig:depth}c).

\subsection{Constraint problems}

\begin{figure}[tbp]
\centering
\includegraphics[width=\linewidth]{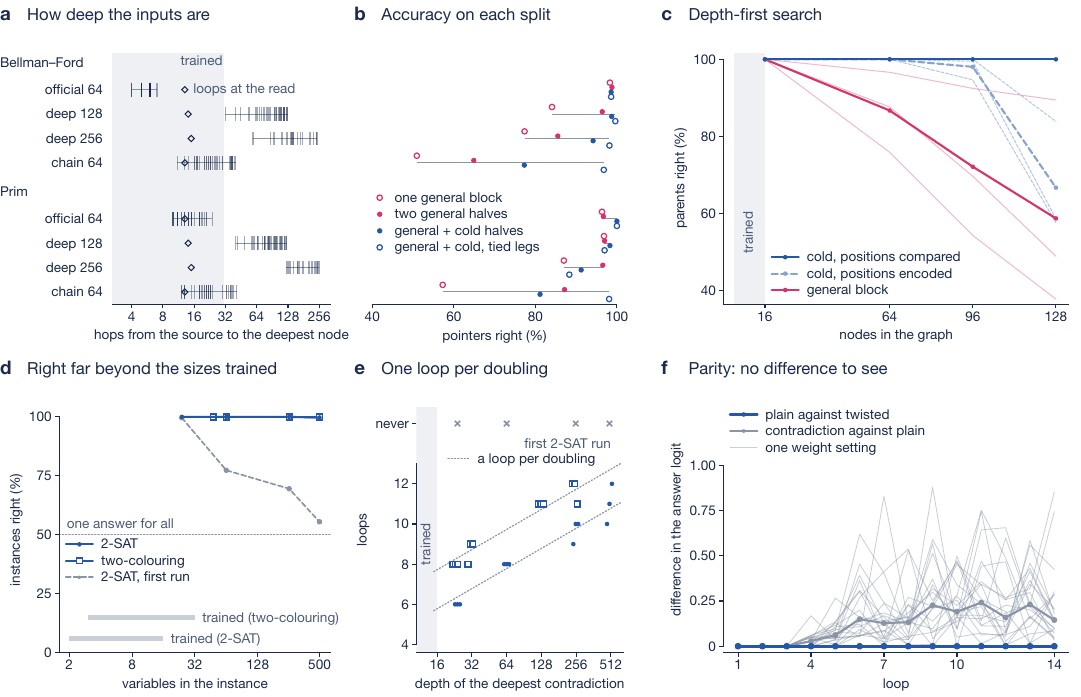}
\caption{\figlabel{Fig. \thefigure}{How far the reach goes.} \textbf{a}, How deep the answers are: one tick per graph at the
largest number of hops from the source along the true answer, for the official 64-node test and for deep graphs of 128 and
256 nodes and chain-shaped graphs of 64 nodes; diamonds, the loops at which each split is read; shaded, the most hops a
training graph can need. On the chain graphs about half the nodes (49\% for Bellman--Ford)
are not reached from the source; they point to themselves and are scored as the library scores them. \textbf{b}, Pointers
right on each split for four shared blocks trained on eleven tasks (one seed each), on the rows of a: one general block
and a general block of two halves (rose), a general-and-cold block, and a general-and-cold block in which eight best-witness channels of the cold half test
both legs of a pair with one test (tied legs; cobalt). \textbf{c}, Depth-first search: parents right against nodes in the graph, read after as many loops
as nodes, three seeds (thin) and their mean (bold), for the cold block with positions compared, the same bit loop with
positions encoded, and the general block. \textbf{d}, Constraint problems from yes-or-no answers alone: instances answered
correctly against the number of variables, three seeds per class (registered read: weight-averaged parameters,
$K(n) + 4$ loops); bars, the sizes trained. Half of every test split is unsatisfiable, so one answer for every instance
scores 50\%. One 2-SAT seed misses one instance at 500 variables. Grey: the first 2-SAT run, with a looser loop
budget and a higher learning rate (one seed). \textbf{e}, The first loop at which every instance of a split is right,
against the depth of its deepest contradiction (implications for 2-SAT, half the shortest odd cycle for two-colouring),
per seed (markers as in d); dotted, one loop per doubling of depth; crosses, the first 2-SAT run, never right on every
instance within the 17 loops read. \textbf{f}, Parity equations over three variables on the cube and the Petersen graph,
random weights and no training: the largest difference in the answer logit between plain and twisted instances (cobalt;
0.0 at every loop) and between a visibly contradictory instance and the plain one (grey; thin, each of 20 weight
settings; bold, their median). Shaded bands throughout: the sizes or depths seen in training. Schematic, not data: the
four sub-rows of each row in b and the side-by-side placing of seeds that share a value in e; every other mark is a
measurement. There is no general-block counterpart for d--f: every constraint-problem run is a cold block.}
\label{fig:depth}\label{fig:csp}
\end{figure}

The sharp line of Section~5 predicts which yes-or-no questions the layer can learn to decide, and we tested both of its
sides. We trained the layer on yes-or-no answers alone, asking only whether an instance can be satisfied. On 2-SAT,
trained on up to 16 variables, and on two-colouring, trained on up to 32, it is right
on every instance at 500 variables on three seeds of each, except for one 2-SAT instance on one seed
(99.22\%). At 500 variables the hardest 2-SAT instances hide a contradiction
256 to 511 implications deep, against at most 16 in training, and the loop at which
every instance is first right grows by about one loop per doubling of that depth, as the theory predicts:
6, 8, 9--10 and 10--12 loops at 24, 64,
256 and 500 variables (Fig.~\ref{fig:csp}d,e). The loop budget in training decides whether this happens. Trained with more
loops than it needed, the block found a rule that reaches contradictions of about 24 to 31 steps and no further. On the
other side of the line, parity equations over three variables, which no position-free pair block can decide, the trained
layer does not decide either: its answer is identical on satisfiable and unsatisfiable instances at every loop
(Fig.~\ref{fig:csp}f).

\section{Failures with named causes}

The results above leave a few large errors, and each has a cause that we can name and a fix that we tested. On strong
components, most of the block's wrong answers name the right component but the wrong one of its entry points, the nodes
that edges from other components point at, and small training graphs rarely contain components with several of them.
Training graphs rich in such components, a sample we changed after this diagnosis, lift the official score to
98.21\% over three seeds.

String matching fails for a different reason. It needs a count carried along the text, and the cold layer passes on
comparisons, not counts. Given the count as a fixed program, the two matchers are exact on 6 of
6 runs, which we report as evidence for the cause rather than as part of the method. The optimal
search tree needs a two-line recurrence over intervals that gradient descent does not find from answers. Listing every
program of a small grammar of joins finds it from 200 training inputs, and the program found is right on
every official test pair; because the grammar was written knowing the answer, this shows only that what is missing is the
search for the program, not the join that would run it. Segment intersection is the one failure without a fix. The
general block reaches 99.33\% on it, but the cold block, which only compares numbers, would need a
product of two coordinates, and given that product as a pair number it stayed near chance (44.30\%).

\begin{figure}[tbp]
\centering
\includegraphics[width=\linewidth]{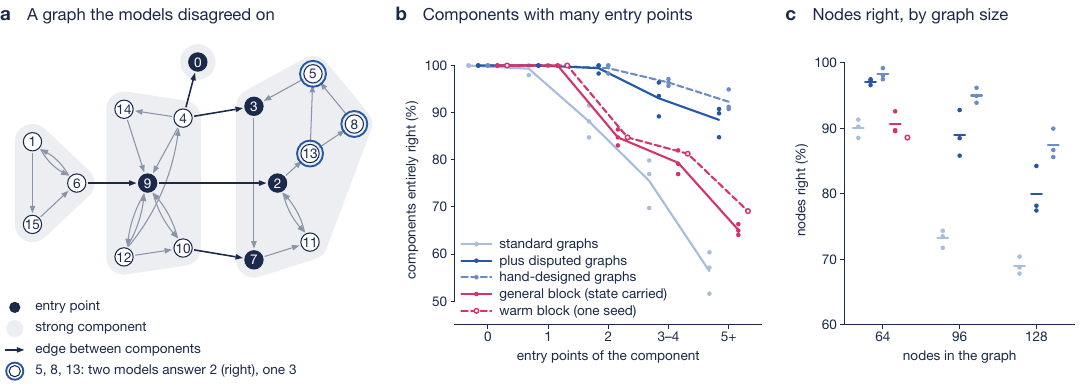}
\caption{\figlabel{Fig. \thefigure}{The models find their own missing training cases.} \textbf{a}, A 16-node graph after
14 random edge changes, on which three saved models first disagree: its four strong components (shaded), their entry
points (filled; an edge from another component points at each) and the edges between components. For nodes 5, 8 and 13
two models name node 2, the right answer, and the third names node 3, another entry point of the same component. The
graph comes from the committee's rule rerun on the saved models, not from the training pool itself; about a quarter of all
disputed nodes are of this kind. \textbf{b}, Components answered entirely right, by their number of entry points, pooled
over the official and a fresh 64-node test file (640 components): cold blocks trained on three pools (the
standard graphs, the standard graphs plus those the models disagreed on, and graphs designed by hand to be rich in entry
points; cobalt tones), the general block, which carries a continuous state from loop to loop (rose; the CLRS-30 headline
series, read after 8 loops), and the same block without its count channels, a fully warm block (rose, dashed; one seed);
three seeds each (dots) and their mean (line). With five or more entry points the general block gets
65\% of components entirely right, the warm block 69\% and the standard cold pool
56\%; the committee's pool lifts the cold block to 88\% and the hand-designed pool to
92\%. \textbf{c}, Nodes right per seed (dots) and mean (bar) on the official 64-node test and on 16 fresh
graphs of 96 and 128 nodes, read after as many loops as nodes; colours as in b. The general and warm blocks were read only
at 64. At 96 and 128 nodes the hand-designed pool was read on a CPU and the other two on a GPU, which can differ by up to
3 points on the same model. Schematic, not data: the node positions in a (components as groups, ordered left to right by
the edges between them); edges, components, entry points, rings and answers are data.}
\label{fig:committee}
\end{figure}

The first of these fixes needed a diagnosis, and a learner of algorithms should not depend on one. We therefore let a
committee of saved models find the missing cases itself. Small training graphs are changed one edge at a time, those on
which the saved models disagree are kept, and they join the training pool. One round lifts the official score on strong
components from 90.02\% to 97.12\% over three seeds (96.58 to 97.46),
against 98.21\% for the hand-designed pool, and components with five or more entry points come out entirely
right in 88.5\% of cases (Fig.~\ref{fig:committee}). The hand-designed pool keeps its lead as the graphs grow,
95.0 against 89.0\% at 96 nodes and 87.4 against 80.0\% at
128. The committee can find only errors that small graphs show, and the errors that remain at 64 nodes almost never
appear among the disputes of 16-node graphs, so we did not run a second round. With the layer's reach and its limits on
algorithms established, we turn to a setting where the witnesses are read from language.

\section{Inside a language model}

Language models read single sentences well and chain many of them poorly. Pairing a language model, which reads the
sentences, with a separate part that chains them is an established design. The separate part has been a logic
solver\cite{yang2023coupling}, a differentiable logic program\cite{zhang2023improved}, a path
search\cite{zhang2026extracting} or a rule table learned by inductive logic programming\cite{chen2026reason}, measured on
benchmarks such as CLUTRR\cite{sinha2019clutrr}. We ask whether the cold block, learned from yes-or-no answers, can serve
as that part, how far it chains, and whether it can be proven exact. Its witnesses are now people, and a pair of people
keeps, as a pair of towns does on the map, the facts that link them through a third.

\subsection{Chaining a thousand statements}

Each context states, one sentence per statement, who is whose parent, in eight wordings such as ``Maria
Lopez is the mother of Tom Baker.'' and ``Tom Baker's mother is Maria Lopez.'', and the question asks whether one person
is an ancestor of another. Training uses contexts of 1 to 16 statements in six of the wordings, while the
test contexts, of 16 to 1,000 statements, use all eight, so that two wordings are
new even though their words are not. The block takes two things from the frozen model. Which mentions name the same
person comes from the model's input embeddings of the two names, before any of its layers, so this bit amounts to
matching names; a threshold set without labels decides it. Which way a statement points comes from the model's own
reading of that sentence alone, two extra passes per sentence, pooled by majority over every sentence of the same
wording in the text being read; wordings are grouped without labels by the words that are not names. The block receives,
for each statement, the model's reading and which name comes first, never the wording, and learns from answers which way
each reading points. The two unseen wordings therefore test the model's reading and the pooling, not the block, and the
pooling relies on the templates: sentence by sentence, the 1.5B model reads the direction of ``a parent of'' right only
57.0\% of the time. The model alone, answering by itself, has neither the matched names nor the
pooling. The block itself keeps three bits per pair of mentions; it is constrained so that it can only add facts, and it
can say yes only when a bit is on.

On frozen Qwen2.5-7B the block is right on every test question at every size up to 1,000 statements, on
six seeds, and on Qwen2.5-1.5B on three of three (Fig.~\ref{fig:organ}). The 7B
model reads 24,190 of 24,192 test sentences correctly, yet on its own it answers at chance
from 64 statements on, and at 43.75\% at 1,000. Used sentence by sentence rather than pooled,
its readings give 98.96\% at 1,000, and the only errors are chains that pass through a sentence
the model misread. Every start learned (15 of 15) when the direction came
from the model; when the block instead had to learn each wording's direction from answers, over a smaller model's reads,
5 of 10 starts learned, and each one that did was exact; there every
wording occurred in training. On a simpler
task, where a general block, which carries a continuous state from loop to loop, was trained on the same reads, the
general block falls to 71\% at 256 facts while the cold block stays exact (Fig.~\ref{fig:organ}c).

\subsection{Proofs at every size}

Measurements at sizes up to a thousand do not show that a block is right at every size, but the checks of Section~5 do.
Of the 23 runs that learned, on the three language models, every one is proven never to say yes on a
context whose answer is no, and to answer every question right at every size, within the loops at which it is read, on
contexts without a cycle (no family has one) and questions about two different people; for
nine of them the proof is also checked in Lean (Fig.~\ref{fig:organ}f). We widened the argument for chains three
times after seeing which blocks it missed; each widening only adds ways to prove, and the direct runs of the chains did
not change (Supplementary Note~5). The checks also reject all five starts that did not learn.
Three give a false yes and one never answers a chain. The fifth is a walker, which answers a chain
of $k$ statements at loop $2k$: run to rest it is right on every one of 3,000 random contexts, but at the
loops at which it is read it is wrong from chains of 4 statements, and the certificate rejects it on
the loop count, which a test run to rest cannot check (Supplementary Table~3). The proofs take the block's input bits as
given; how well the model reads them remains a measurement.

\subsection{CLUTRR}

On CLUTRR, the field's benchmark for this division of labour\cite{sinha2019clutrr}, the block is right on every story of
five to ten relations when given the stories' graphs, and from the stories' text it lifts a frozen open 7B model further
than any other system we found that uses no hand-written rules, apart from frontier reasoning models. CLUTRR's
crowd-written stories mix twenty kinship relations, and each asks for the relation between two people linked by a chain
of 2 to 10 stated relations; systems are trained on short chains and tested on all. The frozen Qwen2.5-7B-Instruct gives,
for each two people named near each other, the relation it reads or ``unknown'', and its probability. The block keeps one
bit per relation for each ordered pair and learns its rule table, which two relations give which, from the answers to
stories of two relations.

Given the stories' graphs in place of the model's reading, as the rule learners are, the block learns exactly the
benchmark's 62 rules in each of six runs and is right on all 826
test stories with 5 to 10 relations, the top of the published table, where the best system in the published comparison
scores 0.98 at ten\cite{cheng2023neural}. It misses only nine short stories, in
which the benchmark's own rules derive two different relations. The test stories took part in designing this trainer:
we scored four earlier builds of it on them before fixing the last one, because the validation stories have only the
training lengths and do not show the failure. The earlier builds learned extra rules and scored from
6.7\% to 98.8\% on the split trained on 2 to 4 relations (Supplementary Note~4).

\begin{figure}[tbp]
\centering
\includegraphics[width=\linewidth]{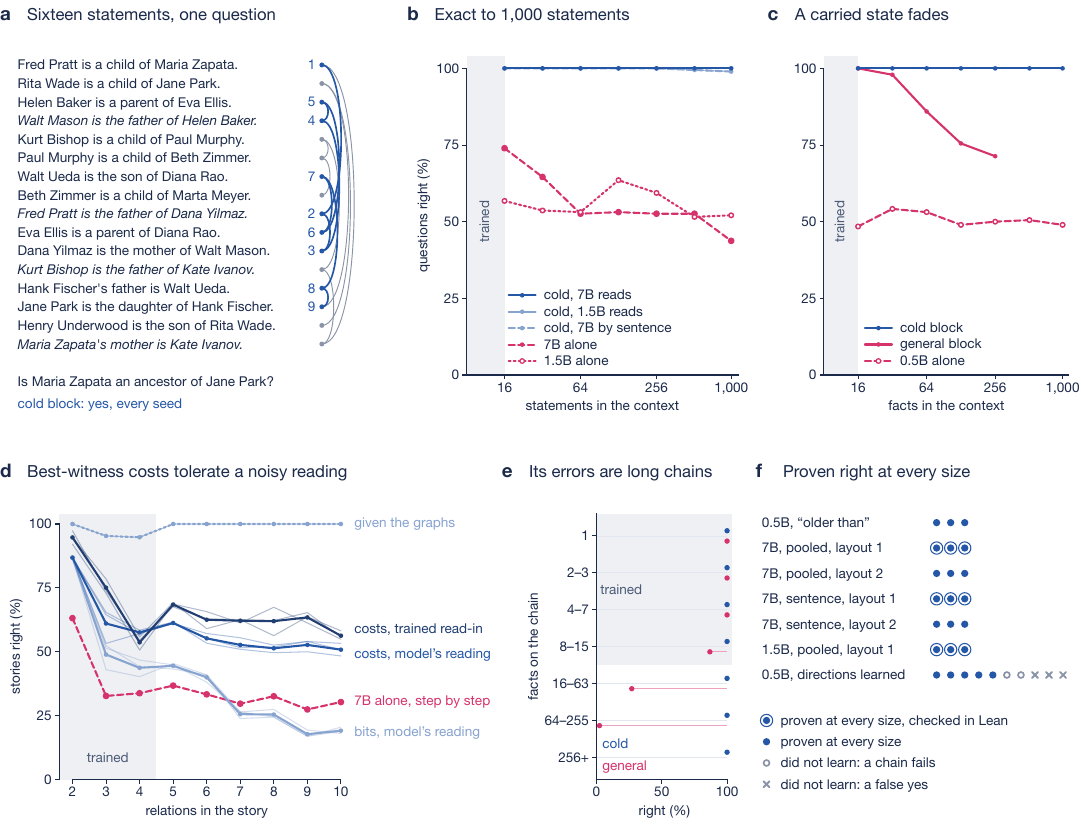}
\caption{\figlabel{Fig. \thefigure}{A cold block beside a frozen language model.} \textbf{a}, A test context of 16
statements and one of its questions, as given to the model. Italic: the two wordings that never occur in training. An arc
joins two statements that chain (the child of one is the parent of the other); the question's chain of nine statements is
in cobalt, numbered from the ancestor down. \textbf{b}, Questions answered correctly against the number of statements in
the context, for blocks trained on contexts of 1 to 16 statements in six wordings from yes-or-no answers alone (192
questions per size; 96 at 1,000). Cold block on the reads of Qwen2.5-7B (six seeds, two input layouts) and of
Qwen2.5-1.5B (three seeds). Sentence by sentence: the 7B model's reading of each sentence used directly, without pooling;
its only errors come from the sentences the model misreads (24,190 of 24,192 test sentences
read right). Rose: each model alone, answering ``Yes'' or ``No'' by itself. \textbf{c}, The first task, ``is $X$ older
than $Y$?'' over Qwen2.5-0.5B with single-token names: the general block, the same read-in and training with strengths carried
from loop to loop (one seed, read up to 256 facts), against the cold block (three seeds), and the model alone.
\textbf{d}, CLUTRR, trained on stories of 2 to 4 relations and tested on 2 to 10 (1,048 test stories): the
block on the frozen Qwen2.5-7B-Instruct's reading of each pair of names, with one bit per relation or with each read
relation costing $-\log p$ and the cheapest derivation answering; the same costs on a linear read-in of the model's state
trained from the answers alone; the same block given each story's graph; and the model alone, reasoning step by step
(rose). Lines are labelled at their right ends. Four earlier builds of the graph trainer were scored on the test stories.
\textbf{e}, The general block's errors on the first task are long chains: yes-questions right by the number of facts on the
question's chain, pooled over contexts of 16 to 256 facts (general, one seed) and 16 to 1,000 facts (cold, three seeds);
shaded, the chain lengths of training. \textbf{f}, Every start of panels b and c and of the kinship runs whose directions
were learned from answers (Supplementary Note~5, Table~3). Proven: two finite checks on the block's table show that it
never says yes when the answer is no and says yes in time on every chain, at every size, given its input bits; ringed: the
proof is also checked in Lean. The 1.5B networks share their tables with the 7B pooled networks of the first layout. Thin
lines: single seeds. The kinship sentences follow templates and state one relation; the held-out wordings reuse words seen
in training; each sentence's direction comes from two extra passes of the model, not from its internal states. Schematic,
not data: the heights of the arcs in a and the row order in f; every plotted point is a measurement.}
\label{fig:organ}
\end{figure}

From text, the bits fail. The model reads 88.6\% of the stated relations right, but it also names a
relation for 31\% of the pairs that a story does not relate, and a wrong bit spreads to many pairs
and cannot be outweighed: the block is right on 34.6\% of the test stories, close to the model alone
reasoning step by step (33.3\%). Read with scores in place of bits, the same block computes the
layer's own zero-temperature limit: each read relation scores the logarithm of the model's probability, scores add over
the two legs of a derivation, and each pair and relation keeps its best witness (Section~5). This read is part of the
method, and nothing in it is trained. It is right on 56.5\% when trained on 2 to 4 relations and
60.6\% when trained on 2 and 3, and on 51\% of the stories with ten relations, where
the model alone is right on 30\% (Fig.~\ref{fig:organ}d). DSR-LM, which fine-tunes a smaller
reader for the task, reports 60.98\% when trained on 2 and 3\cite{zhang2023improved}.

The reading can also be learned from the answers. A linear map of 75,285 weights takes the frozen
model's last-layer state, where the model is about to name what one person is to the other, to the twenty relations and
``nothing stated''. It is first fitted to the model's own reading and then trained from the stories' answers alone,
through the cost read made differentiable, with the rule table held fixed; no fact is ever labelled and the model is
unchanged. The cost read is then right on 64.6\% and 66.6\% of the test stories,
8.1 and 6.1 points above the same read of the model's own reading, and above
DSR-LM's published score on three seeds of three. The organ learns a reading that serves its rule
table, not a more faithful one: the trained map names only 76.5 to 79.1\% of the
stated relations right, against 88.6\% for the model's own reading, and uses relation names that no
story states as codes of its own, which the rule table composes into the right answer, a reasoning
shortcut\cite{marconato2023not}. Before training, the same read-in, fitted to the model's top answer, is right on only
45.8\% and 48.9\%: it is over-sure, so about half of its answers are ties that
fall back to the bits, and it is a weak baseline rather than a measure of what the training adds.

On the test file of our first split, the block with the trained read-in lifts the frozen 7B model from
33.3\%, reasoning step by step, to 64.6\%, 31
points, with no hand-written rule. On the same file, Path-of-Thoughts reports Llama-3-70B, ten times its size, at
56.3\% and GPT-4o at 57.6 to 61.9\% without hand-written rules, GPT-4o at
67.7\% when its reading feeds rules written by domain experts to an answer-set solver, and GPT-5 at
73.2\% asked directly\cite{zhang2026extracting}; on the second split, DSR-LM reports 60.98\% against our
66.6\%. On its own sample of stories, a method that learns composition rules by inductive logic programming
from a model's reasoning traces, and repairs the steps that break them, is right on 62\% of stories
with 8 to 10 relations with a 4-billion-parameter model, close to ours at those lengths, and on 75\%
with GPT-5.4\cite{chen2026reason}; its rule table is checked against the benchmark's own rules and completed from them, whereas
ours is learned by the block from answers alone and keeps the reader's confidence. We claim no new best on CLUTRR: the
block is above every system we found reported on this test without hand-written rules, apart from frontier reasoning
models, and below hand-written rules on GPT-4o and below the frontier reasoning models. One model family, one prompt and
three seeds of the block stand behind each of our rows. The errors come from the reading: given the graphs, the same
block is right on 1,039 of 1,048 test stories. The certificate does not cover the
reads from text, since their answer is a cheapest derivation rather than a kept bit and the trained read-in's relations
are codes of its own. CLUTRR also stops at ten relations, so length is measured only on the template statements, which
state one relation and chain over several hundred statements.

\section{Discussion}

A network can learn an algorithm from answers alone and stay exact far beyond the sizes it was trained on, provided its
attention is read at zero temperature, makes its comparisons inside the layer and rebuilds its state at every loop. Our
best CLRS-30 average comes from a block that takes only part of this recipe, the general block, which reads half of its
channels at zero temperature but embeds the input's numbers and carries its state; where that carried state drifts, as on
depth-first search, it is the cold block, with all three choices, that stays exact. Each
of these choices removes one of the three ways in which soft attention fails at size, and each failure can be seen on its
own: on SET, where a fixed temperature lets the one right pair be outweighed as the table grows; on maximum retrieval,
where learned offsets too small to matter in training decide the answer among thousands of items; and on depth-first
search, where a carried margin erodes with depth. The zero-temperature limit keeps the best witness and its tie count, and
comparing these pairs is what a warm sum does once its temperature is low enough, so a layer that computes the limit
directly never has to tune a temperature to the size of its input. The pair of towns on the map that reports its best
route and the number of routes that tie with it, rather than an average length that no route has, is the whole idea in
small.

The layer pays for this in two ways. It keeps a state for every pair of items and does work cubic in their number at
every loop, and tasks whose answer must be built one step at a time, as depth-first search's must, need loops that grow
with the input, which no layer of this kind can avoid unless NC = P. Some arithmetic also remains built in: on six
weighted graph tasks the cold block's route sums follow a fixed rule, and we treat those results as a ceiling. Learning
that arithmetic, for example as numbers held in bits with a learned carry, is open, as is the class of functions that the
layer computes at fixed precision with its tie count.

The errors that remain show what a learner of exact algorithms still needs. Its training sample must contain the cases
that pin the rule down, and a committee of the learner's own models can find many of them; it needs a store for counts
where the task requires one; and it needs a way to search for programs that gradient descent does not reach. Outside
CLRS-30 the same layer learns constraint problems of bounded width from yes-or-no answers and fails, as the sharp line of
Section~5 says it must, on those it cannot express. Beside a frozen language model it chains a thousand statements that
the model reads well and cannot chain itself, and two finite checks prove every trained block that learned right at every
size, given what the model reads. Where the reading is noisy, as on CLUTRR, the block's best witness tolerates the noise
and its bits do not, and what limits the result there is the reading, not the chaining.

\section*{Methods}

\paragraph{Benchmark protocol.} Inputs, answers and scores come from the benchmark's own library (dm-clrs
2.0.3)\cite{velickovic2022clrs}. Training draws from a fixed pool of 80,000 graphs or arrays at each of the sizes 4, 7, 11, 13 and 16, generated by
the library's sampler with its training settings (cold-setting quickselect also trains on the eight sizes between, drawn
by the same sampler); validation uses its 16-item split and testing its official 64-item split, the same inputs as the
published test file. We also score every model on a second, freshly drawn 64-item test file of 128 inputs per task
(library test sampler, seed 20,261,109) and the cold path and tree models on 32 fresh graphs of 128 nodes. No model sees intermediate
steps of the algorithm (hints): each is trained on the final answer alone. Scores are the library's: the share of
correct pointers, marks or classes over all test nodes, edges or arrays, as the library's evaluation function computes
it. Published scores are taken from each paper's own tables; FloydNet's\cite{yu2026floydnet} per-task score is its
headline Table~9, which takes each task's score from a run with or without hints, and its no-hint score is its Table~10. Every training run, its stopping
rule and its forecasts were written down before the run; Supplementary Table~1 lists every forecast and whether it held.
The configuration reported here is the last of a sequence of designs, each scored on the official test inputs of the
tasks it ran; it was then fixed and run once on all thirty tasks with three seeds. Because those earlier scores informed
the design, we report the fresh test file beside the official one.

\paragraph{The block.} The state $z \in \mathbb R^{n \times n \times 128}$ holds one vector per ordered pair of items. A
per-task embedding writes the input into it once: the pair facts (an edge, its weight, equality, the source marker) through
a small network, and the item facts and numbers through linear maps on the diagonal cells and rows. One loop reads
$t = \mathrm{norm}(z)$, each cell's 128 numbers shifted to mean zero and scaled to unit spread, and forms two test scores
per channel, $a = W_A t + b_A$ and $c = W_B t + b_B$. The count half (channels 1 to 64) uses hard tests
$A_\ell(i, k) = [a_\ell(i, k) > 0]$ and $B_\ell(k, j) = [c_\ell(k, j) > 0]$ and counts
\begin{equation}
N_\ell(i, j) = \sum_{k} A_\ell(i, k)\, B_\ell(k, j) ,
\end{equation}
a whole number at every size, with $k$ over the real items of the input. Channels $2m - 1$ and $2m$ pass on two bits,
$[N_{2m-1} > N_{2m}]$ and $[N_{2m-1} = N_{2m}]$. The best-witness half (channels 65 to 128) gives each leg a cost
$\gamma = -\log \sigma(\cdot)$ and keeps the best witness,
\begin{equation}
S_\ell(i, j) = \exp\Big(-\min_k \big[\max(\gamma^a_{ik}, \gamma^c_{kj}) + \tau_\ell \min(\gamma^a_{ik}, \gamma^c_{kj})\big]\Big),
\label{eq:mbest}
\end{equation}
with $\tau_\ell \in [0, 1]$ learned per channel: $\tau = 0$ scores a route by its weaker leg and $\tau = 1$ by the product
of its two legs. With $y$ the 64 bits and 64 strengths, the loop writes
$z \leftarrow z + \sigma(W_G t + b_G) \odot (W_O y + b_O)$ and then $z \leftarrow z + \mathrm{MLP}(\mathrm{norm}(z))$, with a
GELU network of width 312. The block has 146,552 parameters, of which 128 are shape numbers
that this configuration never reads. On the five weighted graph tasks (Bellman--Ford, Dijkstra, shortest paths in acyclic
graphs, Prim, Kruskal) the embedding also receives each edge weight in the graph's own units, divided by the median over
nodes of each node's lightest edge, through three fixed maps; the block, its width, its loops and its training are the
same on all thirty tasks. Read-outs are linear maps of the state: a pointer from $i$ is a softmax over $j$ of a
score of $z_{ij}$, a mark of item $i$ reads $z_{ii}$, and Floyd--Warshall's pair pointer is a softmax over the middle item
of a bilinear score of $z_{ik}$ and $z_{kj}$. Per task, the embedding and read-outs bring each model to 0.15 to
0.19 million parameters (185,720 to 194,876 counting an edge read-out that
only edge tasks use). One loop costs $O(n^3)$ per channel for the joins and $O(n^2)$
dense maps per pair.

\paragraph{Training an exact layer.} A hard test has no gradient. We use its exact value in the forward pass and the slope
of a sigmoid in the backward pass (straight-through\cite{bengio2013estimating}): $A = [a > 0]$ forward, $\sigma'(a)$
backward. The comparison bits take the slope of $\sigma(N_1 - N_2)$, measured in units of one witness, for ``more'', and ``same'' the
slope of the bump $4\sigma(1 - \sigma)$, which is zero at a tie and points towards it (Supplementary Note~2). All test weights start random with biases at zero, so the counts differ from pair to
pair from the first step; starting every test open makes every count equal and every bit constant, and training then
does not leave that point. The general setting trains each task for 80,000 steps of 32 inputs with AdamW at a
learning rate of $10^{-3}$, 2,000 warm-up steps, a cosine decay to 1\% and weight decay 0.01, with the number of loops
drawn from 4 to 8 for each batch; the final parameters are read at eight loops. A run took a median of
64 minutes on one NVIDIA L4.

\paragraph{The cold setting.} The input's numbers leave the embedding and are kept, exactly, in registers: numbers on
items (keys, positions, start and finish times) in a register per item, edge weights $W(u, v)$ in a register per pair.
Numbers measured in the same units share a register type (a key and the searched-for target in binary search; start and
finish times in activity selection); every other number has a type of its own. Task scheduling compares each deadline
$d_i$ with a count of tasks taken, which the block can form only as a count of positions below the deadline; so for this
task deadlines are given in units of the input's size, $d_i/n$, as positions are ($k/n$), share their register type,
and the positions are kept as given in training instead of randomised by the sampler. A channel pair of the count half that is wired to a register type compares the
best register value among its admitted witnesses, and how many witnesses reach it, between its two channels, and passes on
``more'' and ``same'' of these pairs (Proposition~1); channel pairs wired to no register count as above. The best-witness half
passes on a bit, whether any middle item passes both tests, in place of a strength. For the path and tree tasks the block
also builds one number per pair, $R(u, v)$, starting from $W$ and updated by
\begin{equation}
R(u, v) \leftarrow \min\Big(R(u, v),\ \min_{k \in K_{uv}} \big[R(u, k) \oplus W(k, v)\big]\Big),
\label{eq:grow}
\end{equation}
where $K_{uv}$ are the middle items admitted by a learned pair of tests and $\oplus$ is addition for shortest paths,
$a \oplus b = a + b$, and the larger of the two for Kruskal's and Prim's algorithms, $a \oplus b = \max(a, b)$; for those
two the join takes $R(u, k) \oplus R(k, v)$ in place of $R(u, k) \oplus W(k, v)$. The state receives only bits about
$R$: whether a candidate improves $R(u, v)$, whether the edge $(u, v)$ is its own best route, and, for admitted $k$, the
comparison of $R(k, u) + W(u, v)$ with $R(k, v)$ and its mirror, which compares two routes between the same two ends. Each
loop rebuilds the state from the embedding and the current bits, $s = e + W_O y + b_O$ and $z = s + \mathrm{MLP}(\mathrm{norm}(s))$,
and a bit of the best-witness half, once on, stays on. The comparison half and the grow step read the state rebuilt
from the kept bits and the comparison bits that involve no learned test; the best-witness half reads the state rebuilt
from all the bits. For the comparison bits on registers the backward pass lets the
slope through only where it would flip the bit the loss wants changed, a perceptron-style rule; elsewhere it is zero. The
cold block has no gate: 169,208 parameters in all for a path task, 16,512 fewer than the general
setting's model (Supplementary Note~2).

For Floyd--Warshall the read-out also sees which middle node wins the join, and for Prim's algorithm the block builds
a hop count along the tree by the same join (Supplementary Note~2).

\paragraph{Ablation.} Two arms change only the block's aggregation, on all thirty tasks with seed 0 and the general
setting's width, loops, steps, data and read: a soft witness sum in place of the exact counts (half cold), and the all-warm block of the first version of this paper\cite{mohsenvand2026witnesses} (Supplementary Note~2). The forecasts were registered before the runs.

\paragraph{Reading at rest, and the stopping rule.} The cold block is read when its bits, and $R$, equal those of the
loop before; since the state is rebuilt from the bits, the future then repeats (Proposition~3). We registered a cap of
$4n$ loops; no input reached it. Training draws 4 to 8 loops per batch for the order tasks, and 8 to 12 for the shortest-path tasks and 6 to 10 for
Kruskal's algorithm, where hand-set versions of the block come to rest within 7 and 4 loops at 16 items; runs are capped at 20,000 steps. Training stops, and the model is fixed, at the first check (every 250 steps) at which the 16-item validation
inputs are all right at every loop from the first trained loop to 16, for the fourth check in a row. The rule reads only
validation inputs. For quickselect the first trained loop is 2 instead of 4; its registered run used 4 (Supplementary Note~2).

\paragraph{Theory.} We state the four results the main text uses; proofs are in Supplementary Note~1.

\emph{Proposition 1 (cold limit).} Let $F$ be a finite set of witnesses with scores $s(w)$, $Z_T = \sum_{w \in F}
e^{s(w)/T}$, $s^\ast = \max_w s(w)$ and $g^\ast = |\{w : s(w) = s^\ast\}|$. Then $T \log Z_T \to s^\ast$ and
$Z_T e^{-s^\ast/T} \to g^\ast$ as $T \to 0^+$. For two families, $Z_T(F_1)/Z_T(F_2)$ tends to $\infty$, to $g_1^\ast/g_2^\ast$ or to 0 as $s_1^\ast$ is
greater than, equal to or less than $s_2^\ast$; so if $(s_1^\ast, g_1^\ast)$ exceeds $(s_2^\ast, g_2^\ast)$ in
lexicographic order, $Z_T(F_1) > Z_T(F_2)$ at every small enough $T$. (When the pairs are equal the warm comparison is
decided by lower terms and can depend on $T$; the cold comparison calls it a tie.) The ratio tends to one if and only if the pairs are equal. For a product of independent leg sums (a route of two
legs) the leading terms combine as $(s_1 + s_2, g_1 g_2)$, and across
disjoint families as the better pair, with counts added when the scores tie.

\emph{Theorem 2 (invariance).} Let $\mathcal G$ be a group of recodings of an input's numbers that preserves the
task's answer. Suppose the numbers enter the learned state nowhere except through bits $[q_1 < q_2]$ and $[q_1 = q_2]$:
no embedding of keys or weights, while positions, indices and graph structure, including which edges are present, may
enter. Suppose further that each compared pair $(q_1, q_2)$ is built so that every $g \in \mathcal G$ maps it to
$(\phi(q_1), \phi(q_2))$ for one strictly increasing $\phi$, which may depend on $g$ and on the pair. Then the bits, the states,
the rest point and the answer are the same on $x$ and $gx$ for every $g \in \mathcal G$. Instances: (i) increasing maps of
keys, with built quantities formed by minimum, maximum and counting; (ii) node potentials
$W'(u, v) = W(u, v) + \varphi(u) - \varphi(v)$ and positive factors, with path sums built by equation~(\ref{eq:grow}),
which stays equivariant because $R(u, k) + W(k, v)$ moves by $\varphi(u) - \varphi(k) + \varphi(k) - \varphi(v)$, and with
comparisons only between two routes with the same two ends; potentials keep every cycle sum, so they create no negative
cycle; (iii) increasing maps of weights, with the bottleneck form of equation~(\ref{eq:grow}). The comparisons are made
within one middle item and then asked whether such an item exists: comparing ``the best over $k$'' with ``the best over
$k$'' would mix routes with different ends, which a potential does not preserve, so here the symmetry dictated the
wiring. The theorem is a statement in exact arithmetic; Supplementary Note~1 gives when it holds exactly in 32-bit floating
point. In the orbit test no answer changed.

\emph{Proposition 3 (rest).} Assume no negative cycle, as the benchmark's positive weights guarantee and node
potentials preserve. The cold block's state after a loop is a triple: the comparison bits $c$, the kept bits $L$ and the
built numbers $R$. Kept bits only turn on, and each $R(u, v)$ never rises and takes values that are walk weights (sum
form) or edge weights (bottleneck form), of which only finitely many lie below any bound without negative cycles; so $L$
and $R$ change finitely often, say $M$ times in all. The comparison bits are read on the built numbers a loop starts
with, so a change of $R$ reaches the bits that involve no learned test one loop later, and the learned tests, which
read those bits, one loop after that; once $L$ and $R$ have not changed for three loops the whole triple therefore
repeats, and since
the next triple is a function of the present one it repeats forever: the block rests within $3M + 3$ loops (Supplementary
Note~1).

\emph{Proposition 4 (sequential depth).} One loop of the block is sums and minima over the middle item, comparisons and
a fixed per-pair network on numbers of polynomially many bits, computable in NC, so a polylogarithmic number of loops
computes a function in NC. Lexicographic depth-first search order is P-complete\cite{reif1985depth}, also when the order
is set by a numbering of the vertices, as in the benchmark, since the first branch of the search is then the
lexicographically first maximal path\cite{anderson1987parallelism}; from the
benchmark's depth-first parents the visiting order follows in NC by an Euler tour: a node's children are discovered in
increasing index and the roots are taken in increasing index, so the visiting order is the preorder of the forest with
children sorted by index. So unless NC = P no block looped polylogarithmically many times computes the parents on every graph\cite{greenlaw1995limits};
the argument, as in the first version of this paper\cite{mohsenvand2026witnesses}, does not bound a block looped polynomially many times.

The block distinguishes at most what the folklore two-dimensional Weisfeiler--Leman test
does\cite{maron2019provably,morris2019weisfeiler}; with positions in the starting state that bound does not bind here
(Supplementary Note~1).

\paragraph{The orbit test.} For each model we recode the numbers of every official test input, recompute the true answer
with the library's own algorithm, keep the inputs whose true answer is unchanged (all of them in every case reported), run
the saved model and count the answers that differ from its answers on the original input. One-answer tasks use the first 512 of the 2,048 official arrays. Order tasks use seven
increasing maps of the keys $k$: $k^3$, $\sqrt k$, $e^{3k}$, $\tfrac14 + \tfrac12 k$, $k/10$, $k + 1$ and $4k$. Shortest
paths use four node potentials, $\varphi(u) = c\, d(u)$ with $d(u)$ the shortest distance from the source and $c = 0.5$,
$-0.5$ and $0.9$, or from a second node with $c = 0.9$, and the factors $\tfrac14$ and 4. With $c$ near one, the edges of
shortest routes are recoded to weights near zero, so many routes nearly tie. Kruskal's algorithm uses five increasing maps of the weights
$w$: $\sqrt w$, $\tfrac12 + \tfrac12 w$, $e^{3w}$, $w^3$ and $w^{10}$. Each model first reproduces its saved predictions
exactly.
These symmetries have been used to probe what trained reasoners represent\cite{mirjanic2023latent}, and networks
homogeneous in their edge weights are exactly equivariant to scaling\cite{tang2020towards}.

\paragraph{Depth-first search with more loops.} The same general block was trained on 8 to 32 loops per batch, for
80,000 steps, and read after 64 loops, the number of nodes of the test graphs. Nothing else changed. In a second version, registered before it ran, the best-witness half also passes on a bit, and every loop rebuilds
the state from the embedding and the current bits, as in the cold setting, but no bit is kept and positions stay in the
embedding, so recoding them changes answers (squaring them changes up to 669 of 2,048
parents). It was trained and read alike. A third version, also registered before it ran, keeps the positions out of
the state: they ride in a register and enter only as ``more'' and ``same'' bits among admitted witnesses, as numbers do
in the cold setting. Fresh graphs of 96 and 128 nodes (eight per size and seed, edges drawn with probability one half)
are read after $n$ loops. The bit
block also wanders during training: its official score after 64 loops, checked every 10,000 steps, fell to
81.93\% (seed 0, step 60,000) and 73.10\% (seed 1, step
30,000); all three were at 100\% at steps 70,000 and 80,000, and the registered model is the last.

\paragraph{Dynamic programs with the recurrence given.} For three dynamic programs we also wrote each recurrence as a
fixed program of the join of equation~(\ref{eq:grow}): the program's channels admit every middle item with no learned
test, while the other channels' tests and the read-out are learned; these ceilings enter no score we
report as ours (Supplementary Note~2).

\paragraph{What the block holds.} For three general-setting models (breadth-first search, Bellman--Ford, quickselect;
seed 0) we recorded the state after every loop on the official test inputs, the tests and counts inside the block and
its bits. Principal directions are fitted once on the normalised state pooled over all loops. Linear probes are logistic
regressions fitted on half of the test inputs and scored on the other half; for pointers, the probe scores each candidate
and the highest is taken. A test is matched to a relation (40 to 60 candidates per task) when their balanced accuracy is
at least 95\%. Every model reproduced its saved predictions at every loop before analysis.

\paragraph{One frontier reasoning model on CLRS-Text.} As a probe, not a survey of such models, we queried one frontier
reasoning model, GPT-6 Astra (API identifier gpt-6-astra; results in Supplementary Note~4), through OpenAI's Responses API
on 13 November
2026, with reasoning effort ``high'', no tools and an output limit of 128,000 tokens, one call per question, retrying only
on transport or server errors. Questions are the benchmark library's own text format\cite{markeeva2024clrs} of the first
eight official test inputs (64 items) and the first eight validation inputs (16 items) of eight tasks (insertion sort,
binary search, quickselect, breadth-first search, depth-first search, Bellman--Ford, Prim's algorithm, longest common
subsequence), with full-precision numbers, preceded by two worked examples of the same task and size, as in the
benchmark's text version, and one sentence of instruction. Answers are parsed into the library's output format and scored
with its metric; an answer that does not parse scores zero. Our block is scored on the identical inputs from its saved
predictions, and its time per answer is measured on a laptop processor (Apple M4, one input at a time, after
compilation). We did not record its time per answer on a GPU.

\paragraph{Maximum retrieval.} The task, its sampler and its test follow ref.~\cite{velickovic2025softmax}: a set of
items, each with a class (one of ten) and a priority drawn uniformly from $[0, 1]$; the answer is the class of the
item of highest priority. Training draws set sizes uniformly from 5 to 16, 100,000 steps of 128 sets with Adam at
$10^{-3}$; testing uses 10,240 sets per size from 16 to 16,384, the same sets for every arm and seed. The softmax network
is theirs. Our replication does not reproduce their published curve: it falls up to 18.5 points below
it under one reading of the weight penalty and up to 11.4 under the other (Supplementary Note~4), so
every comparison with softmax attention in the text and in Fig.~\ref{fig:fail}b uses their published curve, and our
runs serve only to locate the cause of its failure. The cold pair count
admits a rival $j$ of item $i$ through a learned test on the bit $[\rho_j > \rho_i]$ and both classes, counts the
admitted rivals exactly, and reads the item with the smallest count at zero temperature, trained with the straight-through
slope. The warm-count control has the same input and is warm throughout: a sigmoid test, a soft count and a softmax over
counts with a learned inverse temperature; it is read warm and, from the same weights, cold. Ten seeds each. The weight
penalty is $10^{-3}$ times the sum of squared parameters in our first softmax replication, following the published code,
and half that in its rerun, the cold count and the control. The offsets-alone predictor takes each softmax network's key
score for an item of each class at priorities 0.9993 and 0.9998, turns the class differences into offsets $b(\kappa)$ in
priority units, and predicts the cold read by simulating 10,240 sets per size, priorities uniform, classes uniform, and
picking the item that maximises $\rho + b(\kappa)$; at 4,096 items it lands within 0.5 points of
each network's measured cold read on average ($r = 0.99$, 20 networks under
both penalties). The offsets are not an artefact of the weaker replication: the warm-count control, which never computes
a softmax key, also learns a class-dependent level inside its soft count, its mean admission varying by
1.2 to 2.7 thousandths across classes, and this level, not the comparison, is what
fails its warm read; read cold, the same weights ignore it.

\paragraph{Constraint problems.} Instances are generated with a known answer, half satisfiable. Each variable and each
constraint is an item, and the answer is read at one global item's own pair. Half of the 2-SAT
test instances are twins, pairs of a satisfiable and an unsatisfiable instance built alike, whose contradiction is a
chain of a known number of implications; two-colouring has the same construction with odd cycles of known length; the
parity check uses plain and twisted systems of equations over three variables on the cube and the Petersen graph,
which no three-variable counting rule separates. 2-SAT trains on 2 to 16 variables and two-colouring on 3 to 32, both with peak learning rate $3 \times
10^{-4}$ and 30,000 updates; 2-SAT trains with a tight loop budget, 7 to 8 loops at its largest training size, where the
first run used 9 to 12 loops and a peak rate of $10^{-3}$. The answer is read after $K(n) + 4$ loops from
weight-averaged parameters (16 snapshots between updates 15,000 and 30,000); positions are withheld, so no answer can
depend on the order of the items, and renumbering four test instances ten times each changed no logit.

\paragraph{The committee.} Starting from the saved models of three seeds, we change the training graphs of 16 nodes one
edge at a time by a random walk, keep the changed graphs on which the saved models disagree by a fixed rule, and fill
half of a new training pool with them; the other half is the standard pool. A model is trained once on this pool, with
the original recipe and the same number of updates, and read after $n$ loops. The rule, the pool size and the forecasts
were registered before the run.

\paragraph{Language models.} Kinship contexts are generated with a known answer: lines of descent or pedigrees over
full names, each parent link stated in one of eight wordings, two of which never occur in training; questions are half
yes, spread over chain lengths, and half no (reversed, unrelated, or naming a person absent from the context). The
frozen model (Qwen2.5-7B or 1.5B) reads each context once. Mentions are found without labels: the commonest words of
the training sentences, cut at the largest ratio between consecutive frequencies, are the closed-class words, and a
mention is a maximal run of other words. A mention's key is the sum of the model's input embeddings of its tokens, centred
and scaled by the keys of the first 20,000 training statements; two mentions name the same person when their cosine
exceeds a threshold set from training contexts alone: among all within-context pairs of mentions, the largest gap between
consecutive cosines in the upper half, less the smaller of 0.03 and half that gap (0.97 in every run). Each sentence is
read alone, twice, followed by `` Therefore, the parent is'' and by `` Therefore, the child is''; the score is the logit
of the first-named person's first distinguishing token minus the second's, and the reading is the sign of the difference
between the two passes. Wordings are grouped by the cosine of the summed embeddings of their closed-class words (a new
group below 0.98), and each group's direction is the majority of its sentences' readings over the whole text being read.
In the first layout a pair of mentions gets a bit for being named first and second in one statement of a given reading;
in the second, each mention also carries the other name of its statement. The model alone receives the context as text,
with none of this. The block has three kept bits per pair of mentions, tied legs and tests
that cannot ask for a bit to be off, and a read-out that can say only yes; it trains for 4,000 updates on contexts of 1
to 16 statements, and a start counts as learned when it is right on 99\% of a 16-statement validation split. Every start
is reported. Reads after $K$, $K + 4$ and $K + 16$ loops, with $K = \lceil \log_2 m \rceil + 3$ for $m$ mentions; $K + 4$
is the registered read.

\paragraph{CLUTRR.} We use CLUTRR's two splits for systematic generalisation, trained on stories of 2 to 4 relations or
of 2 and 3, and tested on 2 to 10, with the people taken from the brackets the data put around names, as other systems
do. For every two people named near each other, the frozen Qwen2.5-7B-Instruct, at a pinned revision, completes
``[B] is [A]'s'' with one of the twenty relation words or ``unknown'', and we keep its probability for the word. The
block keeps one bit per relation for each ordered pair. Its rule table is trained, with weights bounded and a rule on
above one half, on the stories of two relations with one loop, and then frozen; a read-out of 40 numbers gives the
answer. In the cost read, a read relation costs $-\log p$, a derived relation the cost of its cheapest derivation, and
the answer is the cheapest relation at the question's pair. The reading, its threshold and which read to report first
were chosen on validation stories, and the test stories were read once. The model alone gets the story and the question
without examples, apart from one invented example of direction, once answering in one word and once after up to 768
tokens of reasoning, its answer restricted to the twenty relation words. On graph input the block reads each story's own
graph; this trainer went through four earlier builds that were scored on the test stories before its final design (the
table learned from the stories of two relations with one loop, then frozen) was fixed, whereas every choice on text input
was made on validation stories. The trained read-in is a linear map from the model's last-layer state at the position where it would complete
the relation to 21 scores, fitted first to the model's own reading and then trained from the answers of the training
stories through the cost read, with the minimum over derivations relaxed to be differentiable and the rule table of the
zero-shot stage held fixed. The number of updates (12,000 against 3,000) and each seed's threshold were chosen on
validation stories, and the test stories were read once.

\paragraph{Certificates.} The checks of Section~5 run by script on each saved block, on the averaged weights that every
reported number was read with, in the trained code's single-precision arithmetic (Supplementary Note~5). They take the
block's input bits as given and assume that each statement and the question name two different people.

\paragraph{Compute and cost.} General-setting runs took 64 minutes (median) on one NVIDIA L4; cold
runs at most 29 minutes, because the stopping rule ends them early. Each shared block for all thirty
tasks trained for about 10 hours on one NVIDIA B200, the final one about 13. The cloud cost of the runs reported, including
the ablations and development runs that the Supplement uses, is \$1,318 to \$1,408, of which
the main CLRS-30 runs, one general block per task on thirty tasks with three seeds, take \$113 to
\$203. Only \$269 of it was read from billing; the rest are upper bounds from recorded machine
time and rates, and for the language-model runs our own figures. The cost of eight runs, two seeds of the shared
block for all thirty tasks, one earlier shared block and five of the eleven-task blocks, was not recorded, and their caps
add up to \$329. The probe of one frontier reasoning model cost \$22 to \$32 in API calls.
The final shared block for all thirty tasks cost about \$260 more, its test runs included. All analyses of
trained models ran on a CPU from saved parameters.

\paragraph{Use of AI systems.} AI systems were used extensively. The idea, the research direction, the experiment design
and the choice of problems are the author's. The code, the proofs and drafts of the text were produced by AI systems
under the author's direction. Every reported number is generated by script from the saved output of its run, except the
costs above and the forecasts of Supplementary Table~1, which were compiled from billing reads, run logs and the
registers into data files that the script reads.

\paragraph{Data and code availability.} The trained models are public on HuggingFace: one block per task on CLRS-30
(\url{https://huggingface.co/mohsenvand/cold-attention-clrs30}); one block for all thirty tasks
(\url{https://huggingface.co/mohsenvand/cold-attention-clrs30-shared}); the controls, alternative reads and registered
failures reported beside them (\url{https://huggingface.co/mohsenvand/cold-attention-clrs30-variants}); the shared blocks on
eleven tasks of the depth study (\url{https://huggingface.co/mohsenvand/cold-attention-clrs30-eleven}); the 2-SAT and
two-colouring blocks (\url{https://huggingface.co/mohsenvand/cold-attention-csp}); the strong-components block trained on the
committee's graphs (\url{https://huggingface.co/mohsenvand/cold-attention-scc-committee}); the blocks beside a frozen language
model (\url{https://huggingface.co/mohsenvand/cold-attention-organ}) and on CLUTRR
(\url{https://huggingface.co/mohsenvand/cold-attention-clutrr}); and the maximum-retrieval networks
(\url{https://huggingface.co/mohsenvand/cold-attention-max-retrieval}). An earlier release of the per-task CLRS-30 models,
which the first of these supersedes, remains at \url{https://huggingface.co/mohsenvand/witness-clrs}. The code and results of the first version of this paper, from which Figs.~\ref{fig:set} and~\ref{fig:fail}a take the SET joins and the hand-set layers, are at
\url{https://github.com/NeoVand/witnesses}. CLRS-30 is public\cite{velickovic2022clrs}. CLUTRR's data are licensed CC BY-NC
4.0 and are not redistributed: the code downloads the benchmark's own files and checks their SHA-256 hashes.
The code is at \url{https://github.com/NeoVand/cold-attention}, and a browser demonstration at \url{https://neovand.github.io/cold-attention}.

\begin{figure}[tbp]
\centering
\includegraphics[width=\linewidth]{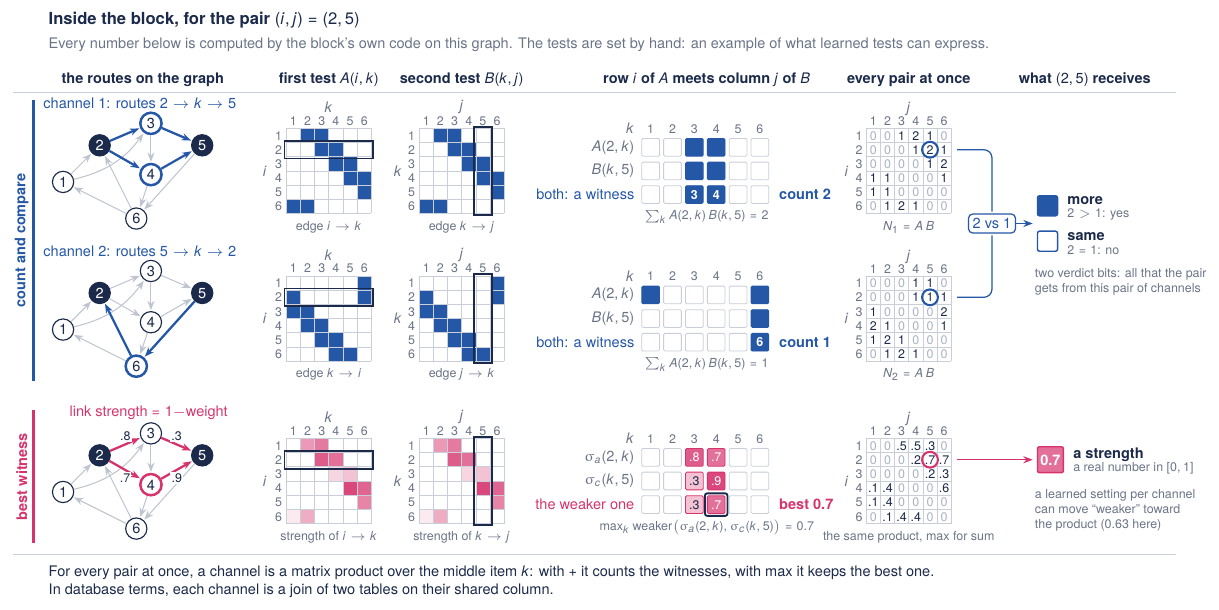}
\refstepcounter{edfig}\label{edfig:arch}
\caption*{\figlabel{Extended Data Fig. \theedfig}{Inside the block, for the pair $(i, j) = (2, 5)$.} One pair of count
channels (cobalt) and one channel of the best-witness half (rose), worked out on the example graph of
Fig.~\ref{fig:arch}a. Row 2 of the first test meets column 5 of the second over every middle item $k$: $A$ and $B$ in the
count channels, the strengths $\sigma_a$ and $\sigma_c$ in the best-witness channel, as in Fig.~\ref{fig:arch}a. The first count channel tests
the legs forward, an edge from $i$ to $k$ and from $k$ to $j$; the second tests them in reverse, an edge from $k$ to $i$
and from $j$ to $k$, so it counts the routes from 5 back to 2. With ``+'' the channel counts the
witnesses, and the pair receives only the two verdict bits (2 against 1: more, not the same); with ``max'' it keeps the
best, and the pair receives a strength (0.7). For all pairs at once each channel is a matrix product over the middle
item. The tests are set by hand, to show what learned tests can express; they are not read from a trained model. Every
count, bit and strength drawn is computed by the block's own code.}
\end{figure}

\clearpage
\section*{Supplementary Information}

\section*{Supplementary Note 1: proofs}
The proofs in this note are not machine-checked; what they state about the benchmark's depth-first search was checked
against its reference code.

\paragraph{Proposition 1 (cold limit).} Let $F$ be finite and non-empty, $Z_T = \sum_{w \in F} e^{s(w)/T}$, $s^\ast = \max_w s(w)$ and
$g^\ast = |\{w : s(w) = s^\ast\}|$. Then
\[
Z_T\, e^{-s^\ast/T} = g^\ast + \sum_{w : s(w) < s^\ast} e^{(s(w) - s^\ast)/T} \;\to\; g^\ast ,
\]
since every exponent in the sum is negative and $T \to 0^+$; and $T \log Z_T = s^\ast + T \log(Z_T e^{-s^\ast/T}) \to s^\ast$,
because the logarithm tends to $\log g^\ast$, a finite number. For two families,
$Z_T(F_1)/Z_T(F_2) = e^{(s_1^\ast - s_2^\ast)/T}\,(g_1^\ast + o(1))/(g_2^\ast + o(1))$, which tends to $\infty$, to
$g_1^\ast/g_2^\ast$ or to 0 as $s_1^\ast$ is greater than, equal to or less than $s_2^\ast$. If $s_1^\ast > s_2^\ast$, or
$s_1^\ast = s_2^\ast$ and $g_1^\ast > g_2^\ast$, the ratio tends to a limit above one, so it exceeds one at every small
enough $T$. Equal pairs give a ratio tending to one, and lower terms decide the warm comparison: the families with scores $\{0, -1\}$ and
$\{0, -2\}$ have the same pair while the first sum is larger at every $T$. \emph{Composition.} For a route of two legs
chosen independently from $F_1$ and $F_2$ with additive scores, $\sum_{a, b} e^{(s(a) + s(b))/T} = Z_T(F_1) Z_T(F_2)$,
whose leading term is $g_1^\ast g_2^\ast e^{(s_1^\ast + s_2^\ast)/T}$. For disjoint families,
$Z_T(F_1 \cup F_2) = Z_T(F_1) + Z_T(F_2)$, whose leading term is the larger of the two, with the counts added when the best
scores are equal. These are the operations of the max-plus semiring with multiplicities, as in counting shortest
paths. $\square$

For the empty family we set $(s^\ast, g^\ast) = (-\infty, 0)$, the value of the limit pair when no term is present.
This is what the block computes: a count over no witness is $0$, the lowest register value over no witness is
$+\infty$, and two empty families compare as ``same''. The pairs that compare built numbers differ in one case: two
absent numbers are not reported as equal, because ``equal'' there asserts that both routes exist.

\paragraph{Theorem 2 (invariance).} We prove instance (ii); (i) and (iii) are the same argument with one strictly
increasing map $f$ in place of the maps $\phi_{uv}$, using that $f$ commutes with minimum and maximum and keeps the
number of items that attain them. Let $g$ act by $W'(u, v) = \gamma\,(W(u, v) + \varphi(u) - \varphi(v))$ with
$\gamma > 0$, and write $\phi_{uv}(\omega) = \gamma\,(\omega + \varphi(u) - \varphi(v))$; each $\phi_{uv}$ is strictly
increasing, and $\phi_{uk}(a) + \phi_{kv}(b) = \phi_{uv}(a + b)$. Let $(c_\ell, L_\ell, R_\ell)$ be the comparison bits,
kept bits and built numbers after $\ell$ loops. \emph{Claim:} for every $\ell$,
\[
c_\ell(gx) = c_\ell(x), \qquad L_\ell(gx) = L_\ell(x), \qquad R_\ell(gx)(u, v) = \phi_{uv}\big(R_\ell(x)(u, v)\big),
\]
with the same pairs absent. \emph{Start.} $R_0 = W$ off the diagonal and $0$ on it, and $\phi_{uu}(0) = 0$; a recoding
adds or removes no edge; the embedding reads no number; the starting bits compare $R_0$ with $W$ on the same pair. \emph{Step.} The state a loop reads is rebuilt from the embedding
and the bits $(c_\ell, L_\ell)$, so it is the same on $gx$ and $x$; hence so is every learned test, every set $K_{uv}$
of admitted middle items, and $L_{\ell + 1}$. In the grow step each candidate $R_\ell(u, k) + W(k, v)$ becomes
$\phi_{uk}(R_\ell(u, k)) + \phi_{kv}(W(k, v)) = \phi_{uv}(R_\ell(u, k) + W(k, v))$, and $\phi_{uv}$, being increasing,
commutes with the minimum over the same set $K_{uv}$; so $R_{\ell + 1}(gx)(u, v) = \phi_{uv}(R_{\ell + 1}(x)(u, v))$.
Every bit compares two quantities that are both routes between the same two ends: a candidate against $R(u, v)$; the
edge $W(u, v)$ against $R(u, v)$; for an admitted $k$, $W(u, v) + R(k, u)$ against $R(k, v)$, both walks from $k$ to $v$;
and its mirror, $W(v, u) + R(k, v)$ against $R(k, u)$, both walks from $k$ to $u$. Here $k$ is the common origin. A
number that is absent (no edge, or no route yet) counts as $+\infty$: ``less'' needs its first member present and
``equal'' needs both, so on a directed graph the mirror pair is $0$ wherever the reverse edge $(v, u)$ is missing. A
recoding adds or removes no edge, so the same members are absent on $gx$ and on $x$; and both members, when present,
are moved by the same $\phi$, which keeps $<$ and $=$. So $c_{\ell + 1}(gx) = c_{\ell + 1}(x)$. \emph{Conclusion.} The read-out reads the state rebuilt from the
bits, so the answers agree after every loop. The block is read when $(c, L, R)$ repeats; the $\phi_{uv}$ are one-to-one,
so $R$ repeats on $gx$ exactly when it does on $x$, and the two runs are read at the same loop. Potentials keep every
cycle sum, so they create no negative cycle.

In instance (i) a channel pair hands on ``more'' and ``same'' of the pairs $(s^\ast, g^\ast)$ of its two channels, with
$s^\ast$ minus the lowest register value among the admitted middle items and $g^\ast$ the number that attain it. A
strictly increasing $f$ sends $(s^\ast, g^\ast)$ to $(-f(-s^\ast), g^\ast)$ in both channels: the same items attain the
lowest value, so $g^\ast$ is unchanged, and the order of the two lowest values is kept. The lexicographic comparison
of the two pairs therefore gives the same two bits.

For Prim's algorithm the state also holds the hop counts $H$. Under a strictly increasing map $f$ of the weights,
$R_\ell(fx) = f(R_\ell(x))$ entrywise, because $f$ commutes with the maxima and minima of the grow step; so
$R(u, v) = W(u, v)$ holds on $fx$ exactly when it holds on $x$, and $R$ moves in a loop on $fx$ exactly when it moves on
$x$, $f$ being one-to-one. $H$ is built from these two facts, from the tests and from sums of hop counts, none of
which reads a weight. Hence $H_\ell(fx) = H_\ell(x)$ for every $\ell$, and the bits that compare hop counts agree. $\square$

\emph{Floating point.} The theorem is a statement in exact arithmetic. In 32-bit floating point it holds exactly for
factors that are powers of two, and for strictly increasing maps of keys or weights that stay strictly increasing after
rounding, because the block then performs the same operations on both inputs. For node potentials it holds exactly when
every sum the block forms is exactly representable; otherwise two routes that tie, or nearly tie, can be ordered
differently after the potential is rounded into the weights. In the orbit test no answer changed.

\paragraph{Proposition 3 (rest).} Assume no negative cycle, as the benchmark's positive weights guarantee and node
potentials preserve. Let $(c_\ell, L_\ell, R_\ell)$ be the comparison bits, the kept bits and the built numbers after
loop $\ell$, and write $c = (c^{\mathrm{own}}, c^{\mathrm{lrn}})$, where $c^{\mathrm{own}}$ are the comparison bits that
involve no learned test: the comparisons of a pair's own item numbers, which depend on the input alone, and the two
bits $[R(u, v) < W(u, v)]$ and $[R(u, v) = W(u, v)]$. One loop is
\begin{align*}
L_{\ell + 1} &= L_\ell \vee E(c_\ell, L_\ell), \\
R_{\ell + 1} &= \min\!\big(R_\ell,\ G(c^{\mathrm{own}}_\ell, L_{\ell + 1};\, R_\ell)\big), \\
c_{\ell + 1} &= C(c^{\mathrm{own}}_\ell, L_{\ell + 1};\, R_\ell),
\qquad\text{so that}\quad c^{\mathrm{own}}_{\ell + 1} = \mathrm{Own}(R_\ell),
\end{align*}
where $E$ is the best-witness half, $G$ the grow step of the Methods and $C$ the comparison half. $E$ reads the state
rebuilt from all the bits. $G$ and $C$ read the state rebuilt from $c^{\mathrm{own}}_\ell$ and $L_{\ell + 1}$ only: the
comparison bits made by learned tests are not fed back to the comparison half or to the grow step, and reach the
future only through $L$, which never loses a bit. Call loop $\ell + 1$ \emph{quiet} if $L_{\ell + 1} = L_\ell$ and
$R_{\ell + 1} = R_\ell$.

\emph{(i) $L$ and $R$ stop.} $L$ only gains bits, so it changes at most once per bit. Each $R(u, v)$ never rises, and
each of its values is the weight of a walk from $u$ to $v$ (sum form) or an edge weight (bottleneck form). Without
negative cycles the walk weights below a bound take finitely many values: each is the weight of a simple path plus a
non-negative combination of the weights of simple cycles, and a cycle of weight zero adds nothing. So $R$ changes
finitely often. Let $M$ be the total number of changes available to $L$ and $R$; at most $M$ loops are not quiet.

\emph{(ii) Three quiet loops give rest.} Let loops $\ell - 1$, $\ell$ and $\ell + 1$ be quiet, with $\ell \ge 2$. Then
$R_{\ell - 2} = R_{\ell - 1} = R_\ell$, so $c^{\mathrm{own}}_{\ell - 1} = \mathrm{Own}(R_{\ell - 2}) = \mathrm{Own}(R_{\ell - 1})
= c^{\mathrm{own}}_\ell$; and $L_\ell = L_{\ell + 1}$. Hence
$c_{\ell + 1} = C(c^{\mathrm{own}}_\ell, L_{\ell + 1}; R_\ell) = C(c^{\mathrm{own}}_{\ell - 1}, L_\ell; R_{\ell - 1}) = c_\ell$,
and the whole triple after loop $\ell + 1$ equals the triple after loop $\ell$. Since the next triple is a function of the
present one and the input, it then repeats forever.

\emph{(iii) Bound.} Before the first repeat no three loops in a row are quiet, so it comes within $3M + 3$ loops. Two
quiet loops can be followed by a change, along the chain $R \to c^{\mathrm{own}} \to c^{\mathrm{lrn}} \to L$. Let $R$
move in loop $\ell$. Then $c^{\mathrm{own}}_{\ell + 1} = \mathrm{Own}(R_\ell)$ can differ from
$c^{\mathrm{own}}_\ell = \mathrm{Own}(R_{\ell - 1})$ while loop $\ell + 1$ is quiet. The learned tests of loop $\ell + 2$
read the state rebuilt from $c^{\mathrm{own}}_{\ell + 1}$, so $c^{\mathrm{lrn}}_{\ell + 2}$ can differ from
$c^{\mathrm{lrn}}_{\ell + 1}$ while loop $\ell + 2$ is quiet. Then $L_{\ell + 3} = L_{\ell + 2} \vee E(c_{\ell + 2},
L_{\ell + 2})$ can gain a bit. By (ii), three quiet loops cannot be followed by a change. Without built numbers $c^{\mathrm{own}}$ depends on the input alone, $c_{\ell + 1}$ is a function of
$L_{\ell + 1}$, one quiet loop after the first gives a repeat, and the bound is $M + 2$ with $M$ the number of kept bits.

\emph{(iv) The read.} The block is read at the first loop whose triple equals the triple of the loop before. By (ii)
this is a fixed point of the loop, for every setting of the weights, and by (iii) it is reached. We registered a cap of
$4n$ loops as a fallback; no input reached it.

\emph{With the hop count.} For Prim's algorithm the state is $(c, L, R, H)$ and a loop is quiet when $L$, $R$ and $H$ are
all unchanged. $H$ is not monotone over the whole run: in a loop in which $R$ moves, $H$ is started again from the new
$R$. But $R$ moves finitely often by (i), and between two moves of $R$ (and after the last) $H$ only falls and its
values are whole numbers of hops, at least 0; so $H$ changes finitely often as well. Let $M_L$ and $M_R$ be the changes
available to $L$ and $R$, as in (i), and $M_H$ the largest number of changes of $H$ in one stretch between two moves of
$R$, or after the last, its restart included: in a stretch each $H(u, v)$ becomes finite at most once and then only
falls through whole numbers, so $M_H$ is finite. There are at most $M_R + 1$ stretches, so
\[
M \le M_L + M_R + (M_R + 1)\, M_H ,
\]
a product because every move of $R$ starts $H$ again. With this $M$, (ii) and (iii) hold as written, with
$c^{\mathrm{own}}_{\ell + 1} = \mathrm{Own}(R_\ell, H_\ell)$: the comparison bits are evaluated on the $R$ and $H$ the
loop started with.

\emph{Not covered.} The depth-first search block keeps no bits: its bits are a function of the previous loop's bits and
can in principle cycle. It is read after $n$ loops, with no claim of rest; measured, every official graph is entirely
right at every loop from 41 to 192 on all three seeds.

\emph{What kind of fixed point.} The kept half is \emph{inflationary}: a loop adds bits and never removes one. That is
all the argument uses. It is not in general monotone, because a learned test may be switched off by a bit that turned
on earlier, so this is the inflationary fixed point of database theory\cite{abiteboul1995foundations} and not the
least fixed point of a monotone map; when every test only responds positively to kept bits the two coincide and the
loop is Datalog evaluation. Equilibrium reasoners stop a continuous state at a tolerance\cite{georgiev2024deep};
this state is discrete and its kept part only grows, so the stop is exact.

\emph{The rest of $R$.} With every middle item admitted, the sum form holds after $\ell$ loops the best walk of at most
$\ell + 1$ edges, as in Bellman--Ford, so $R$ is final within $n - 2$ loops ($n \ge 2$); the bottleneck form,
$\min_k \max(R(u, k), R(k, v))$, doubles the route length each loop, so $R$ is final within
$\lceil \log_2 (n - 1) \rceil$ loops. Measured: every official input rested within $4n$ loops, at most
39 at 64 items and 29 at 128. $\square$

\paragraph{Proposition 4 (sequential depth).} One loop of the block is sums and minima over the middle item, comparisons and a
fixed per-pair network on numbers of polynomially many bits, each computable in NC. So $O(\log^k n)$ loops compute a
function in NC. From the library's depth-first parents and its fixed lexicographic neighbour and
restart order, the depth-first visiting order follows by an Euler tour of the forest, which is in NC: a node's children are
discovered in increasing index and the roots are taken in increasing index, so the visiting order is the preorder of the
forest with children sorted by index. Deciding that order (is $u$ visited before $v$) is P-complete for fixed ordered
adjacency lists, in directed and undirected graphs\cite{reif1985depth}. The library's case, where the search order is set
by a numbering of the vertices, follows from the lexicographically first maximal path, the path from a start vertex that
always moves to the unvisited neighbour of smallest number: deciding whether a vertex lies on it is P-complete for
undirected graphs with numbered vertices, even planar graphs of degree at most three\cite{anderson1987parallelism}
(Problem A.3.1 of ref.~\cite{greenlaw1995limits}). This path is the first branch of the depth-first tree from its start
(ref.~\cite{greenlaw1995limits}, Problem A.3.3, Hint). Number the start vertex first, which leaves the order of the
others unchanged; then the start is the library's first root, its first branch is found from the parents in NC by
following each node's child of smallest index, and a vertex lies on that branch exactly when it lies on the path. So,
unless NC $=$ P, no such block computes the depth-first parents on every graph. The
argument does not bound a block looped polynomially many times, and loops in proportion to $n$ are what the experiments
use. $\square$

\paragraph{Expressive power.} Both halves of the block update $z_{ij}$ as a function of $z_{ij}$ and the multiset $\{(z_{ik}, z_{kj})\}_k$, so pairs
that the folklore two-dimensional Weisfeiler--Leman test does not distinguish get the same
state\cite{maron2019provably,morris2019weisfeiler}; the block therefore distinguishes at most what that test does, which is
what three-variable counting logic\cite{cai1992optimal} and homomorphism counts of patterns of treewidth at most
two\cite{dell2018lovasz} distinguish. This bound is for inputs without names for their items. The benchmark gives
every item a position, and with positions in the starting state the test separates all pairs, so here the limits that
bind are the number of loops and the width, not the test. Within the bound, the block holds a fixed number of bits for
each pair, so it cannot carry a count that grows with the input from one loop to the next, as counting homomorphisms
requires; it can decide whether a pattern of treewidth at most two maps into the graph, and which of two witness counts
is larger.

\section*{Supplementary Note 2: further methods}

\paragraph{The cold forward pass.} Write $t = \mathrm{norm}(s)$ for the normalised state of a cell,
$a_\ell(i, k) = (W_A t_{ik} + b_A)_\ell$ and $c_\ell(k, j) = (W_B t_{kj} + b_B)_\ell$ for the two test scores of channel
$\ell$, $A_\ell(i, k) = [a_\ell(i, k) > 0]$, $B_\ell(k, j) = [c_\ell(k, j) > 0]$, and
$S_\ell(i, j) = \{k : A_\ell(i, k) \text{ and } B_\ell(k, j)\}$ for the admitted middle items. The bit vector $y$ has 128
entries per pair: 64 from the comparison half (32 channel pairs, two bits each) and 64 from the best-witness half. One
loop of the resting block:
\begin{align*}
L_\ell(i, j) &\leftarrow L_\ell(i, j) \vee \big[S_{64 + \ell}(i, j) \ne \emptyset\big], \qquad \ell = 1, \dots, 64
   && \text{tests read from } s = \mathrm{rebuild}(c, L) \\
s' &= \mathrm{rebuild}(c^{\mathrm{own}}, L), \\
\mathrm{rebuild}(y) &= u + \mathrm{MLP}(\mathrm{norm}(u)), \qquad u = e + W_O y + b_O, \\
c &\leftarrow \text{the comparison half's 64 bits, tests read from } s' \text{ (below)} \\
y &= (c, L) .
\end{align*}
The comparison half and the grow step read the state rebuilt from the kept bits and the comparison bits that involve
no learned test, the learned comparison bits being zero in this input only ($y$ keeps them for the read-out and the next
loop's best-witness half); the best-witness half reads the state rebuilt from all the bits. The first loop starts from
$L = 0$ and $c = (c^{\mathrm{own}}_0, 0)$, with $c^{\mathrm{own}}_0$ computed on the registers and, on path tasks, on
$R_0$: $W$ off the diagonal, and on it $0$ for sums (a weighted self-loop's weight on Floyd--Warshall) and $-\infty$ for
the bottleneck form. Channels $2m - 1$ and $2m$ form a
pair, wired to one register type or to none:
\begin{align*}
\text{plain facts:}\quad & N_\ell = |S_\ell(i, j)|, \qquad
   \text{more} = [N_{2m-1} > N_{2m}], \quad \text{same} = [N_{2m-1} = N_{2m}] . \\
\text{item register } \rho_\ell:\quad & E_\ell = \min_{k \in S_\ell(i, j)} \rho_\ell(k) \ \ (+\infty \text{ if empty}), \qquad
   E^\ast = \min(E_{2m-1}, E_{2m}), \\
 & n_\ell = \big|\{k \in S_\ell(i, j) : \rho_\ell(k) = E^\ast\}\big|, \qquad
   \text{more} = [n_{2m-1} > n_{2m}], \quad \text{same} = [n_{2m-1} = n_{2m}] . \\
\text{built numbers:}\quad & \text{less} = [V_1 < V_2], \quad \text{level} = [V_1 = V_2, \text{ both present}]; \\
\text{per witness:}\quad & \text{less} = \big[\exists k \in S_{2m-1}(u, v) : V_1(k) < V_2(k)\big], \\
 & \text{level} = \big[\exists k \in S_{2m-1}(u, v) : V_1(k) = V_2(k), \text{ both present}\big] .
\end{align*}
$V_1$ and $V_2$ are as in the table below. ``Best'' is the lowest register value, so in the scores of Proposition~1 $s^\ast = -E$. An item register enters a
witness through the middle item, $\rho(k)$; a number given once per input (the target of binary search) is the same for
every $k$. ``More'' is exactly $[E_{2m-1} < E_{2m}$, or equal with more witnesses attaining it$]$. For each combination
of register types $(\rho, \rho')$ the first two pairs are fixed, with no learned test: $S = \{i\}$ against $S = \{j\}$,
which hands on $[\rho(i) < \rho'(j)]$ and $[\rho(i) = \rho'(j)]$, and the reverse. In the general setting every admitted
middle item has the same score, so the pair $(s^\ast, g^\ast)$ reduces to the count $g^\ast$, compared within one loop;
both members are used only on register-wired channels, where ties in $g^\ast$ arise for numbers given once per input,
for deadlines in units of $n$, and for empty sets.

\begin{center}\small
\begin{tabular}{@{}l r r r r@{}}
\toprule
Setting & plain facts & item registers & of which fixed & built numbers \\
\midrule
General & 32 & 0 & 0 & 0 \\
Cold order tasks & 16 & 16 & 2 per combination & 0 \\
Cold depth-first search & 16 & 16, (position, position) & 2 & 0 \\
Cold path and tree tasks & 16 & 8, (position, position) & 2 & 8 \\
\bottomrule
\end{tabular}

\vspace{2mm}
\begin{tabular}{@{}>{\raggedright\arraybackslash}p{38mm} >{\raggedright\arraybackslash}p{40mm} >{\raggedright\arraybackslash}p{52mm} r@{}}
\toprule
Task & register kinds & combinations & pairs each \\
\midrule
minimum, four sorts, quickselect & \{position\}, \{key\} & (pos, pos), (key, key) & 8 \\
binary search & \{position\}, \{key, target\} & (pos, pos), (key, key), (key, target), (target, target) & 4 \\
activity selection & \{position\}, \{finish, start\} & (pos, pos), (f, f), (f, s), (s, s) & 4 \\
task scheduling & \{position, deadline$/n$\}, \{weight\} & (pos, pos), (pos, d), (d, d), (w, w) & 4 \\
\bottomrule
\end{tabular}

\vspace{2mm}
\begin{tabular}{@{}l l l >{\raggedright\arraybackslash}p{50mm}@{}}
\toprule
Pair & $V_1$ & $V_2$ & over $k$ \\
\midrule
grow & $\min_{k \in K_{uv}} R(u, k) \oplus W(k, v)$ & $R(u, v)$ & the minimum, which becomes the new $R(u, v)$ (bottleneck form: $R(u, k) \oplus R(k, v)$) \\
edge & $R(u, v)$ & $W(u, v)$ & none (no test) \\
forward $\times 3$ & $R(k, u) \oplus W(u, v)$ & $R(k, v)$ & some admitted $k$ \\
mirror $\times 3$ & $R(k, v) \oplus W(v, u)$ & $R(k, u)$ & some admitted $k$ \\
\bottomrule
\end{tabular}
\end{center}

The eight built-number pairs are the last eight of the 32; each has one learned pair of tests (its first channel's),
except ``edge'', which has none; none keeps a tie count. An absent number counts as $+\infty$. $K_{uv}$ in Methods eq.~(\ref{eq:grow}) is the admitted set of
the grow pair's first channel, one set per pair and block. The eight pairs fill the last 16 comparison entries of $y$, in
place of their ``more'' and ``same'', and $W_O$ reads them as any bit; every built-number bit is computed on the $R$ (and
$H$) the loop started with. Floyd--Warshall replaces the first forward and mirror pairs by two ``via'' pairs
($V_1 = R(u, k) + W(k, v)$, $V_2 = R(u, v)$; the read-out also receives, per $k$, ``admitted and equal''). Prim's
algorithm replaces the first two forward and the first two mirror pairs by four hop pairs: the grow pair on $H$, with
$+$, and three that compare $H(k, v)$ with $H(k, u)$ in either order. The depth-first search block keeps no bits: $y$ is
computed afresh each loop from the state rebuilt from the whole previous $y$, with the best-witness bit not latched.

\paragraph{Parameters.} Both norms are parameter-free (subtract the mean, divide by the spread).
\begin{center}\small
\begin{tabular}{@{}l r@{}}
\toprule
Part & numbers \\
\midrule
tests $W_A$, $W_B$ ($128 \times 128$ each), $b_A$, $b_B$ & 33,024 \\
write $W_O$, $b_O$ & 16,512 \\
gate $W_G$, $b_G$ (general setting only) & 16,512 \\
network $128 \to 312 \to 128$ with biases & 80,312 \\
shape numbers, three vectors of 64 & 192 \\
\textbf{general block} & \textbf{146,552} \\
\textbf{cold block} (no gate) & \textbf{130,040} \\
graph-task embedding and read-outs & 39,168 \\
\textbf{cold path model in all} & \textbf{169,208} \\
general model on a graph task & 185,720 \\
\bottomrule
\end{tabular}
\end{center}
The 128 shape numbers the general configuration never reads are a temperature and a size exponent per channel, used only
by the ablation's soft sum; the third vector is $\tau$ of Methods eq.~(\ref{eq:mbest}). In the cold setting $W_G$ and $b_G$ do not
exist, and $\tau$ shapes only the backward slope of the best-witness bit; it receives no gradient there, so it keeps its starting
values, evenly spaced from 0 to 1. Floyd--Warshall's model has two more numbers, the
gains of its via pairs (169,210). Test weights of channels whose tests are fixed or absent are allocated and never read:
in a path block 13 of the 64 comparison channels ($4 + 2 + 7$: both channels of the two fixed position pairs, both of
the ``edge'' pair, and the second channel of each of the other seven built-number pairs, which admit through their first
channel's tests), 3,354 numbers at 258 per channel; in an order block, the four channels of the two fixed pairs of each
combination of register types. The released files omit the 128 shape numbers that only the ablation reads (general
block 146,424, cold block 129,912).

\paragraph{Backward rules.} \emph{Tests.} $A = [a > 0]$ forward; $\partial A / \partial a := \sigma(a)(1 - \sigma(a))$.

\emph{Counts of plain facts (the general setting, and the plain pairs of the cold setting).} With
$d = N_1 - N_2$ and $\sigma = \sigma(d)$,
\[
\frac{\partial\,\text{more}}{\partial N_1} := \sigma(1 - \sigma) = -\frac{\partial\,\text{more}}{\partial N_2}, \qquad
\frac{\partial\,\text{same}}{\partial N_1} := \frac{d}{dd}\big[4\sigma(1 - \sigma)\big] = 4\sigma(1 - \sigma)(1 - 2\sigma)
 = -\frac{\partial\,\text{same}}{\partial N_2} .
\]
So ``same'' is given the slope of a bump of height one centred on the tie. The slope is zero at a tie and points
towards the tie on either side; it is not sign-blind. $N_\ell = \sum_k A_\ell(i, k) B_\ell(k, j)$ is differentiated as
written.

\emph{The gate (perceptron-style).} For the bits of register-wired pairs and of built numbers the arriving gradient is
replaced by
\[
\Gamma_b(g) = \begin{cases} \max(g, 0) & \text{if the bit } b = 1, \\ \min(g, 0) & \text{if } b = 0, \end{cases}
\]
that is, a bit at one passes only requests to fall and a bit at zero only requests to rise. A bit that is already as
the loss wants it sends nothing back.

\emph{Pairs on item registers.} With $n_1, n_2$ the counts at the joint lowest value $E^\ast$ and $d = n_1 - n_2$, let
$\tilde d = d$ if both channels have a witness at $E^\ast$, $\tilde d = \mathrm{clip}(d, -1, 1) = \pm 1$ if only one has,
and $\tilde d = d = 0$ if neither has ($E^\ast = +\infty$, both counts zero); $\sigma = \sigma(\tilde d)$. Then
\[
\frac{\partial \mathcal L}{\partial P_1(i, k, j)} := \Gamma_{\text{more}}(g_{\text{more}})\,\sigma(1 - \sigma)\,[\rho_1(k) \le E^\ast]
 \;+\; \Gamma_{\text{same}}(g_{\text{same}})\,4\sigma(1 - \sigma)(1 - 2\sigma)\,[\rho_1(k) = E^\ast],
\]
with $P_\ell(i, k, j) = A_\ell(i, k) B_\ell(k, j)$ the admission of $k$, and the opposite sign for the second channel.
The slope of ``more'' goes to every middle item, admitted or not, whose value is at or below $E^\ast$, since those are
the items whose admission changes the comparison; the slope of ``same'' goes to the items exactly at $E^\ast$. When only
one channel attains $E^\ast$, the other's slopes go only to items it does not admit, at or below $E^\ast$, whose admission
would let it reach $E^\ast$, and the clip holds the slope of ``more'' at $\sigma'(1)$ however large $d$ is; when neither
channel has a witness, ``more'' sends $\sigma'(0) = 1/4$ to every middle item and ``same'' sends nothing. The
gradient of a test score follows by the product rule, for example
$\partial \mathcal L / \partial a_1(i, k) = \sum_j \big(\partial \mathcal L / \partial P_1(i, k, j)\big)\, B_1(k, j)\, \sigma'(a_1(i, k))$.
No register value receives a gradient.

\emph{Built numbers.} For $V = \min_{k \in K} X(k)$: $\partial V / \partial P(k) := -1$ for the admitted $k$ that attain the
minimum and for every item not admitted whose value is below it (admitting it would lower $V$), and $0$ otherwise;
and $\partial V / \partial X(k) := 1$ at the admitted $k$ that attain it (shared equally among ties), so slopes also reach
the $R$ of earlier loops and, through it, their admissions.
For ``less'' $= [V_1 < V_2]$: $\partial\,\text{less} / \partial V_1 := -1$, $\partial\,\text{less} / \partial V_2 := +1$. For
``level'' $= [V_1 = V_2]$, the slope of $-|V_2 - V_1|$: $\partial\,\text{level} / \partial V_1 := \mathrm{sign}(V_2 - V_1) = -\partial\,\text{level} / \partial V_2$,
where, here and below, the slope of $|x|$ at $x = 0$ is $+1$ (the convention of the automatic differentiation we train
with), so a level bit at a tie still passes slopes of $\pm 1$.
For the per-witness pairs: $\partial\,\text{less} / \partial P(k) := [V_1(k) < V_2(k)]$, and
``level'' takes the slope of $\sum_k P(k)\,[V_1(k) = V_2(k)] - |V_1(k^\circ) - V_2(k^\circ)|$, with $k^\circ$ the admitted
item whose two values are present and closest (no second term if there is none): so
$\partial\,\text{level} / \partial P(k) := [V_1(k) = V_2(k)]$ and $\partial\,\text{level} / \partial V_1(k^\circ) :=
-\mathrm{sign}(V_1(k^\circ) - V_2(k^\circ)) = -\partial\,\text{level} / \partial V_2(k^\circ)$; ``less'' sends no slope to
the values. $\Gamma$ acts once, on the gradient that arrives at each bit, before every slope above; the admission and
value slopes behind it are not gated again. No slope depends on the size of a number.

\emph{The best-witness bit.} Forward $[S \ne \emptyset]$; backward the slope of the strength of Eq.~(4).

\emph{Ties.} At a tie the slope of ``same'' is zero, for plain facts and item registers alike, so a wrong ``same'' is
undone only through ``more'', whose slope there is $\sigma'(0) = 1/4$, and only when the loss also asks ``more'' to
change. On register-wired pairs $\Gamma$ lets a ``more'' at zero pass only requests to rise, so a tie there can be broken
only in favour of the pair's first channel. No other backward rule breaks ties. Built numbers do not share this limit:
at a tie ``level'' still passes slopes of size one, to the two values and, per witness, to the admission of each tying
item.

\paragraph{Floyd--Warshall and Prim's algorithm in the cold setting.} For Floyd--Warshall, whose answer for $(u, v)$ is the last middle node of a shortest route, the
read-out adds, for each candidate $k$ and each of the two via pairs, that pair's learned gain (starting at 8) times the
bit ``$k$ is admitted by the pair and
$R(u, k) + W(k, v) = R(u, v)$'', the winner of the join; an edge from a node to itself counts as an edge, as the library's
files require. For Prim's algorithm the block builds a second number per pair, the hops $H(u, v)$ along tree edges: 0 on
the diagonal, 1 where $R(u, v) = W(u, v)$, grown as $H(u, v) \leftarrow \min\big(H(u, v), \min_{k \in K^H_{uv}} H(u, k) +
H(k, v)\big)$, with $K^H_{uv}$ the middle items admitted by the hop pair's own tests, and started again whenever $R$
moved in that loop; the state receives, for admitted $k$, whether $H(k, v) <
H(k, u)$, and the reverse. Training draws 8 to 12 loops per batch for
Floyd--Warshall and 10 to 14 for Prim's algorithm, where hand-set versions of the block come to rest within 7 and 9
loops at 16 nodes; the stopping rule and the read are as for the other path tasks. In the orbit test Floyd--Warshall uses
the same potentials as the shortest-path tasks, measured from node 0, and the same factors; Prim's algorithm uses
Kruskal's five increasing maps. Both additions were registered, with forecasts, before the runs.

\paragraph{The ablation blocks.} In the half-cold arm the count-and-compare half is replaced by a soft
witness sum, $\big(n^{-\alpha}\sum_k (a_{ik} c_{kj})^{1/\tau}\big)^{\tau}$ with sigmoid legs $a$, $c$ and a temperature
$\tau \in [1/16, 1]$ and size exponent $\alpha \in [0, 1]$ learned per channel; the best-witness half is kept. In the
all-warm arm the block is the earlier witnesses paper's: an OR-join half and a power-mean half with
inputs shifted by a learned multiple of $\log n$. Each arm runs the general setting's job for every task with only the
aggregation changed: width 128, eight loops at test, 80,000 steps, the same data and seed 0, read at eight loops with
the final parameters. Each job carried a registered time cap (5,210 to 10,880 seconds); every run recorded its full
80,000 steps, so no cap stopped a run. The forecasts were registered before the runs.

\paragraph{The general-and-cold block.} Its two halves are whole blocks of 128 channels: a general block, as in Methods
with its gate, and a cold block, as in the cold setting, with no gate. They act on two states per pair: the general state
$z$, written once by the task's general embedding, and the cold bits $y = (c, L)$, from which the cold state is rebuilt
each loop with the task's cold embedding $e^{\mathrm c}$, which receives no number. The cold half's wiring names no task.
Of its 32 comparison pairs, 12 count plain facts; 12 compare item registers, four in each of three register rows
(positions, then the task's other item numbers and any number given once per input; a row a task does not fill is
zero), each pair within one row, the first two of each row with the fixed tests $\{i\}$ against $\{j\}$ and the reverse;
and 8 are the built-number pairs on a pair register $W$ of edge weights in the bottleneck form, so that every built value
is an input weight and no sum is formed ($W$ is absent on tasks without a graph). One loop is
\[
(y, R) \leftarrow \mathrm{cold}(y, R), \qquad
z \leftarrow z + [i, j \text{ real}]\,(U y + u), \qquad
z \leftarrow \mathrm{general}(z),
\]
one loop of the cold setting, then one of the general setting, with $U$ ($128 \times 128$) and $u$ starting at zero. The
cold half never reads $z$, so the coupling runs one way. Each task has its own embedding and read-outs for each half,
and its answer logits are the sum of its general read-out of $z$ and its cold read-out of the state rebuilt from $y$. The
tasks share 293,104 parameters: 146,552 in the general half, 130,040 in the cold half and 16,512 in the
coupling; each task adds 6,016 to 46,688 for each half. The block trains on all tasks at once, on $K(n)$ to $K(n) + 3$
loops with $K(n) = \lceil \log_2 n \rceil + 3$, and is read after $K(n)$ loops from weight-averaged parameters, not at
rest. In the tied-legs variant the last eight of the cold half's 64 best-witness channels use their first test on both
legs, so that each hands on $[\exists k : A_\ell(i, k) \wedge A_\ell(k, j)]$, the square of the relation its test reads;
this adds no parameter and leaves those channels' 1,032 second-leg weights unread.

\paragraph{Dynamic programs with the recurrence given.} For the maximum subarray, the longest common subsequence and
optimal binary search trees we also wrote each recurrence as a fixed program of the Methods' grow step:
which number each register pair builds and from which registers, with the items on a line (each has a next item) that
carries the task's numbers as edge weights. In the program's channels every middle item is admitted, with no learned
test; the block's other channels, which compare the items' stored numbers and keep the cold bits, still learn their
tests, as do the embedding and the read-out. The models
are read at rest; they enter no score we report as ours. With the program's tests learned from a random start instead,
training stopped short of exactness on all three (45.95\%,
97.85\% and 74.61\%), while 7 of the 7
fixed-program models read so far are exact. These models use the same stopping rule and read as the
cold setting, with 19 to 24 training loops (24 to 28 for the longest common subsequence).

\paragraph{Listing number programs.} Where gradient descent does not find optimal binary search trees' program,
listing does. As in enumerative program synthesis\cite{alur2013syntax}, we listed every program of a small grammar in
which each built number grows only by the join of the Methods' grow step: a step register
$S(k, k') = b_k + c_{k'}$ on each item $k$ and its next item $k'$, with $b$ and $c$ each absent, $p$ or $q$ (the task's
two lists of probabilities); one or two built registers, each starting on the diagonal at $0$, $p_i$ or $q_i$ and grown
by one line, $T(i, j) \leftarrow \min\big(T(i, j), \min_k [X(i, k) + Y(k, j)] + Z(i, j)\big)$, where $X$ and $Y$ are $S$,
a built register, or a built register read one item later, and $Z$ is absent or a built register; the answer for a pair
is the winning $k$ of the last line. That gives 219,672 programs. Each was run to rest in 64-bit sums on
200 training inputs (40 at each of the five training sizes) and kept if its answer equalled the stored
root on every pair. 6 fit, and they are one program in different spellings: the interval weight grows
from $q$ by steps of $p + q$, and the cost table grows from itself, itself read one item later, and the interval weight,
as in the textbook recurrence; the spellings differ only in the cost table's start on the diagonal, which adds the same
amount to every tree over an interval and so never changes a winner, and in the tie rule. All are right on all
66,560 pairs of the official test file. In a grammar about six times larger (a step may also carry $p + q$
of one item, and a register may be read one item earlier on its second index; 1,357,824 programs),
12 fit, again one program with a second way to form the interval weight, and again all are right on
every test pair. The two listings took 3 and 18 seconds on one processor, with no
model and no gradient. They do not show that the program can be found without knowing it: we wrote the grammar knowing
the answer, and it contains the right program by construction. What they show is that nothing else in the grammar fits
and that 200 small inputs are enough to say so. Listing is a different learner from gradient descent,
and no score we report as ours comes from it. Matrix chain order is outside this grammar: its extra term is a product of
three numbers.

\section*{Supplementary Note 3: what the channels' tests match}

For three general-setting models (breadth-first search, Bellman--Ford and quickselect, seed 0, official inputs) every
channel's two tests were compared, at loops 1, 2, 3, 4, 8 and 16, with 40 to 60 candidate relations on the same cells,
keeping the best by balanced accuracy (the agreement averaged over the cells where the relation holds and where it does
not, in either polarity). Loop 16 lies beyond the trained loops (4 to 8); the block is run on unchanged. The candidates:
\begin{itemize}\setlength{\itemsep}{0pt}
\item graphs: same node; edge; edge or same node; self-loop; the source (row or column); index order; hop distance of the
row or column at most $t$; hop-distance order; ``$v$ could be $u$'s parent'' (an edge to a node one hop closer); the
answer pointer and its transpose; siblings; ancestor; rejected candidate; pair hop distance at most $t$; tree depth at
most $t$; the model's own pointer after the previous loop;
\item Bellman--Ford adds: distance order; distance quantiles; closer neighbour; light or lightest edge; pair
shortest-path quantiles; the predecessor on the best route of at most $t$ edges;
\item quickselect: key order; rank thresholds; is the median; neighbours in rank; closer to the median; position order;
first or last; the model's current pick.
\end{itemize}
A channel is \emph{dead} when its output never varies (a count with one value on at least 99.5\% of cells, or a
strength with standard deviation below 0.01); \emph{explained} when it is live and every varying test matches a
candidate at 95\% balanced accuracy or more, with any constant test always open; \emph{partly} explained when one test
matches; and \emph{not} explained otherwise. The share of the write weights each channel by the standard deviation of
its output times the norm of its row of $W_O$.

\begin{center}\small
\begin{tabular}{@{}l r c c r@{}}
\toprule
Task & Loop & Count half & Witness half & Share of the write, count half \\
 & & \multicolumn{2}{c}{explained / partly / not / dead, of 64} & \\
\midrule
BFS & 1 & 0 / 12 / 26 / 26 & 10 / 19 / 29 / 6 & 22.1\% \\
BFS & 2 & 6 / 4 / 20 / 34 & 13 / 27 / 20 / 4 & 14.3\% \\
BFS & 3 & 2 / 3 / 21 / 38 & 2 / 14 / 43 / 5 & 13.9\% \\
BFS & 4 & 5 / 3 / 18 / 38 & 5 / 24 / 32 / 3 & 14.9\% \\
BFS & 8 & 5 / 3 / 15 / 41 & 2 / 14 / 43 / 5 & 14.2\% \\
BFS & 16 & 5 / 1 / 11 / 47 & 11 / 15 / 33 / 5 & 15.1\% \\
Bellman--Ford & 1 & 2 / 1 / 6 / 55 & 16 / 13 / 31 / 4 & 6.3\% \\
Bellman--Ford & 2 & 1 / 0 / 6 / 57 & 7 / 9 / 46 / 2 & 6.1\% \\
Bellman--Ford & 3 & 0 / 0 / 15 / 49 & 3 / 8 / 52 / 1 & 12.4\% \\
Bellman--Ford & 4 & 0 / 3 / 19 / 42 & 7 / 5 / 51 / 1 & 18.8\% \\
Bellman--Ford & 8 & 5 / 5 / 13 / 41 & 9 / 10 / 29 / 16 & 47.6\% \\
Bellman--Ford & 16 & 2 / 4 / 14 / 44 & 13 / 5 / 21 / 25 & 58.9\% \\
Quickselect & 1 & 27 / 16 / 17 / 4 & 11 / 20 / 33 / 0 & 49.0\% \\
Quickselect & 2 & 33 / 16 / 5 / 10 & 15 / 20 / 25 / 4 & 35.8\% \\
Quickselect & 3 & 25 / 20 / 9 / 10 & 15 / 21 / 24 / 4 & 37.2\% \\
Quickselect & 4 & 26 / 19 / 9 / 10 & 12 / 20 / 28 / 4 & 37.0\% \\
Quickselect & 8 & 9 / 22 / 23 / 10 & 2 / 23 / 36 / 3 & 38.3\% \\
Quickselect & 16 & 0 / 11 / 41 / 12 & 1 / 4 / 56 / 3 & 42.7\% \\
\bottomrule
\end{tabular}

\end{center}

\section*{Supplementary Table 1: registered forecasts}
Before each run reported here we wrote down what we expected it to show, with odds, and afterwards scored each forecast on
the read the register named. The table lists all 213 forecasts of the 18 runs and follow-ups behind the
paper's main results: 110 came true, 75 did not, and 28 could not be scored because the arm
they were written for never ran or the forecast became void. A miss is a
forecast that did not come true, whatever its odds. Two odds separated by a slash were written independently by two
forecasters before the run, and a forecast marked as added after launch was written after its run had started. Where a
run was rebuilt and its forecasts scored again on the later build, the cell says so. Each row's register and result file
are listed with the released code.

{\small
\begin{longtable}{@{}>{\raggedright\arraybackslash}p{0.21\linewidth}>{\raggedright\arraybackslash}p{0.40\linewidth}>{\raggedright\arraybackslash}p{0.10\linewidth}>{\raggedright\arraybackslash}p{0.21\linewidth}@{}}
\toprule
Run & Forecast & Odds & Outcome \\
\midrule
\endfirsthead
\toprule
Run & Forecast & Odds & Outcome \\
\midrule
\endhead
\bottomrule
\endlastfoot
General block for all 30 tasks, second seed & 30-task mean within 0.4 of the first seed's 96.96 & 70\% & hit: 96.71 (0.25 under) \\
 & 30-task mean above FloydNet's 96.64 & 65\% & hit: 96.71 \\
 & Depth-first search between 70 and 80 after 8 loops & 80\% & miss: 69.29 \\
\addlinespace
General-and-cold block for all 30 tasks (3 seeds) & Three-seed mean at least 97.04 (best single shared block so far) & 70\% / 60\% & miss: 96.55 \\
 & Three-seed mean at least 97.5 & 50\% / 40\% & miss: 96.55 \\
 & Some seed below FloydNet's 96.64 & 15\% / 20\% & hit: two seeds (96.41, 96.48) \\
 & Three-seed mean above the first shared block's 96.87 & 70\% & miss: 96.55 (96.40 like for like) \\
 & Depth-first search three-seed mean at least 80 (first shared block 73.03) & 55\% & miss: 75.50 \\
 & Task scheduling three-seed mean at least the first shared block's 91.85 & 35\% & miss: 88.17 \\
 & Task scheduling below 93 (deadlines lack the positions' units) (added after launch) & 75\% & hit: 88.17 \\
\addlinespace
The same trained block with parts switched off (no training) & Unchanged model reproduces its saved predictions: under 0.1\% of entries differ, every task and split & 90\% & hit: 50, 97 and 90 of 67.4 M entries differ \\
 & Cold read-out off: maximum subarray (Kadane) gains at least 2.9 and task scheduling at least 2.2 (half their losses) & 40\% (25\% on rereading, before the run) & miss: Kadane +0.00; task scheduling +2.66 \\
 & Hand-over from the cold to the general half off: at least 10 of 30 tasks fall below 60 & 80\% & hit: 12 of the 25 tasks read \\
 & Read-out chosen per task on the 16-node validation split differs from the full model on at most 3 tasks per seed & 80\% & hit: 2, 1 and 1 tasks \\
 & Read-out chosen per task on a fresh 64-node split: test mean 96.87 or more & 20\% & miss: 96.23 (25 tasks) \\
\addlinespace
General block for all 30 tasks, with deadlines in the positions' units and depth-first search read after $n$ loops (3 seeds) & Three-seed mean at least 97.3 (first shared block 96.87) & 55\% & miss: 97.20 \\
 & Three-seed mean at least 97.6 & 30\% & miss: 97.20 \\
 & Every seed's mean at least 96.87 & 65\% & hit: 97.34, 97.26, 97.01 \\
 & Task scheduling three-seed mean at least 97 (first shared block 91.85) & 60\% & hit: 99.07 \\
 & Depth-first search read at n loops: three-seed mean at least 85 (first shared block 73.03) & 45\% & miss: 79.62 (76.42, 91.02, 71.44) \\
 & No other task falls by more than 1.0 against the first shared block & 70\% & miss: DAG shortest paths $-2.10$ \\
\addlinespace
Strong components retrained on graphs where saved models disagree & Official 64-node test at least 96 (standard pool 90.0; hand-made pool 98.2) & 55\% & hit: 96.58 \\
 & Components with five or more entry points entirely right in at least 85\% & 50\% & hit: 90.8\% \\
 & The new committee disputes at most a fifth as many changed graphs as the old one & 60\% & miss: ratio 0.27 (490 vs 1,843 of 2,000) \\
 & At 128 nodes, at least 10 points above the standard pool's 68.4 & 55\% & miss: 78.17 (+9.8) \\
\addlinespace
The same, two more seeds & Official test at least 96 on both new seeds & 50\% & hit: 97.31, 97.46 \\
 & Three-seed mean at least 96 & 60\% & hit: 97.12 \\
 & Three-seed mean at least 3 above the standard pool's best seed (92.24) & 80\% & hit: +4.88 \\
 & Five or more entry points at least 85\% on both new seeds & 55\% & miss: 89.86, 84.79 \\
 & 128 nodes: three-seed mean at least 76 & 50\% & hit: 79.95 \\
\addlinespace
A second round of that loop (stopped at its free gate) & Free gate: at least 20\% of disputed nodes on 16-node walks show the component-joining error & 50\% & miss: 1.56\% (17 of 1,087); round 2 not run \\
 & Official test at least round 1's 96.58 & 70\% & not scored: round 2 not run (gate) \\
 & Official test at least 97.5 & 40\% & not scored: round 2 not run (gate) \\
 & Components with 3 to 4 entry points entirely right in at least 93\% & 45\% & not scored: round 2 not run (gate) \\
 & Component-joining errors at most half of round 1's (about 72) & 40\% & not scored: round 2 not run (gate) \\
 & Stopping signal: at most 294 disputed walks (0.6 of 490) & 60\% & not scored: round 2 not run (gate) \\
 & 128 nodes: at least 80.2 & 45\% & not scored: round 2 not run (gate) \\
\addlinespace
Constraint problems learned from yes/no answers & 2-SAT trained to 16 variables: at least 99\% at 500 variables, deep twins included & 50\% & miss: 55.47\% \\
 & Two-colouring trained to 16 variables: at least 99\% at 500 variables & 60\% & not scored: cannot be built as registered; replacement (trained to 32): 100\% \\
 & Three-variable parity without positions, random weights: identical answer on every hard pair, Petersen graph, all ten weight settings & 85\% & hit: bit-identical \\
 & The same on the cube & 60\% & hit: bit-identical \\
 & Parity with positions given: at least 90\% on graphs twice the training size & 15\% & miss: 50.0\% \\
 & Parity with positions given fits its training instances (at least 95\%) & 50\% & miss: 45.83\% \\
\addlinespace
Two-colouring trained to 32 variables & Two-colouring trained to 32 variables: at least 99\% at 500 variables & 55\% & miss: 50.0\%: the run collapsed to a constant answer \\
 & Rerun at a lower learning rate: at least 99\% on validation & 55\% & hit: 100\% \\
 & Rerun: at least 99\% at 500 variables, twins included & 15\% & hit: 100\% \\
\addlinespace
2-SAT with a lower learning rate, then also a tight loop budget & 2-SAT, lower learning rate: at least 99\% at 500 variables & 25\% & miss: 67.97\% \\
 & Same arm: deep twins at 64 variables at least 90\% & 35\% & hit: 100\% \\
 & Lower learning rate and tight loop budget: at least 99\% at 500 variables & 45\% & hit: 100\% \\
 & Same arm: deep twins at 64 variables at least 90\% & 55\% & hit: 100\% \\
 & Same arm: at least 99\% on validation (24 variables) & 75\% & hit: 100\% \\
 & The tight budget changes little or nothing above depth 32 (added after launch) & none stated & miss: right at every depth to 498 \\
\addlinespace
2-SAT, tight loop budget, two more seeds & Both new seeds at least 99\% at 500 variables & 55\% & hit: 99.22\%, 100\% \\
 & At least one new seed at least 99\% at 500 variables & 85\% & hit: both \\
 & Both new seeds 100\% at 64 variables & 80\% & hit: 100\%, 100\% \\
\addlinespace
Two-colouring, two more seeds & Both new seeds 100\% at 500 variables & 65\% & hit: 100\%, 100\% \\
 & At least one new seed 100\% at 500 variables & 90\% & hit: both \\
 & Neither new seed collapses to a constant answer & 75\% & hit: neither \\
\addlinespace
2-SAT on the constructed (forced) block & Pooled read-out learns (at least 99\% on validation) on at least 1 of 5 seeds & 35\% & miss: 0 of 5 (50.0 to 50.8\%) \\
 & Read-out at the global item learns on at least 1 of 5 seeds & 15\% & not scored: not run (stop rule) \\
 & Every seed that learns is right on every instance at 64, 256 and 500 variables & 85\% & not scored: no seed learned \\
 & Every such seed has all bits at rest at 500 variables & 80\% & not scored: no seed learned \\
 & Every such seed's weights read as the two intended rules & 70\% & not scored: no seed learned \\
\addlinespace
Maximum retrieval as sets grow & Softmax replication within 5 points of the published curve at 64 to 4,096 items & none stated & miss: misses by up to 18.5 points \\
 & Rerun with half the weight penalty: within 5 points of the published curve & none stated & miss: misses by 9.6, 11.4, 6.8 at 512 to 2,048 \\
 & Softmax-trained weights read cold: at least 95\% at 4,096 (kill below 90\%) & none stated & miss: 36.3\% and 32.5\% (killed) \\
 & Cold-trained head: at least 95\% at 4,096 & none at registration; 30\% (stated at the rerun) & not scored: not run (replication gate failed) \\
 & Where cold reads fail, the cause is ranking against a rival within about 1/n, not dispersion & none stated & not scored: needs the cold-trained arm; on cold reads, ranking errors 100\%, rival within 2/n only up to 4,096 \\
 & Rerun's weights read cold: at most 60\% at 4,096, limited by ranking resolution & none stated & hit: 32.5\% \\
 & Cold pair count: at least 99.5\% at every size to 16,384 (kill below 99\%) & none stated & hit: lowest ten-seed mean 99.97\% \\
 & Every pair-count error comes from a learned test that admits by class & none stated & hit: no seed admits by class; all 91 errors on tied-top sets \\
 & Resolution law: class offsets predict cold accuracy within 2 points, and the 50\% point lies at n times offset spread in [0.5, 2] & none stated & miss: offsets predict within 0.7 points; 50\% point at 2.70 and 2.26 \\
 & Same comparison inputs, warm throughout: below 90\% at 4,096 & none stated & hit: 41.97\% \\
 & The same weights read cold: at least 99\% at every size & none stated & hit: at least 99.98\% \\
\addlinespace
Cold block beside a frozen 0.5B model: ordering facts & Without training, the nearest-match read is right on every name mention at 1,000 facts for one of layers 4, 8, 12 & 70\% & miss: best 99.96\% (layer 4); only the input layer is exact \\
 & Block on true relation bits: at least 99\% at 1,000 facts in every family & 65\% & miss: 64 of 96 (67\%) \\
 & Middle-layer arm: at least 99\% at 1,000 facts on 2 of 3 seeds & 35\% (revised to 25\% before any run) & not scored: arm dropped \\
 & Middle-layer arm: under one wrong comparison bit in a million at 1,000 facts & 30\% & not scored: arm dropped \\
 & General block: at least 95\% at 16 facts and at most 80\% on long chains at 1,000 & 65\% & hit: 100\%; 0\% (12 questions) \\
 & Input-layer arm within 2 points of the middle-layer arm at every size & 55\% & not scored: arm dropped \\
 & The language model alone: at most 65\% at 16 facts & 85\% & hit: 48.4\% \\
 & Middle-layer arm learns (at least 99\% on validation, 2 of 3 seeds) & 55\% (revised to 40\% before any run) & not scored: no longer applies; arm dropped \\
 & Some attention head ranks the stating line first on at least 90\% of one-step questions at 16 facts & 70\% & miss: best 25\% \\
 & That head's share falls 20 points or more from 16 to 256 facts & 55\% & not scored: void (no such head); best heads fall 25 and 19 points \\
 & When that head is wrong at 256 facts, its top line is a rival line over half the time & 65\% & miss: 6\% \\
 & The model says yes to at least 90\% of one-step questions at 16 facts & 30\% & miss: 61\% \\
 & Laptop: input-layer read at least 95\% on new contexts at 16 facts & 60\% & hit: 100\% \\
 & Laptop: input-layer read at least 90\% at 128 facts & 40\% & miss: 81.3\% \\
 & Laptop: layer-4 read at least 5 points under the input layer at 128 facts & 60\% & miss: 12 points above \\
 & Laptop: a tight loop budget makes the block double its reach on at least 2 of 3 seeds & 55\% & hit: 3 of 3 \\
 & Laptop: the registered budget doubles on at least 1 of 2 new seeds & 60\% & hit: 1 of 2 \\
 & Laptop, kept bits without counts: 100\% at 16 facts on both seeds & 75\% & hit: both \\
 & Every question right at 128 facts on at least one seed & 50\% & hit: seed 0 \\
 & Doubles on both seeds & 60\% & miss: seed 1 misses two long chains \\
 & At least 95\% of questions at rest by the late read at 128 facts, both seeds & 70\% & miss: 93.8\% and 85.4\% \\
 & Late read at least as good as the primary read at 64 and 128 facts, both seeds & 65\% & miss: one late wrong answer (seed 0, 128 facts) \\
 & Kept bits with counts: every question right at 128 facts on one seed & 35\% & not scored: not run \\
 & Laptop, a second late read in the loss: every question right at 128 facts, both seeds & 55\% & miss: 87.5\% (seed 0) \\
 & Every validation question right on the late read, both seeds & 75\% & not scored: held on seed 0; second seed not run \\
 & At least 95\% at rest by the late read at 128 facts, both seeds & 45\% & not scored: not found; second seed not run \\
 & Late read at least as good as the primary read, both seeds & 65\% & not scored: held on seed 0; second seed not run \\
 & Doubles on both seeds & 80\% & miss: seed 0 walks a few facts per loop \\
 & Laptop, tests that cannot ask for a bit to be off: learns on at least one seed & 55\% & miss: 86.5\%, 51.0\% on validation \\
 & Every question right at 128 facts on both reads, one seed & 40\% & miss: no learner \\
 & No learner turns a right answer wrong with more loops & 75\% & not scored: no learner \\
 & Learners: at least 95\% at rest at 128 facts & 50\% & not scored: no learner \\
 & Laptop, one test for both legs and 3 kept bits: learns on at least 2 of 3 seeds & 50\% & miss: 60.4, 67.7, about 71\% \\
 & Learners right on every question at 128 facts & 80\% & not scored: no learner \\
 & Learners double & 85\% & not scored: no learner \\
 & Learners: at least 95\% at rest at 128 facts & 70\% & not scored: no learner \\
 & Laptop, plus a read-out that can only speak for yes: learns on at least 2 of 3 seeds & 55\% & hit: 3 of 3 \\
 & Learners right on every question at 128 facts, both reads & 85\% & hit: all three, also to 1,000 facts \\
 & Learners: at least 95\% at rest at 128 facts & 75\% & hit: every question at rest \\
 & Model's real states: at least 2 of 3 seeds right on every question at 1,000 facts & 85\% & hit: 3 of 3 (96 of 96 each) \\
 & Every seed right on validation is right on every test question, 16 to 1,000 facts & 80\% & hit: 3 of 3 \\
 & No wrong comparison bit at any size & 90\% & hit: 0 of 96.2 M pairs at 1,000 facts \\
 & True relation bits: learns and is right on every question at 1,000 facts & 50\% & miss: 64 of 96 \\
\addlinespace
Cold block beside frozen models: kinship in eight wordings & Laptop, sentence direction given: learns on at least 2 of 3 seeds & 75\% & hit: 2 of 3 \\
 & Laptop, wording pairs given, 4,000 updates: learns on at least 1 of 3 & 35\% & miss: 0 of 3 (68 to 69\%) \\
 & Same at 16,000 updates: learns on at least 1 of 3 & 60\% & hit: 1 of 3 \\
 & Every learner right on every question at 128, 256 and 1,000 statements & 85\% & hit: 4 of 4 \\
 & Control with no direction in the input: validation below 90\% & 85\% & hit: 63.0\% \\
 & Name read: no wrong same-person bit at 1,000 statements & 75\% & miss: no name found (0 of 12,000) \\
 & Wording read right at every size & 85\% & not scored: void (depends on the failed name cut) \\
 & Model's own reading of direction right on at least 99\% of sentences for 6 of 8 wordings & 30\% & miss: no wording above 97.2\% \\
 & Model's reading right on every sentence of all 8 wordings & 10\% & miss \\
 & Kinship is not the relation family the model reads best & 60\% & hit: vacuously: no family is read \\
 & Names found by word: every name found, and no wrong same-person bit at 1,000 statements & 70\% & hit: all spans found; no wrong bit \\
 & Every wording class right at every size & 85\% & hit \\
 & Two-prompt contrast reading right on at least 90\% of test sentences & 40\% & miss: 70.0\% \\
 & Contrast reading right on every sentence & 3\% & miss \\
 & Laptop, trained on six wordings, read on eight: below 90\% at 16 statements, every error a yes-question & 85\% & hit: 66.7\% \\
 & Input bits built from the names' token ids, eight wordings: learns on at least 1 of 3 seeds & 65\% & hit \\
 & On at least 2 of 3 seeds & 45\% & hit: 2 of 3 \\
 & Learners right at 128, 256 and 1,000 statements, all reads & 85\% & hit \\
 & Learners' weights read as the intended program & 70\% & hit \\
 & 0.5B model's reads: no wrong same-person bit or wording class at any size to 1,000 statements & 80\% & hit: 0 wrong in 385 M pairs \\
 & Learns on at least 2 of 3 seeds & 65\% & miss: 1 of 3 \\
 & Learns on at least 1 of 3 seeds & 88\% & hit: 1 of 3 \\
 & Every learner right on every registered question, 16 to 1,000 statements, all reads & 85\% & hit \\
 & General-block control: at least 99\% on validation at 16 statements & 55\% & hit: 100\% \\
 & General-block control: below 90\% at 256 statements & 85\% & hit: 78.1\% \\
 & 7B model's reading, pooled by wording, right for all 8 wordings & 55\% & hit \\
 & 7B per-sentence reading right on at least 95\% of sentences & 25\% & hit: 99.99\% \\
 & 1.5B pooled reading right for all 8 wordings & 35\% & hit: narrowest 57.0\% \\
 & 7B reads `C is a child of P' right on over half its sentences & 60\% & hit: 4,835 of 4,835 \\
 & Each mention also carries its sentence's other name: learns on at least 4 of 5 seeds & 55\% & miss: 3 of 5 \\
 & On at least 3 of 5 & 75\% & hit: 3 of 5 \\
 & Every learner of that layout right on every registered question & 88\% & hit: 3 of 3 \\
 & Earlier layout, five seeds: at most 3 of 5 learn & 80\% & hit: 2 of 5 \\
 & 7B read-in exact: every name found, no wrong same-person bit at any size & 80\% & hit \\
 & Pooled reading gives every wording class the right direction, two unseen wordings included & 85\% & hit \\
 & Pooled: learns on at least 2 of 3 seeds in one layout & 75\% & hit: 3 of 3 in both \\
 & Every pooled learner right on every registered question, 16 to 1,000 statements, all 8 wordings & 80\% & hit: 6 of 6 \\
 & Every per-sentence learner right on every registered question & 45\% & miss: each misses 2 questions \\
 & Every per-sentence learner at least 98\% at every size & 85\% & hit: lowest 95 of 96 \\
 & 7B model alone: at most 65\% at 1,000 statements & 75\% & hit: 43.75\% \\
 & 7B model alone: at least 75\% at 16 statements & 50\% & miss: 73.96\% \\
 & 1.5B pooled: some seed learns and every learner is exact & 60\% & hit: 3 of 3 \\
 & Laptop, restart on high training loss: the dead start is restarted and a later attempt learns & 85\% & hit \\
 & No attempt the rule keeps ends below 99\% on validation & 85\% & hit: one run \\
\addlinespace
Cold block beside a frozen 7B model: CLUTRR & Graph input, trained on 2 to 4 relations: at least 0.97 at every length 5 to 10, each of 3 seeds & 70\% & hit: 1.00 in every cell (final build); missed on builds 1--3 \\
 & Learned table is exactly the benchmark's 62 rules, every seed & 55\% & hit: final build; builds 1--3 had extra rules (10 to 7,938) \\
 & Trained on 2 to 3 relations: at least 0.97 at every length 4 to 10, each seed & 65\% & hit: at least 0.992 (final build); missed on builds 1--3 \\
 & Trained on 2 to 4: at least 96\% overall, each seed & 60\% & hit: 99.1\% (final build); held on build 2, missed on 1 and 3 \\
 & Answers unchanged between the primary and the late read & 90\% & hit: final build; failed on 2 of 3 seeds of one split in build 2 \\
 & Text: the better reader gets at least 90\% of stated relations right (test) & 70\% & miss: 88.6\%, 89.0\% \\
 & At least 95\% right & 35\% & miss: 88.6\%, 89.0\% \\
 & At threshold 0.9, a relation asserted for at most 30\% of unrelated pairs & 45\% & hit: 15.8\% \\
 & Text, trained on 2 to 4: at least 60\% overall (mean of 3 seeds) & 55\% (later estimate 35\%, not scored) & miss: 34.6\% (bit read); cost read 56.5\% \\
 & Text, trained on 2 to 4: at least 75\% overall & 20\% & miss: 34.6\%; cost read 56.5\% \\
 & Text, trained on 2 to 4: at least 40\% at 10 relations & 45\% (later estimate 30\%, not scored) & miss: 19\% (bit read); cost read 51\% \\
 & Text, trained on 2 to 3: above DSR-LM's 60.98\% & 50\% & miss: 36.9\% (bit read); cost read 60.6\% \\
 & Table learned from text holds all 62 true rules & 70\% & miss: 3 of 6 runs \\
 & On test stories read without error: at least 97\% & 75\% & miss: 49\% (bits), 79\% (costs) \\
 & Validation smoke, with an `unknown' answer: a relation for at most 40\% of unrelated pairs & 65\% & hit: 36\% \\
 & With `unknown', still at least 90\% of stated relations right & 60\% & miss: 89.3\% \\
 & The wider sentence window reads more right than the closest window & 60\% & hit: 89.3 vs 85.5\% \\
 & With one made-up example in the prompt: at least 92\% of stated relations right & 50\% & miss: 91.5\% \\
 & With the example: more error-free stories than 86 of 302 & 60\% & hit: 106 \\
 & Model alone, one word: at most 45\% overall & 70\% & hit: 31.3\% \\
 & Model alone, step by step: at most 55\% overall & 60\% & hit: 33.3\% \\
 & Both at most 30\% at 10 relations & 70\% & miss: 26.2\% and 30.3\% \\
 & Step by step above one word & 65\% & hit: 33.3 vs 31.3\% \\
 & Block on text above the better of the two, overall & 65\% & hit: 56.5 vs 33.3\% (cost read) \\
 & Block on text above the better of the two at 10 relations & 60\% & hit: 51 vs 30\% (cost read; bit read 19\%) \\
 & Cost read at least 2 points above the bit read on validation (2 to 4 split) & 70\% & hit: 76.4 vs 68.2\% \\
 & Cost read above the bit read on test & 70\% & hit: +22 points \\
 & Cost read at least 60\% overall on test (2 to 4 split) & 45\% & miss: 56.5\% \\
 & Cost read above DSR-LM's 60.98\% (2 to 3 split) & 45\% & miss: 60.6\% \\
 & Cost read at least 5 points above the bit read at 10 relations & 55\% & hit: 51 vs 19\% \\
 & Read-in fitted to the model's own reading agrees with it on at least 95\% of validation pairs & 70\% & hit: 97.9\%, 99.3\% \\
 & Read-in trained from answers: validation at least 3 points above 76.4\% (2 to 4 split) & 55\% & hit: 89.8\% \\
 & Test at least 60\% (2 to 4 split) & 50\% (lowered to 40\% before the test read) & hit: 64.6\% \\
 & Test at least 65\% (2 to 4 split) & 25\% & miss: 64.63\% \\
 & Test above DSR-LM's 60.98\% (2 to 3 split) & 60\% & hit: 66.6\% \\
 & Stated relations read right on validation rise to at least 93\% & 40\% & miss: fell to 79\% from 90\% \\
 & Table relearned from the new reading is exactly the 62 rules in 2 of 3 seeds & 35\% & miss: 0 of 3 \\
 & At least 55\% at 10 relations (2 to 4 split) & 40\% & hit: 56.3\% \\
 & 12,000 training updates at least 1 point above 3,000 on validation & 65\% & hit: 89.2 vs 87.3\% \\
 & The 12,000-update version is chosen on both splits & 60\% & hit: both \\
 & Ablation: the untrained read-in within 2 points of 56.5\% on test & 60\% & miss: 45.8\% \\
 & The trained read-in at least 4 points above the untrained one on test, both splits & 75\% & hit: +18.9 and +17.8 \\
\addlinespace
Finite proof of exactness for the trained blocks & Every one of the 15 kinship runs says no on its one-context dual & 90\% & hit: 15 of 15 \\
 & The simulation check passes for all 9 runs of one input layout & 60\% & hit: 9 of 9 \\
 & The simulation check passes for all 6 runs of the other layout & 45\% & hit: 6 of 6 \\
 & A closure set reached within 3 loops in all 15 runs & 70\% & miss: none as written; 19 of 23 in a later chain form \\
 & No run is a tower (right to 1,000 only through squarings that never feed back) & 85\% & hit \\
 & All five 0.5B learners pass both checks, with 5 child-first and 3 parent-first wordings & 60\% & hit: 5 of 5 \\
 & Question about one person also named in a statement: every run answers yes & 80\% & miss: 19 of 20 \\
 & Question about one person named nowhere else: at least 12 of 15 runs answer no & 50\% & miss: 9 of 15 \\
 & The third argument proves every length for the 4 networks left & 55\% & miss: 2 of 4 as written; 4 of 4 after additions made after seeing results \\
 & It also proves at least 17 of the 19 already proved & 80\% & miss: 7 of 19 as written; 19 of 19 after additions \\
\end{longtable}}

\section*{Supplementary Table 2: what each setting builds in}

What each row of the column set by the theory receives beyond the benchmark's inputs, and which rows changed after
their first results. The general setting is the same block and configuration on all thirty tasks; the cold settings add
registers, and on the path and tree tasks a built number per pair. ``Changed'' marks the rows whose recipe was
changed after the results of their first registered recipe; Results reports the column both ways (94.77\%
registered before results; after the changes, 97.98\% with every row on the benchmark's standard
sampler and inputs, 98.35\% with task scheduling and quickselect as trained, and 98.44\% when the
six path and tree tasks also take built route sums, which Results does not use). The three changes and their reasons:
\begin{itemize}\setlength{\itemsep}{0pt}
\item task scheduling: deadlines given in units of the input's size, as positions are, and positions not randomised,
because the general block had learned that a large deadline means yes, true of every training input; registered
before results it is the general setting's 89.00\%. With the sampler's randomised
positions the same cold block scores 98.35\% (seeds 98.18 to 98.49), 1.51 points
lower and below the general block given the same deadlines (99.34\%): the gain over
89.00\% comes from the deadlines' units, and the cold block's lead over the general block needs the
positions kept as given;
\item quickselect: every training size from 4 to 16 and the loss from loop 2, because models fitted on the library's
five sizes were wrong at sizes between them; registered before results it is the two registered seeds' mean,
16.19\% (32.37\% and 0.00\%). An earlier run on seed 0, outside the registered recipe,
reaches 100\% when read at the step where the stopping rule would have stopped it; it is not in the average;
\item depth-first search: 8 to 32 training loops, a read after $n$ loops, then positions in a register, because a fixed
number of loops falls short on this task and the encoded positions failed at 128 nodes; registered before results it
is the general block read after $n$ loops, 86.75\%. Topological sort and strong components took the same
position register once it had fixed depth-first search; registered before results they are the loop with positions
encoded, 100.00\% and 87.79\% (now 100.00\% and
90.02\%).
\end{itemize}

\begin{center}\scriptsize\setlength{\tabcolsep}{3pt}
\begin{tabular}{@{}>{\raggedright}p{27mm} >{\raggedright}p{14mm} >{\raggedright}p{35mm} >{\raggedright}p{30mm} >{\raggedright}p{32mm} l@{}}
\toprule
Tasks & Setting & Inputs, sampler, loops & Registers and built numbers & Bits and read-out extras & Changed \tabularnewline
\midrule
all thirty (general setting) & general & library sampler, 4 to 16 items; trained on 8 loops at test; 80,000 updates; on
Bellman--Ford, Dijkstra, acyclic paths, Prim and Kruskal the weights also enter as $x = w/s$, with $s$ the median over
nodes of each node's lightest edge, through $x/(1 + x/16)$, $\mathrm{clip}(\log x, \pm 4)$ and $e^{-x}$ & none & none & no \tabularnewline
four sorts, minimum & cold & training loops 4 to 8; read at rest & position; key & none & no \tabularnewline
binary search & cold & as above & position; key and target (one type) & none & no \tabularnewline
activity selection & cold & as above & position; start and finish (one type) & none & no \tabularnewline
quickselect & cold & every size from 4 to 16 (eight sizes added to the library's five); first trained loop 2 & position; key & none & yes \tabularnewline
task scheduling & cold & deadlines given as $d/n$; positions as given, not randomised by the sampler & position and deadline (one type) & none & yes \tabularnewline
Bellman--Ford, Dijkstra, acyclic paths & cold & training loops 8 to 12 & weight $W$ per pair; $R$ built by Methods eq.~(\ref{eq:grow}) with $\oplus = +$ & $R$ improves; edge is its own best route; $R(k,u) + W(u,v)$ against $R(k,v)$ and the mirror & no \tabularnewline
Kruskal & cold & training loops 6 to 10 & $W$; $R$ with $\oplus = \max$ & as above & no \tabularnewline
Floyd--Warshall & cold & training loops 8 to 12; a self-loop counts as an edge & $W$; $R$ with $\oplus = +$ & as above; read-out gain (8 at the start) times the join-winner bit & no \tabularnewline
Prim & cold & training loops 10 to 14 & $W$; $R$ with $\oplus = \max$; hop count $H$, restarted when $R$ moves & as above; $[H(k,v) < H(k,u)]$ and the reverse & no \tabularnewline
depth-first search & bit loop & training loops 8 to 32; read after $n$ loops; no kept bits & position & none & yes \tabularnewline
topological sort, strong components & bit loop & as depth-first search & position & none & yes \tabularnewline
the other twelve tasks & general & as the general setting & none & none & no \tabularnewline
\bottomrule
\end{tabular}
\end{center}

\section*{Supplementary Note 4: further results and every loss}

\paragraph{CLRS-30, every way of counting.} Supplementary Table~4 gives every task. The general block (one block per task, three seeds) averages 97.04\%;
FloydNet reports 96.64\% (10 scores from runs with hints) and averages 94.81\% without hints. The setting the theory prescribes
averages 97.98\% on the standard sampler and inputs, 98.35\% with task scheduling and
quickselect as trained (both off the standard inputs), and 98.44\% when the six weighted graph tasks take
the cold block with built route sums (12 wins, 12 ties and 6 losses
against FloydNet's published scores); we treat the latter as a ceiling. With every choice as first registered, before any
result, the column averages 94.77\%. The best of our blocks per task, a selection over blocks and so
optimistic, averages 98.48\%. With built route sums, the cold block is right on every test input on every seed
on 14 of 18 tasks; without them, on 9 of 12.

\paragraph{The shared block for all thirty tasks.} The final shared block is our first one, the general block for all
thirty tasks (96.87\% over three seeds), with deadlines divided by the input's size on task scheduling and
depth-first search read after $n$ loops; both changes were registered before its runs. Its registered read, every task
after 8 loops except depth-first search, averages 97.20\%, 0.33 points above the first block; read like
the first block, every task after 8 loops, 96.92\%. On the fresh 64-node split it averages 97.06\%
against 96.65\%. As a secondary read, not registered as the headline, its weight-averaged parameters average
97.57\%. The gain comes from the two changes: task scheduling rises from 91.85 to 99.07\% and
depth-first search from 73.03\% at 8 loops to 79.62\% at $n$ (seeds 76.42, 91.02
and 71.44\%; read at 8 loops the same models score 71.21\%). It loses on 15 other
tasks against the first block: directed acyclic shortest paths 2.10 points, all on one seed
(93.07\% on its final parameters, 99.40\% over the three seeds weight-averaged), longest common
subsequence 0.99, activity selection 0.54, binary search
0.44 and the optimal search tree 0.36, and less than a quarter of a point on
each of the others. 2 of its 6 registered forecasts held (Supplementary Table~1).

\paragraph{The general-and-cold block.} A shared block with a general half and a cold half averages 96.55\% on its
registered read and 96.40\% read like our first general shared block, 0.46 points below it, and
below the final shared block's 97.20\%; it gains on strong
components and Prim's algorithm and loses on the tasks that compute with numbers. A knock-out read on 25
tasks and three seeds shows how it uses its cold half. With the coupling between the halves switched off, the mean falls
from 96.1\% to 46\%. With only the cold half's read-out switched off, strong components lose
17 points, shortest paths in acyclic graphs
11, Floyd--Warshall 13, matrix chain
39 and the optimal search tree 39, while maximum subarray and
binary search do not move. The deficit therefore lies in how the two halves train, not in the cold read-out: choosing the
read per task after training recovers at most 0.13 points, even when chosen on the test itself.

\paragraph{Item memory against pair memory.} With one state per item instead of one per pair, and the cold reading
unchanged (one seed, a tie counted wrong), the block averages 76.76\% against 98.44\% for the pair
block's column. It stays within half a point on 12 tasks (Breadth-first search, Bellman--Ford, Dijkstra, DAG shortest paths, Insertion sort, Bubble sort, Heapsort, Quicksort, Minimum, Binary search, Quickselect, Activity selector) and averages
61.31\% on the other 18, where the pair block averages 97.40\%: sorting,
selection, search, greedy selection and single-source shortest paths need no middle item; all-pairs paths, depth-first
order and what rests on it, spanning trees, dynamic programs, geometry and string alignment do.

\paragraph{Maximum retrieval.} Our softmax network did not reproduce the published curve: it fell up to
18.5 points below it between 64 and 4,096 items under one reading of the weight penalty and up to
11.4 under the other. The cold-read results do not depend on this. The cold pair count's admission
table is exact on every seed, but its reject margin is thin; a network trained warm and read cold has a wider one.

\paragraph{Constraint problems.} The first 2-SAT run, with a looser loop budget and a higher learning rate, is right on
77.34\% at 64 variables and 55.47\% at 500: it finds contradictions out to about 24 to
31 implications and calls deeper ones satisfiable, and more loops at test do not help. The first two-colouring run, at
the higher learning rate, collapsed to one answer. Parity with positions given, trained on graphs of 4 to 8 vertices,
did not fit its own training instances.

\paragraph{Language models.} On the first task (``is $X$ older than $Y$?'', Qwen2.5-0.5B, single-token names), the general
block was also read at 512 and 1,000 facts on 48 and 12 questions only:
64.58\% and 33.33\%. A control that gave the block hand-made relations in place of the
model's read failed (about two thirds right), because of how those inputs were built. On the kinship task, using the
0.5B model's reads with the wording directions learned from answers, 5 of 10
starts learned.

\paragraph{CLUTRR.} Every CLUTRR row reads the frozen Qwen2.5-7B-Instruct at revision \texttt{a09a35458c70} (Hugging Face; full hash in the released code). Trained on 2 and 3 relations, the block with one bit per relation is right on
36.9\% of the test stories, and the model alone on 38.0\% step by step and
34.3\% in one word; trained on 2 to 4, the model alone in one word is right on
31.3\%. DSR-LM reports GPT-3 without examples at 28.6\% on the split trained on 2 and
3\cite{zhang2023improved}. On that split the cost read is above DSR-LM's 60.98\% on two
of three seeds (59.6 to 61.1\%). With ten relations, the cost read, the bits and the
model alone are right on 48, 14 and 27\% on that
split. Given the graphs and trained on 2 and 3, the block is right on 1,139 of
1,146 test stories on each seed. The four earlier builds of the graph trainer, all scored on the test
stories, did worse. The first (split trained on 2 to 4, one seed) switched on all 8,000 possible rules and
was right on 6.7\%. The second learned the 62 true rules with 58 to
72 extra ones on the split trained on 2 to 4 (97.5 to 98.8\%
over three seeds) and 65 to 73 on the split trained on 2 and 3
(78.1 to 80.8\%). The third helped the second split (10 to
15 extra rules, 95.5 to 98.7\%) and broke the first
(89 to 111 extra, 46.6 to 62.9\%). The
fourth, designed after these test scores were seen, learned exactly the 62 rules (99.1\% and
99.2 to 99.4\%), and the final trainer differs from it on one seed by two stories. The rule
tables learned from the model's reading hold
74 to 81 rules, among them all 62 true ones in
3 of 6 runs. Our model-alone baseline is asked without examples beyond one
invented example of direction, reasons for up to 768 tokens and answers with one of the twenty relation words; examples
from the training stories or a larger model would score higher. With the read-in trained from answers, validation
accuracy is 89.8\% and 95.6\% on the two splits; before any training from answers,
the read-in, fitted to the model's top answer, is at 74.7\% and 81.2\%
on validation, close to the model's next-word reading (76.3\% and 80.2\%), but
at 45.8\% and 48.9\% on test, read once per split after the reported number
was fixed. Its top probability is exactly 1 on about 80\% of the test pairs, so costs of zero cannot
rank derivations and about half of its test answers are ties; on stories of 2 to 4 relations this costs little, on long
ones half the answers. We had forecast it within 2 points of the zero-shot reading on test; it is 10.7 and 11.7 points under. With ten relations the trained read-in is right on
56\% and 55\%; on the first split it loses ground at four relations
(57.6\% to 53.7\%). A rule table relearned from the trained read-in is not the benchmark's (80 to 113 rules,
58 to 62 of them true), and the bit block on the trained reading is right on only 33.5\% and
31.5\%; the reported rows therefore keep the zero-shot stage's table. As the code stored the table, the cost read keeps one rule per pair of input relations; keeping both rules in the one table where two apply would get one more test story right and leaves the mean at 60.6\%.

\paragraph{The committee.} Two forecasts failed: the new committee still disputes about a quarter of fresh changed
graphs, and on the first seed the gain at 128 nodes fell just short of the forecast. At 96 and 128 nodes the
committee-trained block scores 89.0\% and 80.0\% (three seeds).

\paragraph{Against every published model trained without hints.} Besides FloydNet's runs without hints, four papers
report CLRS-30 scores at 64 nodes for models trained on answers alone. Mahdavi et al.\ ran three processors on all thirty
tasks; the best of the three per task averages 63.24\%\cite{mahdavi2023towards}. Bevilacqua et al.\ report
a triplet message-passing network without hints on 18 tasks, averaging
80.49\%\cite{bevilacqua2023neural}, and Rodionov and Prokhorenkova raise it on the same tasks to
86.00\%, or 87.91\% with their contrastive term where they apply it\cite{rodionov2023neural}.
Georgiev et al.\ report a deep-equilibrium reasoner on 10 tasks, on test sets of their own, averaging
75.71\%\cite{georgiev2024deep}, and discrete neural algorithmic reasoning reports one task without hints,
breadth-first search, at 79\%\cite{rodionov2025discrete}. The general block is ahead on every task that
each of these covers: on the 18 tasks of Rodionov and Prokhorenkova it averages 97.76\%
(99.29\% in the setting the theory prescribes), on the 10 of Georgiev et al.\
96.06\% (98.82\%), and on all thirty of Mahdavi et al.\ 97.04\%. The best
score published without hints for each task, from any of these papers or from FloydNet, averages
94.85\%; against it the general block wins 13 tasks, ties 7 and
loses 10, and the setting the theory prescribes on the standard inputs wins 13, ties
8 and loses 9; every loss is to FloydNet's runs without hints. The per-task scores
are listed with the code.

\paragraph{One frontier reasoning model, as a probe.} We gave one frontier reasoning model, GPT-6 Astra (API identifier
gpt-6-astra), reasoning on and no tools, the official questions in the benchmark's text format\cite{markeeva2024clrs}: the first eight inputs of eight
tasks at 16 and 64 items. It answered 64 of 64 exactly at 16 items and
63 of 64 at 64, writing 3,759 to 23,575 tokens per
answer on the graph tasks at 64 items; the block takes 90 milliseconds per answer on a laptop processor.
This probes one model with one prompt; it is not a measure of frontier models in general.

\section*{Supplementary Table 4: CLRS-30 task by task}

The official 64-node test, task by task: the general block, one per task (mean of three seeds, lowest to highest); the
cold setting where it was run (mean, number of seeds; on the six path and tree tasks with the built route sums that
Results treats as a ceiling, on task scheduling with deadlines divided by the input's size and positions as given, and
on quickselect with extra training sizes and a loss from the second loop); the final shared block for all thirty tasks on
its registered read (every task after 8 loops, depth-first search after $n$ loops; mean of three seeds,
97.20\% on average); FloydNet's published score and its runs without hints; and the best other published score with its model, trained with or without hints. The
fresh 64-node split and the ablations are in the result files released with the models.
\begin{center}\small
\resizebox{\linewidth}{!}{\begin{tabular}{@{}l r r r r r r l@{}}
\toprule
Task & \multicolumn{3}{c}{Ours} & \multicolumn{2}{c}{FloydNet} & \multicolumn{2}{c}{Best other} \\
\cmidrule(lr){2-4}\cmidrule(lr){5-6}\cmidrule(l){7-8}
 & general (3 seeds) & cold (seeds) & shared (3 seeds) & reported & no hints & score & model \\
\midrule
\multicolumn{8}{@{}l}{\emph{Graphs}} \\
Breadth-first search & 100.00 (100.00--100.00) & -- & 100.00 & 100.00 & 100.00 & 100.00 & RANR \\
Depth-first search & 72.97 (70.80--74.76) & 100.00 (3) & 79.62 & 100.00 & 75.10 & 100.00 & Triplet-GMPNN, RANR \\
Topological sort & 99.99 (99.98--100.00) & 100.00 (3) & 99.67 & 100.00 & 99.75 & 100.00 & FloydNet (small) \\
Articulation points & 100.00 (100.00--100.00) & -- & 100.00 & 100.00 & 100.00 & 97.97 & G-ForgetNet \\
Bridges & 100.00 (100.00--100.00) & -- & 100.00 & 100.00 & 100.00 & 99.45 & FloydNet (small) \\
Strong components & 90.61 (89.55--92.58) & 90.02 (3) & 88.00 & 90.07 & 90.07 & 89.27 & FloydNet (small) \\
\multicolumn{8}{@{}l}{\emph{Paths and trees}} \\
Bellman--Ford & 99.67 (99.66--99.71) & 100.00 (3) & 99.51 & 100.00 & 100.00 & 99.18 & G-ForgetNet \\
Dijkstra & 99.63 (99.61--99.66) & 100.00 (3) & 99.50 & 97.67 & 99.40 & 99.14 & G-ForgetNet \\
DAG shortest paths & 99.84 (99.80--99.85) & 100.00 (3) & 96.84 & 100.00 & 100.00 & 99.37 & G-ForgetNet \\
Floyd--Warshall & 99.71 (99.70--99.73) & 100.00 (3) & 99.22 & 96.20 & 96.20 & 81.40 & FloydNet (small) \\
Kruskal & 99.49 (99.41--99.57) & 100.00 (3) & 99.28 & 99.63 & 99.63 & 93.47 & FloydNet (small) \\
Prim & 98.93 (98.73--99.07) & 99.98 (3) & 97.48 & 99.67 & 99.65 & 95.19 & G-ForgetNet \\
\multicolumn{8}{@{}l}{\emph{Order}} \\
Insertion sort & 100.00 (100.00--100.00) & 100.00 (3) & 100.00 & 100.00 & 100.00 & 98.40 & G-ForgetNet \\
Bubble sort & 100.00 (100.00--100.00) & 100.00 (3) & 99.98 & 100.00 & 100.00 & 95.78 & RANR \\
Heapsort & 100.00 (100.00--100.00) & 100.00 (3) & 100.00 & 100.00 & 100.00 & 99.00 & FloydNet (small) \\
Quicksort & 99.98 (99.95--100.00) & 100.00 (3) & 100.00 & 100.00 & 100.00 & 99.80 & FloydNet (small) \\
Minimum & 100.00 (100.00--100.00) & 100.00 (3) & 99.98 & 100.00 & 100.00 & 99.26 & G-ForgetNet \\
Binary search & 98.89 (98.44--99.12) & 100.00 (3) & 98.80 & 94.47 & 94.47 & 85.96 & G-ForgetNet \\
Quickselect & 99.77 (99.71--99.85) & 100.00 (5) & 99.82 & 80.33 & 80.33 & 87.08 & RANR \\
\multicolumn{8}{@{}l}{\emph{Greedy}} \\
Activity selector & 99.40 (99.10--100.00) & 99.94 (3) & 99.16 & 95.30 & 93.40 & 99.03 & G-ForgetNet \\
Task scheduling & 89.00 (88.84--89.07) & 99.86 (3) & 99.07 & 91.05 & 91.50 & 88.08 & RANR \\
\multicolumn{8}{@{}l}{\emph{Dynamic programming}} \\
Maximum subarray (Kadane) & 87.86 (87.06--89.01) & -- & 87.55 & 86.20 & 80.57 & 83.53 & RANR \\
LCS length & 95.66 (95.53--95.77) & -- & 92.43 & 95.70 & 86.10 & 91.70 & FloydNet (small) \\
Matrix chain order & 99.80 (99.78--99.85) & -- & 99.64 & 91.50 & 85.15 & 91.89 & RT \\
Optimal BST & 82.71 (77.86--85.19) & -- & 82.44 & 82.87 & 74.50 & 83.58 & G-ForgetNet \\
\multicolumn{8}{@{}l}{\emph{Strings and geometry}} \\
KMP matcher & 99.15 (98.29--99.66) & -- & 99.63 & 99.43 & 99.43 & 27.37 & FloydNet (small) \\
Naive string matcher & 99.07 (98.54--99.80) & -- & 99.69 & 100.00 & 100.00 & 100.00 & FloydNet (small) \\
Graham scan & 99.80 (99.64--99.88) & -- & 99.56 & 99.63 & 99.63 & 97.67 & G-ForgetNet \\
Jarvis march & 99.92 (99.88--100.00) & -- & 99.52 & 99.50 & 99.50 & 99.00 & FloydNet (small) \\
Segments intersect & 99.33 (98.97--99.56) & -- & 99.67 & 99.90 & 99.90 & 99.06 & G-ForgetNet \\
\midrule
Average & 97.04 & & 97.20 & 96.64 & 94.81 & 92.69 & \\
\bottomrule
\end{tabular}
}
\end{center}

\section*{Supplementary Note 5: certificates for trained blocks}

\paragraph{The block as a table.} The blocks of Section~8 keep bits for each pair of items, which are mentions of people
or, on the first task, lines. One loop turns bit $c$ on at $(a, b)$ if some item $k$ passes test $c$ at $(a, k)$ and at
$(k, b)$, and bits stay on. A test adds a number looked up from the pair's pattern, which is everything the block is given
about the pair (the relation between its items: none, the arrow of a statement or the arrow of the question; whether they
name the same person; whether each is in the question), to non-negative weights on the kept bits that are on, and passes
above zero. A trained block is therefore a finite table: for each pattern and each set of kept bits, which tests pass.
The read-out says yes on some sets of kept bits at the question's pair, and more bits never turn a yes into a no. The
table is computed in the arithmetic of the trained code; Supplementary Table~3 gives each group's smallest margin over
the entries used.

\paragraph{Never a false yes.} Take any context whose answer is no, in which each statement and the question name two
different people. Call a person reachable if it is the question's first person $X$ or a descendant of $X$. The template
has one mention of each kind of person in each place of each kind of statement (reachable to reachable, unreachable to
unreachable, unreachable to reachable; none runs from reachable to unreachable, and $Y$ is unreachable, since the answer
is no), and the
question; it is run to rest. Each item of the context is sent to the template's item of the same kind and place. By
induction on the loops, the bits a pair of the context keeps are among those its image keeps at rest: if bit $c$ turns
on at $(a, b)$ through $k$, test $c$ passed at $(a, k)$ and at $(k, b)$ with bits that the images keep; a test only gains
from bits; and the table, enumerated over every pattern that a pair and its image can have, shows that the test passes
on the images too, so the image of $(a, b)$ already holds $c$. If the read-out says no on the template, it says no on the
context. The plainer statement, that a yes survives every map of the items that keeps the input bits, fails for these
blocks: ``same person'' excludes the same mention, so merging two mentions of one person turns it off.

\paragraph{Every yes in time.} Take a context without a cycle whose answer is yes. It holds a chain of parent links from
$X$ to $Y$, and the mentions of the chain's statements and of the question carry among themselves the patterns of the
bare chain, the context with only those statements; further items only add bits. It is therefore enough that every
bare chain says yes in time. We look for a set $S$ of kept bits that single links carry after $t_0$ loops and that a
pair of the chain carries one loop after the two halves of its split at the middle carry it, counting a test as passing
on a kind of pair only if it passes on every pattern that kind can have. A pair $L$ links apart then carries $S$ after
$t_0 + \lceil \log_2 L \rceil$ loops, and the question's pair one split later. The script checks this count against the
registered number of loops for every length up to 65,536; both grow as $\log_2 L$, so the check holds at every length.
Some blocks keep no bit on a single statement, raise a bit through an item's pair with itself, or attach the question's
two ends only through the last step; for them a wider argument runs every chain of up to 48 items, with pairs grouped
into classes, to find a loop by which each pair holds each bit, and carries the halving induction from distances 8 to
15 to every larger distance. A block counts as proven at every size if either argument succeeds within the registered
loops. The plain halving argument proves 21 of the 23 blocks that learned and the wider
one all 23; the proved loop count is 1 to 3 loops under the registered
one. We widened the argument three times after seeing which blocks it missed (splits next to the question, raising
through an item's pair with itself, and separate patterns for the two pairs next to the question); each widening only
adds ways to prove, and the direct runs of the chains were never changed. The script also runs the bare chains
themselves, to 500 statements on the kinship task and 1,000 facts on the first.

\paragraph{Lean.} For the nine runs of the first layout on the 7B and 1.5B readings, which have five
distinct tables, both checks are proved in Lean 4 in a form with separate caps at the two ends of the chain (theorems
\texttt{correct\_<run>}, using only the axioms \texttt{propext}, \texttt{Classical.choice} and \texttt{Quot.sound}). The
table itself is read from the weights outside Lean, by evaluating the block on every pattern and set of kept bits; its
signs agree in single, double and 50-digit arithmetic.

\paragraph{What the checks found.} Supplementary Table~3 lists every start. The table computes the trained code's own
answer on 40 random contexts per network, and every network that learned is right on
each of 3,000 random contexts, cycles included, both read at rest and read at the registered loops. Of
the five starts that did not learn, three fail the template check. The other two
pass it and fail on chains: one never answers a chain, and one is a walker that answers a chain of $k$ statements at
loop $2k$ (loops 8, 96 and 200 for 4, 48 and 100 statements,
against 7, 10 and 11 registered). Read at rest the
walker is right on all 3,000 random contexts; read at the registered loops it is wrong on
4 of them. No trained network among the 28 keeps squarings in separate bits that
stop at a fixed length. The
networks also differ at one corner that the theorems exclude, a question whose two names are the same person: of the
20 kinship networks that learned, 6 say yes always, 13 only
if that person is also named in a statement, and 1 never; the 3 networks of the
first task say no in every such case. Some networks thus read ``ancestor'' as ``ancestor or self'', and others strictly.

\section*{Supplementary Table 3: the certificate for every trained block}

Every start of Section~8, grouped by the model read and the layout. Learned: passed the validation check of
Methods. Never a false yes: the template check passes. Chains in time: every bare chain run says yes within the
registered loops. Every size: one of the two arguments for chains holds within the registered loops, so the block is proven right on
every context of every size without a cycle, for questions about two different people, given its input bits. Smallest margin: the smallest distance from zero of any table entry
used, among the networks that learned.
\begin{center}\small
\resizebox{\linewidth}{!}{\begin{tabular}{@{}l r r r r r r@{}}
\toprule
Runs & Starts & Learned & Never a false yes & Chains in time & Every size & Smallest margin \\
\midrule
Qwen2.5-0.5B, ``older than'', one wording (line items) & 3 & 3 & 3 & 3 & 3 & 0.0006 \\
Qwen2.5-7B, kinship, pooled reading, first layout & 3 & 3 & 3 & 3 & 3 & 0.0063 \\
Qwen2.5-7B, kinship, pooled reading, second layout & 3 & 3 & 3 & 3 & 3 & 0.0031 \\
Qwen2.5-7B, kinship, sentence by sentence, first layout & 3 & 3 & 3 & 3 & 3 & 0.0151 \\
Qwen2.5-7B, kinship, sentence by sentence, second layout & 3 & 3 & 3 & 3 & 3 & 0.0028 \\
Qwen2.5-1.5B, kinship, pooled reading, first layout & 3 & 3 & 3 & 3 & 3 & 0.0063 \\
Qwen2.5-0.5B, kinship, directions learned from answers & 10 & 5 & 5 & 5 & 5 & 0.0024 \\
\midrule
All & 28 & 23 & 23 & 23 & 23 & 0.0006 \\
\bottomrule
\end{tabular}
}
\end{center}

\section*{Supplementary Note 6: the sharp line for constraint problems}

\paragraph{Setting.} Fix a finite constraint language $\Gamma$, finitely many relations of bounded arity on a finite
domain $D$. The network sees a position-free encoding of an instance: for relations on two variables, one input label per
ordered pair of variables (which relations of $\Gamma$ hold on the pair, and on the variable with itself); for higher
arity, one token per constraint, labelled by its allowed tuples, with pair labels that say which variable sits at which
place of which constraint (a pair's label includes the labels of its two ends, so the pair $(C, x)$ also carries the relation of $C$). No core or idempotency condition is needed, since the problem and bounded width are both
unchanged by taking the core of $\Gamma$.

\paragraph{Ceiling.} Let a network's accept or reject answer be a function of the stable colouring that the
$k$-dimensional Weisfeiler--Leman refinement gives the encoding, for a fixed $k$ ($k = 2$ for our block). This allows any
weights, joins that test existence, keep the best witness or count, any number of loops, any read-out, and even a
different network for each size. If such networks decide the problem on every instance, $\Gamma$ has bounded width.
(a) The $k$-dimensional refinement distinguishes exactly what counting logic with $k + 1$ variables, $C^{k+1}$,
distinguishes\cite{cai1992optimal}, on any relational structure\cite{grohe2021logic}. (b) With pair labels the encoding
is the instance itself, each ordered pair labelled with the relations that hold on it, so instances that $C^{k+1}$ does
not tell apart have encodings that the refinement does not tell apart. With constraint tokens the encoding is an
interpretation of the instance of dimension at most the largest arity $r$, a token standing for the tuple of its
scope, and instances that $C^{m}$ does not tell apart, for some $m$ that depends on $r$ and $k$ only, have encodings
that $C^{k+1}$ does not tell apart; this is the standard transfer of counting logic through interpretations, which we
sketch rather than prove. (c) So the set of instances that the networks accept is closed under equivalence in $C^{m}$
($m = k + 1$ for pair labels), and a class of finite structures closed under that equivalence is definable in
infinitary counting logic with $m$ variables. (d) Atserias, Bulatov and Dawar show that systems of equations on three
variables over any non-trivial finite abelian group are not definable in infinitary counting logic with any fixed number
of variables\cite{atserias2009affine} (their Theorem 4). Their construction places a twist on a graph of large
treewidth, as Cai, F\"urer and Immerman did, and gives a solvable and an unsolvable system that the logic cannot tell
apart. They carry this to every $\Gamma$ whose algebra of polymorphisms admits the unary or affine type, by a gadget
reduction from such equations to $\Gamma$ that a Datalog program computes and counting logic can therefore follow
(their Theorem 15 and Corollary 16). Barto and Kozik proved that every $\Gamma$ whose algebra omits both types has
bounded width\cite{barto2014local}; all of this may be applied to the core of $\Gamma$, which poses the same problem. So
if $\Gamma$ does not have bounded width, its problem is not definable in counting logic, which contradicts (c). For pair
labels the proof is complete given the cited theorems; with constraint tokens it also rests on the sketched step (b).

\paragraph{Reach.} If $\Gamma$ has bounded width, Barto's theorem gives relational width $(2, 3)$: an instance has a
solution exactly when its $(2,3)$-minimal reduct, the largest sub-instance in which every allowed pair of values on two
variables extends to every third variable, has no empty constraint\cite{barto2016collapse}. One looped cold block with
fixed weights computes that reduct, using only its best-witness half, whose bits are kept.

\emph{State.} Kept bits only turn on (Supplementary Note~1), so the block records what the check has ruled out, not what
it still allows: for each ordered pair of variables $(x, y)$, the set $F_{xy} \subseteq D^2$ of value pairs ruled out,
$|D|^2$ bits, which the network applied to each pair computes from the input label and the kept bits. At the start
$F_{xy}$ holds the pairs that a constraint on $(x, y)$ or on $(y, x)$ excludes, and $F_{xx}$ holds every $(a, b)$ with
$a \neq b$ and every $(a, a)$ that a constraint on $x$ alone excludes.

\emph{Update.} The check rules $(a, b)$ out on $(x, y)$ when, for some variable $z$, every value $c$ satisfies
$(a, c) \in F_{xz}$ or $(c, b) \in F_{zy}$. The condition ``for every $c$'' is not one test on each leg, but it is an OR,
over the subsets $U \subseteq D$, of such a pair of tests:
\[
\forall c \,\big[(a, c) \in F_{xz} \,\vee\, (c, b) \in F_{zy}\big]
\iff
\exists U \subseteq D:\ \big(\forall c \notin U:\ (a, c) \in F_{xz}\big) \wedge \big(\forall c \in U:\ (c, b) \in F_{zy}\big),
\]
where, from left to right, $U$ is the set of $c$ with $(a, c) \notin F_{xz}$. The block has one channel for each
$(a, b, U)$, $|D|^2 2^{|D|}$ in all. Channel $(a, b, U)$ admits the middle item $z$ when the first conjunction holds on
the leg $(x, z)$ and the second on the leg $(z, y)$; its bit is kept; and $F_{xy}$ gains $(a, b)$ once any of its
$2^{|D|}$ channels has fired. Each test is a conjunction of bits, which is one threshold: coding each bit as a pair of
coordinates, $(+1, -1)$ or $(-1, +1)$, keeps the mean and the spread of every state fixed, so the normalisation before
the tests is a fixed scaling; and the network applied to each pair computes $F$ exactly, as a ReLU network computes any
Boolean function of finitely many bits.

\emph{Read-out.} No, when some variable has every value ruled out ($F_{xx}$ contains every $(a, a)$).

\emph{Correctness.} For constraints on two variables the reduct is fixed by its pairs: its constraint on three variables
$\{x, y, z\}$ is the set of triples whose three pairs it allows, so $|D|^2$ bits per pair suffice. The update is
monotone and only adds bits, so its result does not depend on the order in which the check is run, and by induction on
the loops the block never rules out a pair that the reduct allows. At rest the pairs not ruled out satisfy the check, so
they form a $(2,3)$-minimal sub-instance and lie within the reduct, which is the largest one. A variable has every
value ruled out exactly when the reduct has an empty constraint.

\emph{Loops.} The kept bits of the next loop are a function of $F$ and the present kept bits. If a loop leaves $F$
unchanged, the loop after it adds no kept bit. Every earlier loop adds a pair to some $F_{xy}$, so $F$ is final within
$n^2 |D|^2$ loops and the block is at rest within $n^2 |D|^2 + 2$.

\emph{Constraints on three or more variables.} Pairs alone are not enough: the pair projections of a Horn clause on three
variables, or of a parity equation, are full\cite{barto2016collapse}. Each constraint $C$, of arity $r_C$, then has a
token, and its ruled-out tuples $G_C \subseteq D^{r_C}$ are kept on the pairs $(C, x)$, one copy for each variable $x$ of
its scope, $|D|^{r_C}$ bits per copy. Three more kinds of channel, each with one test on each leg, do the rest. A tuple
$t$ is ruled out on $(C, y)$ when some variable $k$ at place $p$ of $C$ has $(t_p, t_q) \in F_{ky}$, with $y$ at place
$q$; the network applied to the pair $(C, y)$ reads $y$'s places from its label. A tuple ruled out on one copy
$(C, k)$ is ruled out on every copy $(C, y)$. And $(a, b)$ is ruled out on $(x, y)$ when, for a token $C$ with $x$ at
place $p$ and $y$ at place $q$, every tuple $t$ with $t_p = a$ and $t_q = b$ is ruled out on $(C, y)$. A deduction can
now take two loops, from a pair of variables to one copy and from that copy to the others, but the count is unchanged in
kind: if a loop leaves every ruled-out pair and tuple unchanged, the block rests one loop later. So it rests within
$n^2 |D|^2 + \sum_C r_C |D|^{r_C} + 2$ loops, with $n$ the number of variables. At rest the pairs and tuples not ruled out
form the reduct, by the argument above, so the block decides the problem. The number of channels depends on $\Gamma$
alone.

\paragraph{Loops for 2-SAT and two-colouring.} On two values a ruled-out pair is an implication between literals:
$(a, b) \in F_{xy}$ says that $x = a$ forces $y \neq b$. The channel with $U = \{c\}$ composes two of them, $x = a$
forcing $z = c$ and $z = c$ forcing $y \neq b$. By induction on $t$, splitting a path at its middle, every implication
along a path of at most $2^t$ steps is in $F$ after $t$ loops. A contradiction $L$ implications deep, $x = 1$ forcing
$x = 0$ and $x = 0$ forcing $x = 1$ along paths of at most $L$ steps, therefore rules out every value of $x$ by loop
$\lceil \log_2 L \rceil$, which matches the measured growth of one loop per doubling of depth (Section~6). We checked
the construction against exhaustive search on random instances of 2-SAT, two-colouring, a three-valued language closed
under the minimum, and Horn clauses on three variables written as tokens: every answer was right, every loop count
within its bound, and every 2-SAT and two-colouring contradiction found by loop $\lceil \log_2 L \rceil$. On parity
equations over three variables, which do not have bounded width, the same block accepted every unsolvable instance it
was given.

\paragraph{What is ours.} The two directions rest on the cited theorems. The statement for pair networks, the
construction of the block and the learned results are ours; we found no earlier statement of the line for networks,
and no network reported to learn a class of bounded width exactly from yes-or-no answers.

\end{document}